\documentclass[10pt,twocolumn,letterpaper]{article}

\usepackage{iccv}
\usepackage{times}
\usepackage{epsfig}
\usepackage{graphicx}
\usepackage{amsmath}
\usepackage{amssymb}
\usepackage{multirow}
\usepackage{multicol}
\usepackage{subfigure}
\usepackage{appendix}

\usepackage[pagebackref=true,breaklinks=true,letterpaper=true,colorlinks,bookmarks=false]{hyperref}

\iccvfinalcopy 

\def\iccvPaperID{7182} 

\ificcvfinal\fi

\begin{document}

\title{CANet: A Context-Aware Network for Shadow Removal}

\author{Zipei Chen\textsuperscript{1}, 
Chengjiang Long\textsuperscript{2}\thanks{This work was co-supervised by Chengjiang Long and Chunxia Xiao.}, 
Ling Zhang\textsuperscript{3}, 
{Chunxia Xiao\textsuperscript{1}}$^{\tt\small*}$\thanks{Corresponding author.} \\
\textsuperscript{1}{School of Computer Science, Wuhan University, Wuhan, Hubei, China} \\
\textsuperscript{2}{JD Finance America Corporation, Mountain View, CA, USA} \\
\textsuperscript{3}{Wuhan University of Science and Technology, Wuhan, Hubei, China} \\
{\tt\small czpp19@whu.edu.cn, cjfykx@gmail.com, zhling@wust.edu.cn, cxxiao@whu.edu.cn}
}

\maketitle
\ificcvfinal\thispagestyle{empty}\fi

\begin{abstract}
In this paper, we propose a novel two-stage context-aware network named CANet for shadow removal, in which the contextual information from non-shadow regions is transferred to shadow regions at the embedded feature spaces. At Stage-I, we propose a contextual patch matching (CPM) module to generate a set of potential matching pairs of shadow and non-shadow patches. Combined with the potential contextual relationships between shadow and non-shadow regions, our well-designed contextual feature transfer (CFT) mechanism can transfer contextual information from non-shadow to shadow regions at different scales. With the reconstructed feature maps, we remove shadows at L and A/B channels separately. At Stage-II, we use an encoder-decoder to refine current results and generate the final shadow removal results. We evaluate our proposed CANet on two benchmark datasets and some real-world shadow images with complex scenes. Extensive experimental results strongly demonstrate the efficacy of our proposed CANet and exhibit superior performance to state-of-the-arts. Our source
code is available at  \color{magenta}{\url{https://github.com/Zipei-Chen/CANet}}.
\end{abstract}

\section{Introduction}
Shadow is a natural phenomenon appearing when the light is partially or completely blocked. As a fundamental challenge in the field of computer vision, the existence of shadow in images or videos inevitably degrades the accuracy and effectiveness of general application tasks such as intrinsic image decomposition~\cite{intrinsic_image_decomposition1, intrinsic_image_decomposition3}, visual recognition~\cite{Long:CVPR2017, Hua:TPAMI2018, Long:WACV2019, Hu:TIP2021}, object detection and tracking~\cite{object_detection1, object_detection2, object_detection3}, trajectory prediction~\cite{manh2018scene, Shi:CVPR2021}, single image super-resolution~\cite{Zhang:ICME2020, Zhang:TCSVT2021} and image captioning~\cite{Dong:MM2021}. Therefore, shadow removal is important and necessary to improve the visual effects and avoid the performance drop on the above-mentioned computer vision tasks. However, due to the complex interactions of geometry and illumination, shadow removal remains a challenging problem.



Current shadow removal methods can be mainly divided into two categories: physical-based methods~\cite{tradition_removal1, tradition_removal2, Guo_2011, tradition_removal3, tradition_removal4, Xiao_2013, Zhang_2015} and learning-based methods~\cite{Deshadownet, Stc-gan, DSC, TBGANs, ARGAN, RIS-GAN, siggraph}. Compared to physical-based methods, which apply a physical model to analyze each pixel's intensities,
learning-based methods analyze the image in feature maps. Recently, learning-based methods with proper model have presented potential advantages~\cite{RIS-GAN,DSC,ShadowDetectionandRemovalfromaSingleImageUsingLABColorSpace}. However, these methods mainly focus on increasing the receptive field of the model without considering other particular context-sensitive shadow-aware components, which may easily ignore the contextual matching information hidden in images. 

\begin{figure*}[ht]
\vspace{-0.5cm}
\begin{center}
\includegraphics[width=1\linewidth]{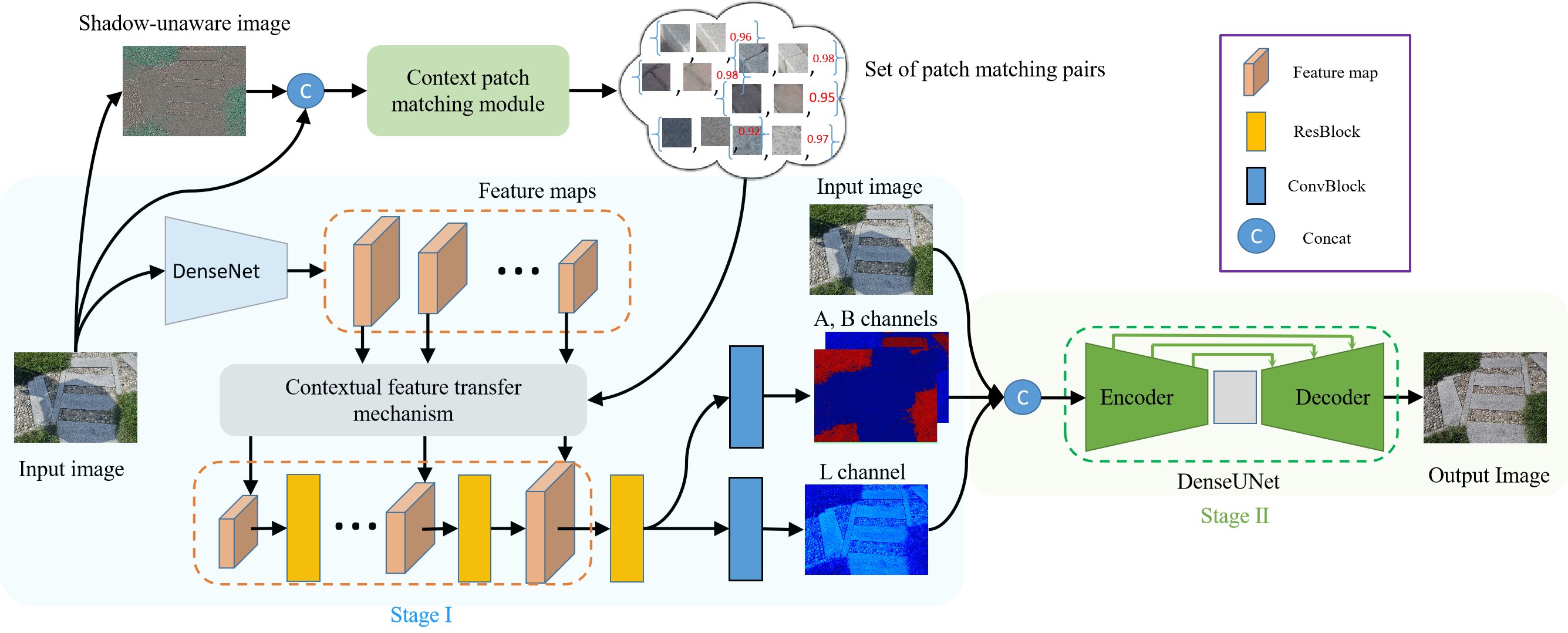}
\end{center}
\vspace{-0.3cm}
 \caption{The overview of our proposed CANet, which takes two stages for shadow removal. At Stage-I, the contextual feature is firstly extracted via a pretrained DenseNet~\cite{Densenet}; meanwhile, the designed contextual patch matching module (CPM) (see Figure~\ref{fig:CPM_network}) is used to acquire a set of contextual matching pairs; then, applying a contextual feature transfer mechanism (see Figure~\ref{fig:CTF_strategy}) to transfer contextual information from non-shadow patches to shadow patches to recover the L and A/B channels of the shadow-removal image. At Stage-II, we integrate the recovered L and A/B channel information with the input shadow image and feed them into a DenseUNet to generate the final shadow-removal result.
  }
\label{fig:model_overiew}
\vspace{-0.3cm}
\end{figure*}

In this paper, we propose a novel two-stage context-aware network CANet for shadow removal in an end-to-end manner. As shown in Figure~\ref{fig:model_overiew}, our CANet integrates a contextual patch matching (CPM) module and a contextual feature transfer (CFT) mechanism at Stage-I and takes Stage-II as a refinement step for shadow removal. In particular, the CPM module is designed to search for the corresponding potential relationships between shadow and non-shadow patches, which demonstrates the contextual mapping between shadow and non-shadow regions. The CFT mechanism is utilized to transfer the contextual feature at different scales from non-shadow regions to shadow regions based on the output patch matching pairs from the CPM module and the extracted contextual features.




Our CPM module is designed as a dual-head structure network with the shared patch feature extractor to predict the degree of context matching between two patches from the image, as well as determine the type of the patch pair without a shadow mask. We only focus on contextual information transfer from non-shadow regions to shadow regions. Therefore we can define three types of patch pairs, {\em i.e.}, (1) both from shadow or non-shadow regions, (2) the first one from the shadow region and the second one from the non-shadow region, and (3) the first one from the non-shadow region and the second one from the shadow region. With these prediction types, we can filter out most irrelevant patch pairs.
Unlike those traditional patch match methods, our CPM is learning-based to adaptively handle complex scenes by data-driven and can effectively avoid matching errors caused by shadows the impact of shadows by averaging the lightness. What's more, our type classification head can be used to filter out mostly patch pairs from the same shadow or non-shadow regions and only focus on other pairs with high correlation scores.

We train the CPM module with sizeable self-collected training data and apply the learned CPM module to obtain a set of patch matching pairs. Then, inspired by the idea of information transfer~\cite{2002Color}, we introduce a contextual feature transfer (CFT) mechanism to transfer the contextual feature at different scales from non-shadow patches to shadow patches, resulting in a series of feature maps without shadow information. Different from the existing information transfer strategies used in shadow removal task~\cite{Xiao_2013, Zhang_2015, Zhang:CGF2020}, which search a most relevant non-shadow patch/sub-region for each shadow patch/sub-region, our CFT mechanism performs feature transfer by applying several patch matching pairs for one shadow patch according to the similarity between two patches.
With the reconstructed shadow-less feature maps, we remove shadows in the L and A/B channels separately at Stage-I. Finally, to ensure the robustness of our results, with the recovered L and A/B channel images and the shadow image as inputs, we use an encoder-decoder to predict the final shadow removal image at Stage-II.

In summary, our main contributions are three-fold as follows:
\begin{itemize}
\item We propose a two-stage context-aware network CANet for shadow removal in an end-to-end manner, in which the contextual information from non-shadow regions is transferred to shadow regions at the embedded feature spaces. 
\end{itemize}
 
\begin{itemize}
\item We design and train a context patch matching (CPM) module to acquire the potential contextual relationships between shadow and non-shadow regions in the image, which automatically distinguishes shadow patches from non-shadow patches during the matching processing.
\end{itemize}

\begin{itemize}
\item The proposed contextual feature transfer (CFT) mechanism transfers the extracted contextual features from non-shadow regions to shadow regions in different scales, which remove features associated with shadows and produce superior shadow removal results.
\end{itemize}

Both quantitative and qualitative experiments demonstrate the effectiveness and efficiency of our proposed CANet, as well as its superior performance in generating realistic shadow-removal images.


\section{Related work}
\textbf{Physical-based methods for shadow removal.}
Physical-based methods for shadow removal are traditional methods, which usually formulate a physical model using some prior knowledge to recover illumination in shadow regions \cite{tradition_removal3,Guo_2011,tradition_removal2,Xiao_2013,tradition_removal3,tradition_removal4}. Finlayson {\em et al.} proposed a series of shadow removal methods \cite{tradition_removal1, tradition_removal2} based on gradient consistency, which reconstructs the shadow removal images based on the prior that the gradient information of the image is immutable after shadow removal. 
Due to the change of illumination, these methods can present obvious shadow boundary artifacts. 

Another strategy is information transfer, which transfers information like color or light from one picture/region to another picture/region. It has been widely used in image process tasks. Wen {\em et al.}~\cite{2008Example} proposed a user-interactive multiple local color transfer method, which sets a proper gradient-guided color transfer function for each pixel. Zhang {\em et al.}~\cite{2013A} incorporated the color-transfer techniques and gradient fusion method to altering the illumination effects of an image from another. Shor {\em et al.}~\cite{2008_trandition} build a linear mapping model between the shadowed and non-shadow regions. 
Xiao {\em et al.}~\cite{Xiao_2013} conducted shadow removal task using sub-region matching illumination transfer. Zhang {\em et al.}~\cite{Zhang_2015} proposed a local-to-global shadow removal method based on illumination transfer. Although those methods using information transfer can produce pleasant shadow removal results, the effectiveness of these methods depends on the accuracy of texture matching.


\textbf{Learning-based methods for shadow removal.}
Unlike traditional physical-based methods, learning-based methods tend to learn high-level contextual features for shadow removing~\cite{DSC,Zhang:CGF2020,shadow_decomposition,RIS-GAN}. Qu {\em et al.}~\cite{Deshadownet} proposed a multi-context embedding network DeshadowNet to integrate the information from different levels for shadow removing. Wang {\em et al.}~\cite{Stc-gan} analyzed the relationship of shadow detection and removal and then proposed a stacked conditional generative adversarial network (ST-CGAN) model to perform shadow detection and removal jointly. Hu {\em et al.}~\cite{DSC} used direction-aware spatial context attention features for shadow detection and removal. Zhang {\em et al.}~\cite{RIS-GAN} explore relationship of the residual and inverse illumination for shadow removal and proposed a general RIS-GAN. Hieu {\em et al.}~\cite{shadow_decomposition} regard shadow image as the combination of shadow-free image, shadow parameters and shadow matte, and use neural networks to predict them to remove shadows. Liu {\em et al.}~\cite{2020Shadow} presented a LG-ShadowNet for shadow removal by training on unpaired data. Lin {\em et al.}~\cite{2020BEDSR} proposed a BEDSR-Net for document shadow removal. The BEDSR-Net is specifically designed for document image shadow removal, which may be lack of expandability for other kinds of shadow images. Cun {\em et al.}~\cite{2020Towards} designed a network named SMGAN for shadow-removal, which can produc ghost-free shadow-removal images.
Although those existing methods achieve some advances, they only focus on increasing the receptive field of the model, ignoring the paired matching information in the image. In contrast, our proposed CANet is proposed to explore the underlying contextual information between shadow and non-shadow regions. 


  


\section{Method}
Intuitively, two patches with similar textures should have similar illumination and context under the same shadow-free environment. Based on this, we explore the idea of context-aware information transfer to conduct our shadow removal task. The proposed two-stage context-aware network (CANet) for shadow removal is illustrated in Figure \ref{fig:model_overiew}. At Stage-I, given a shadow image, the L and A/B channels of a shadow-removal image are recovered with contextual information transferred from non-shadow patches, relying the obtained contextual patch matching information. At Stage-II, a DenseUNet is designed to integrate the recovered L and A/B channel information with the input shadow image to generate a high-quality shadow-removal image. 


In the following subsections, we will introduce our contextual patch matching module, contextual feature transfer mechanism, and the two-stage CANet for shadow removal.

\subsection{Contextual Patch Matching Module}
As shown in Figure~\ref{fig:model_overiew}, the CPM module is designed to generate a set of ordered patch matching pairs of non-shadow and shadow patches with the prediction correlation scores together. 
\begin{figure}[ht]
\begin{center}
\includegraphics[width=0.90\linewidth]{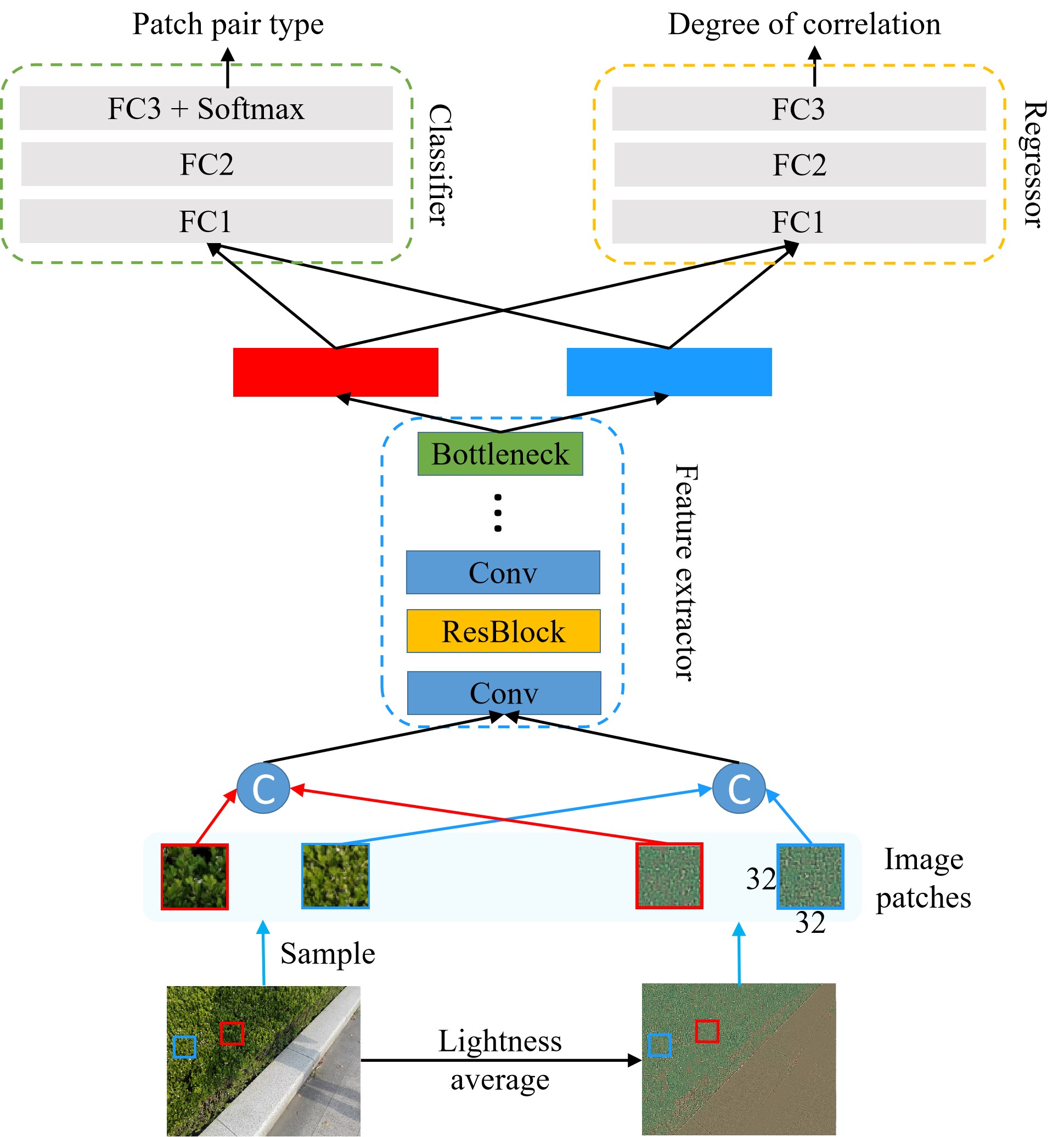}
\end{center}\vspace{-0.3cm} 
   \caption{The architecture of our dual-head contextual patch matching module. We first calculate the shadow-unaware image by lightness average with a mean-filter operation. Then, we extract $32\times32$ patches from the shadow image and the shadow-unaware image, and feed them into the shared patch descriptor to extract deep learning features. Finally, the extracted features are fed into the two heads to predict the type (denoted as -1, 0, or 1) of the input patch pair and the corresponding degree as a continuous value in the range of 0 to 1 to measure their correlation.}
\label{fig:CPM_network}
\vspace{-0.3cm}
\end{figure}

\begin{figure*}[t]
\vspace{-0.5cm}
\begin{center}
\includegraphics[width=1\linewidth]{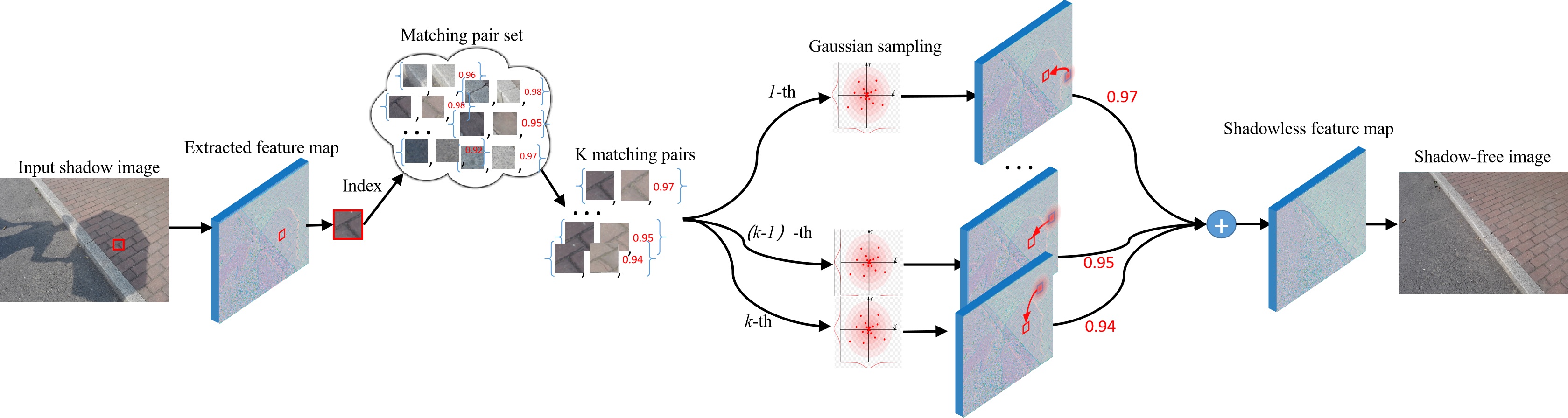}
\end{center}\vspace{-0.3cm}
   \caption{The illustration of how we use our contextual feature transfer mechanism to transfer the contextual information according to captured contextual matching pairs. Note that we use the Gaussian sampling to sample contextual information from non-shadow patches and integrate them to the query shadow patches adaptively.}
\label{fig:CTF_strategy}
\vspace{-0.3cm}
\end{figure*}
To avoid matching errors caused by shadows, given an input shadow image, we first apply a mean-filter with kernel size 3 to get a shadow-unaware image by averaging the lightness on the shadow image. Especially, the shadow-unaware image is calculated as:
\begin{equation}
    I_{i, j} = I_{i, j} - \frac{\begin{matrix} \sum_{(i, j)\in P} I_{i, j} \end{matrix}}{N} + I_{avg}
\end{equation}
where $I_{i, j}$ denotes the lightness value at pixel $(i, j)$, $P$ denotes a $3\times 3$ patch around pixel $(i, j)$, $N$ is the total number of pixeles in the patch, and $I_{avg}$ is the global average lightness value of the image. 
We empirically observe that the shadow-unaware image is supplementary to the input shadow image as the input source for our CPM module, especially valuable to eliminate the effects of the shadow and distinguish shadow patches from non-shadow patches. More details can be seen in the Appendices~\ref{appendix}.

We then extract $32\times32$ patches from both the input shadow image and the shadow-unaware image are concatenated and fed into our dual-head CPM module, as illustrated in  Figure~\ref{fig:CPM_network}. 
Unlike traditional methods, which use the hand-craft local descriptor such as SIFT~\cite{SIFT} to represent image patches and apply the Euclidean distance to measure the similarity between them, our CPM is learning-based so that it can adaptively handle the complex scenes by data-driven. We make full use of convolution layers with residual blocks as the shared feature extractor to extract deep learning features for the dual heads to conduct a regression task and a classification task. To specify, one head is a correlation regressor to output a continuous value at the range from 0 to 1 as the degree of correlation, and the other one is a type classifier to predict the type of pair as one of three types denoted as -1, 0, and 1. 

It is worth mentioning here we care about the order between each output patch matching pair, especially when the two patches are not from the same regions (shadow or non-shadow). In particular, 0 indicates two patches from the same shadow or non-shadow regions, -1 indicates the patch pair starting with the non-shadow patch and then the shadow patch, while 1 indicates the patch pair starting with the shadow patch and then the non-shadow patch. Note that we use -1 and 1 to distinguish the shadow patch from the non-shadow patch, as we only transfer the contextual information from non-shadow regions to shadow regions. By doing this, our type classifier head can be used to filter out most of the patch pairs from the same shadow or non-shadow regions and only focus on other pairs with high correlation scores.

To learn a solid and robust CPM module, we make full use of the existing shadow benchmark datasets to collect a large training data set. We randomly sample $32 \times 32$ patches from shadow images. Note that we use cosine similarity as the measurement between shadow and non-shadow patches on the corresponding shadow-free images to generate ground-truth for correlation regression. For those with cosine similarity higher than 0.95, we set the ground-truth of correlation degree ${\bf s}_{gt}$ as 1 and 0 smaller than 0.6. For the ground-truth type ${\bf t}$, we utilize the ground-truth shadow mask to determine the ground-truth type to be -1, 0, or 1 for any patch pairs. 

 

We optimize the overall loss $\mathcal{L}_{CPM}$ to train the CPM module. It contains a regression loss $\mathcal{L}_{reg}$ and a classification loss $\mathcal{L}_{cls}$, {\em i.e.}, 
\begin{equation}
    \mathcal{L}_{CPM} = \mathcal{L}_{reg} + \mathcal{L}_{cls}
\end{equation}

The regression loss $\mathcal{L}_{reg}$ is used to optimize the correlation regressor for CPM module, {\em i.e.},
\begin{equation}
    \mathcal{L}_{reg} = ||{\bf s}_{out} - {\bf s}_{gt}||_2
\end{equation}
where ${\bf s}_{gt}$ is the correlation degree label for input pair and ${\bf s}_{out}$ is the output of the correlation regressor. 

The classification loss $\mathcal{L}_{cls}$ is used to optimize the type classifier, which is defined as a cross-entropy:
\begin{equation}
    \mathcal{L}_{cls} = -\sum\limits_{i=1}^{3}t_ilog(p_i) 
\end{equation}
where $t_i$ denotes the ground truth matching type of the patch pair and $p_{i}$ is the output of our type classifier.

\subsection{Contextual Feature Transfer Mechanism}
The target of our contextual feature transfer mechanism is to transfer the contextual feature from non-shadow regions to shadow regions, resulting in feature maps without shadow information. Generally, directly replacing the features from non-shadow regions to shadow regions may cause sub-optimal results, such as discontinuity, artifacts. Therefore, we introduce a contextual feature transfer (CFT) mechanism to perform information transfer in feature space.

The process of our contextual feature transfer model is shown in Figure \ref{fig:CTF_strategy}. Given a shadow patch from the input feature map, we first retrieve the matched non-shadow patches from the generated set of patch matching pairs. Then, for each shadow patch, we use Gaussian sampling to perform contextual feature transfer using the matched non-shadow patches. Finally, we integrate the top-$k$ transferred feature patches according to the correlation degree between each matching pair.


Let $n$ be the kernel size of Gaussian sampling and $k$ be the feature transfer times. The Gaussian sampling can be written as:
\begin{equation}
    F'_{x, y} = \sum\limits_{\Delta x=0}^{n} \sum\limits_{\Delta y=0}^{n}\frac{\varphi(\Delta x, \Delta y)}{\begin{matrix} \sum_{\Delta x=0}^n \begin{matrix} \sum_{\Delta y=0}^n \varphi(\Delta x, \Delta y) \end{matrix}  \end{matrix}} F_{x+\Delta x, y+\Delta y}
\end{equation}
where $F'_{x, y}$ and $F_{x, y}$ are the feature maps after and before sampling at position $(x, y)$, respectively. $\varphi(\Delta x,\Delta y)$ is the Gaussian weight at the position $(x + \Delta x,y + \Delta y)$ and $\varphi(\Delta x, \Delta y) = exp\left( -\frac{\Delta x^2 + \Delta y^2}{2\sigma^2} \right)$, where $\sigma$ is the variance of the Gaussian distribution.

To better integrate the transferred features, we adaptively integrate the $k$ sampling results according to the correlation degree of each matching pair. The reconstructed shadow-less feature $F$ can be written as: 
\begin{equation}
    F = \sum\limits_{i=1}^{k} \frac{w_i}{\begin{matrix} \sum_{i=1}^k w_i \end{matrix}} F'_i,
\end{equation}
where $F'_i$ is the $i$-th sampling result, $w_i$ is the correlation degree between the matching pair. 
Due to Gaussian sampling has a larger receptive field and takes surrounding information into consider when sampling, the Gaussian sampling in the contextual feature transfer mechanism can help to better transfer the contextual information and get pleasant results.

\subsection{Shadow Removal with Two-stage Strategy}
\begin{figure}[ht]
\begin{center}
\includegraphics[width=1\linewidth]{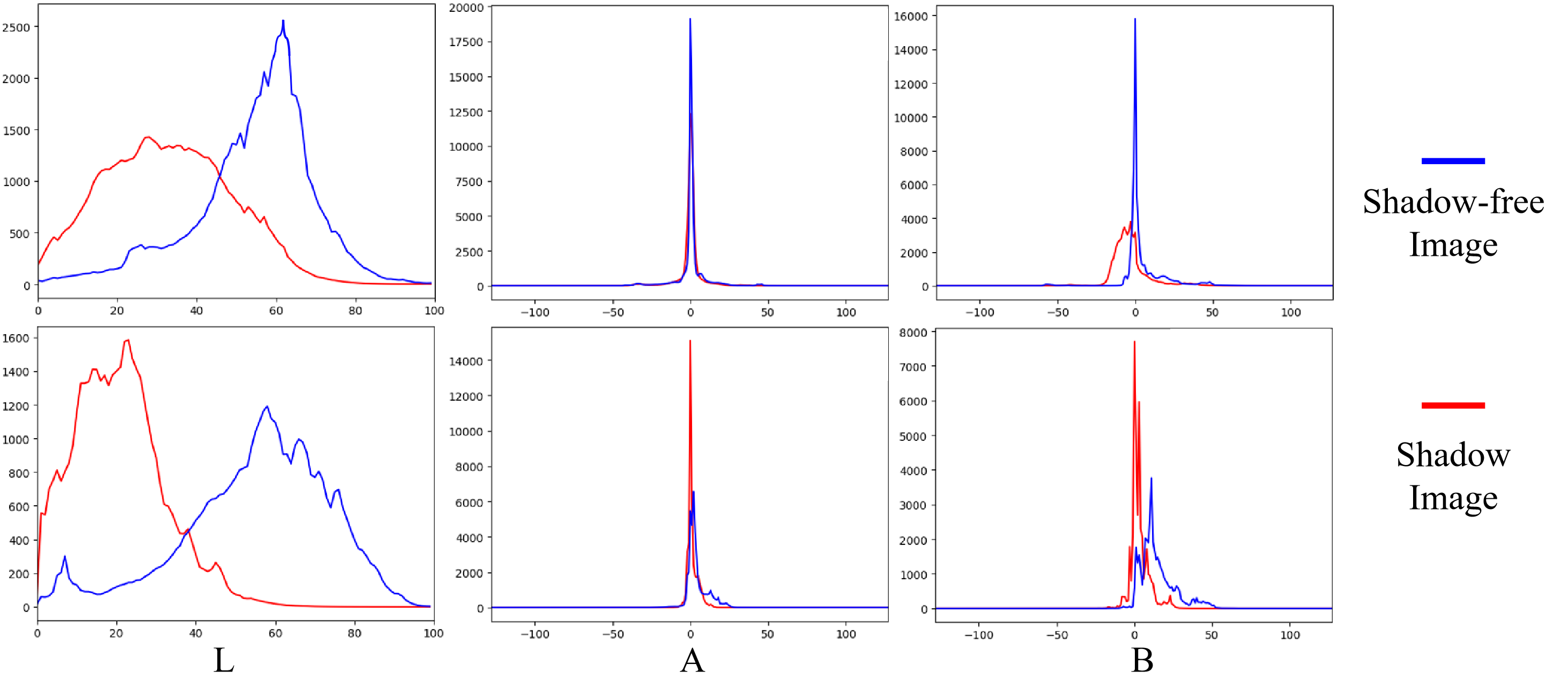}
\end{center}\vspace{-0.3cm}
   \caption{The difference between the input shadow image and the ground truth shadow-free image in shadow areas of each channel in LAB color space on the ISTD dataset~\cite{Stc-gan}(first row) and the SRD dataset~\cite{Deshadownet}(second row). 
   }
\label{fig:LAB_analyze}
\end{figure}

Our shadow-removal network CANet adopts a two-stage strategy. At Stage-I, we first use a DenseNet~\cite{Densenet}, pre-trained on ImageNet\cite{ImageNet} as the feature extractor to extract the contextual features. 
Then, with the extracted contextual features as inputs, we apply our contextual feature transfer mechanism to transfer the features from non-shadow regions to shadow regions at different scales. With a series of upsampling and residual blocks, we can remove shadows at the L and A/B channels separately. As we can see from the statistical analysis in Figure~\ref{fig:LAB_analyze}, the L channel is more sensitive than A/B channels to highlight the difference between shadow and non-shadow regions. The separation treatment can avoid the over-processing of A and B channels and the inadequate processing of the L channel, making it more propitious to feature transfer in the contextual feature transfer model.


Note that the recovered L and A/B channels are shadow-removal results using transfer operation, which maybe contain undesirable areas due to the inaccurate match in CPM module. Hence we turn to Stage-II to produce a fine shadow-removal result, which takes the recovered L and A/B channel as strong guidance information for shadow-removal result generation. At Stage-II, with the recovered L and A/B channel images and the shadow image as inputs, we use DenseUNet, an encoder-decoder structure, to predict the final shadow-removal image. 
To get a robust parametric model for shadow removal, we use a total loss $\mathcal{L}_\text{CANet}$ to train our CANet. It is defined with a removal loss $\mathcal{L}_{rem}$, a perceptual loss $\mathcal{L}_{per}$, and a gradient loss $\mathcal{L}_{grad}$, {\em i.e.},
\begin{equation}
    \mathcal{L}_\text{CANet} = \lambda_1\mathcal{L}_{rem} + \lambda_2\mathcal{L}_{per} + \lambda_3\mathcal{L}_{grad}
\end{equation}
where $\lambda_1$, $\lambda_2$, $\lambda_3$ are the hyperparameters. In this paper, we set $\lambda_1 = 1, \lambda_2 = 25, \lambda_3 = 5$.

The removal loss $\mathcal{L}_{rem}$ is the visual-consistency loss between the shadow removal result of two stages $I_{out\_1}, I_{out\_2}$ generated by our CANet and the corresponding ground-truth $I_{gt}$, {\em i.e.},
\begin{equation}
    \mathcal{L}_{rem} = \left\| I_{gt} - I_{out} \right \|_{ 2 }.
\end{equation}

The perceptual loss $\mathcal{L}_{per}$ is perceptual-consistency loss which aims to preserve the structure of image and is defined as: 
\begin{equation}
    \mathcal{L}_{per} = \left\| VGG(I_{gt}) - VGG(I_{out}) \right \|_{ 1 },
\end{equation}
where $VGG(\cdot)$ is the feature extractor from the VGG19 model.

The gradient loss $\mathcal{L}_{grad}$ is used to encourage the result to be smooth and is defined as:
\begin{equation}
    \mathcal{L}_{grad} = \left\| \nabla I_{gt} - \nabla I_{out} \right \|_{ 1 },
\end{equation}
where $\nabla$ is the gradient of the image at pixel-level.

\section{Experiments}
\textbf{Benchmark datasets.} We conduct various experiments on two shadow removal benchmark datasets to verify the effectiveness of our CANet. One is ISTD dataset~\cite{Stc-gan}, which includes 1330 training triples of shadow image, shadow mask and shadow-free image and 540 testing triplets. The other is the SRD dataset~\cite{Deshadownet} with 2680 training pairs of shadow and shadow-free images and 408 testing pairs. 

\textbf{Implementation details.}
Our proposed method is implemented in PyTorch on two GPUs (NVIDIA GeForce 2080Ti) with the input size of the image as 400 $ \times $ 400 and mini-batch size as 2. 
We empirically use the Adam Optimizer to optimize our network. In our experiments, we set the first momentum value, the second momentum value and weight decay are 0.9, 0.999 and $5 \times 10^{-4}$, respectively. We train our CPM module for 30 epochs and CANet for 50 epochs. The initial learning rate is set as 0.0001. 
Besides, we also partly verify our method on the Huawei MindSpore platform.

Similar to~\cite{matchnet},  we run our CPM module with two phases to avoid repeated operation on the same patch. We first extract features of all patches, then feed them respectively into two heads of our CPM module to produce the matching information of the input patch pair. Besides, to achieve a good trade-off between efficiency and accuracy, we set $k=3$ and $n=5$ in our experiments. 

\subsection{Comparison with State-of-the-arts}
We compare our CANet with eight state-of-the-art methods, {\em i.e.}, Guo~\cite{Guo_2011}, Zhang~\cite{Zhang_2015} DeshadowNet~\cite{Deshadownet}, ST-CGAN~\cite{Stc-gan},  Mask-shadowGAN~\cite{hu2019mask}, ARGAN~\cite{ARGAN}, DSC~\cite{DSC}, and RIS-GAN~\cite{RIS-GAN}.  Among these competing methods, the first two are traditional methods, while the last six are learning-based methods. Note that all of these learning-based methods try to explore the context information via a deep-learning model with a larger receptive field and use this information to remove shadows in the image. In particular, DSC~\cite{DSC} exploits the direction-aware spatial RNN, DeshadowNet~\cite{Deshadownet} uses the multi-context model to capture the spatial context information.

To ensure a fair comparison, we use the same training data with the same input size ({\em i.e.}, 400 $ \times $ 400) to train all the learning-based methods. 
We calculate the root mean square error (RMSE) in LAB color space between the generated shadow-removal images and the shadow-free ground truth image to quantitatively evaluate the performance of shadow removal.

{\noindent\bf Quantitative Evaluation}. We summarize the quantitative results on the test data of both SRD and ISTD in Table \ref{table:table1}. As we can see, all the competing learning-based methods perform worse than our proposed CANet. It can be explained by the fact that these baselines ignore the potential correlation between shadow and non-shadow regions Therefore these methods may fail in handling some complex shadow scenarios, especially when the most correlative non-shadow regions are not close to the shadow regions. 
Conversely, by explicitly capturing the useful potential contextual matching information globally, our CANet can handle the complicated case and therefore significantly improve the results of shadow removal. Table \ref{table:table1} reports the quantitative evaluation results compared with state-of-the-art methods, where we can see that our CANet outperforms the other state-of-the-art methods in shadow area, non-shadow area and whole image on the two datasets, which clearly demonstrates the effectiveness of our CANet.

\begin{table}[ht]
\centering
\caption{The quantitative comparison results of shadow removal between our method and recent methods on ISTD and SRD datasets in terms of RMSE (where S, N, A represent the shadow area, non-shadow area and whole image respectively).}
\resizebox{\linewidth}{!}{
\begin{tabular}{c|c c c|c c c}
\hline
\multirow{2}{*}{Method} & \multicolumn{3}{c|}{ISTD} & \multicolumn{3}{c}{SRD} \\
& S & N & A & S & N & A\\
\hline
Guo~\cite{Guo_2011}                 & 18.95 & 7.46 & 9.3 & 29.89 & 6.47 & 12.60 \\
Zhang~\cite{Zhang_2015}             & 13.77 & 7.17 & 8.16 & 9.50 & 6.90 & 7.24 \\
\hline
DeshadowNet~\cite{Deshadownet}      & 12.76 & 7.19 & 7.83 & 17.96 & 6.53 & 8.47 \\
ST-CGAN~\cite{Stc-gan}              & 10.33 & 6.93 & 7.47 & 12.65 & 6.37 & 7.83 \\
Mask-shadowGAN~\cite{hu2019mask}    & 10.35 & 7.03 & 7.61 & 10.32 & 6.83 & 7.32 \\
ARGAN~\cite{ARGAN}                  & 9.21 & 6.27 & 6.63 & 8.13 & 6.05 & 6.23  \\
DSC~\cite{DSC}                      & 9.22 & 6.39 & 6.67 & 8.22 & 6.01 & 6.21 \\
RIS-GAN~\cite{DSC}                  & 9.15 & 6.31 & 6.62 & 8.09 & 6.02 & 6.17 \\
\hline
\textbf{CANet}               & {\bf 8.86} & {\bf 6.07} & {\bf 6.15} & {\bf 7.82} & {\bf 5.88} & {\bf 5.98}\\
\hline
\end{tabular}
}
\label{table:table1}
\end{table}

\begin{figure*}[ht]
\vspace{-0.5cm}
\centering
\includegraphics[width=1.86cm]{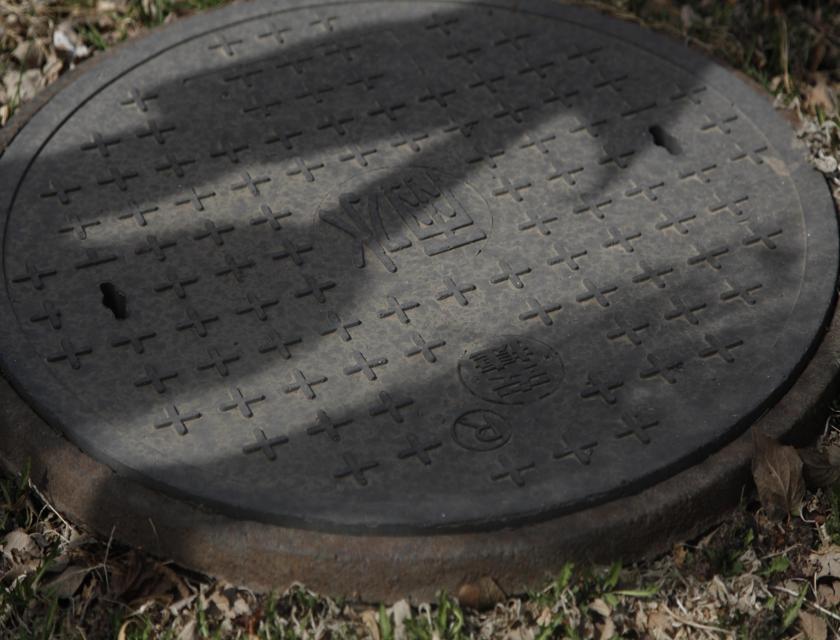}
\includegraphics[width=1.86cm]{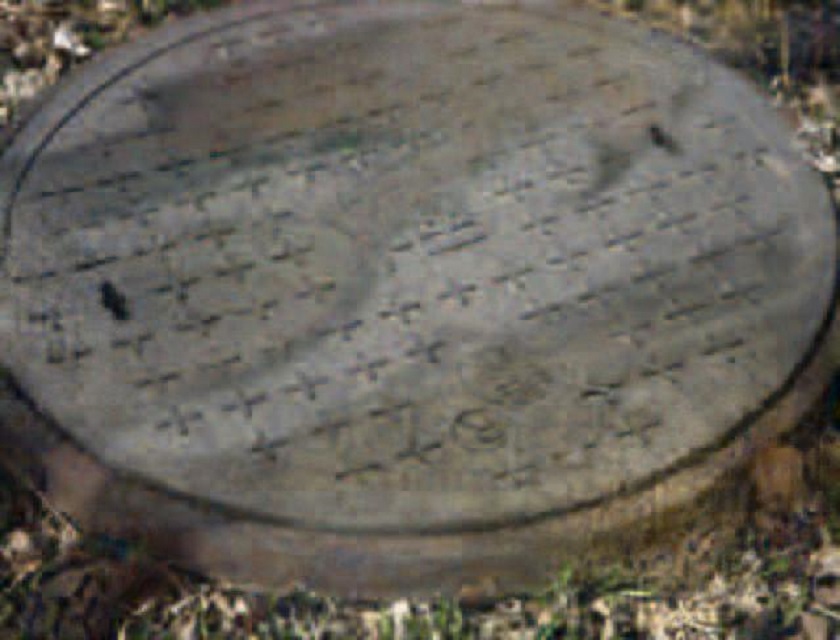}
\includegraphics[width=1.86cm]{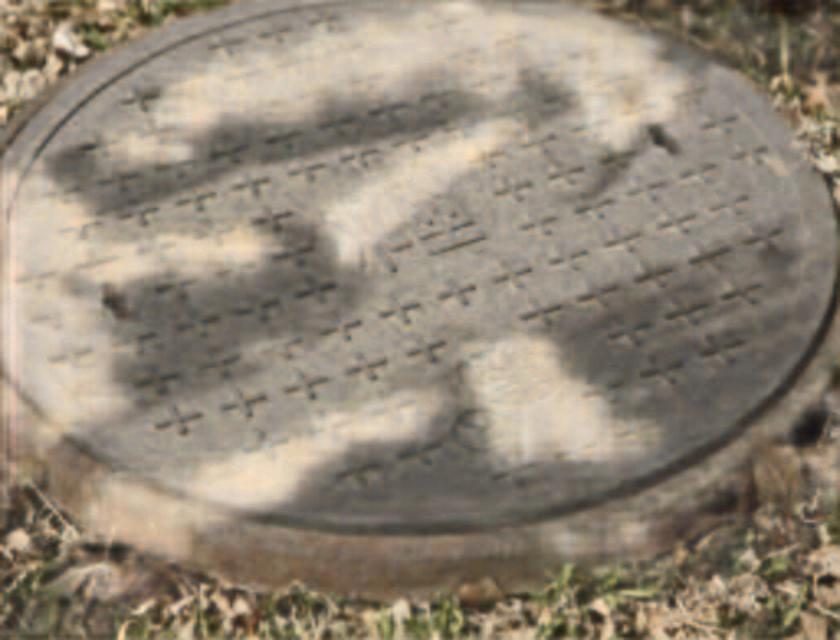}
\includegraphics[width=1.86cm]{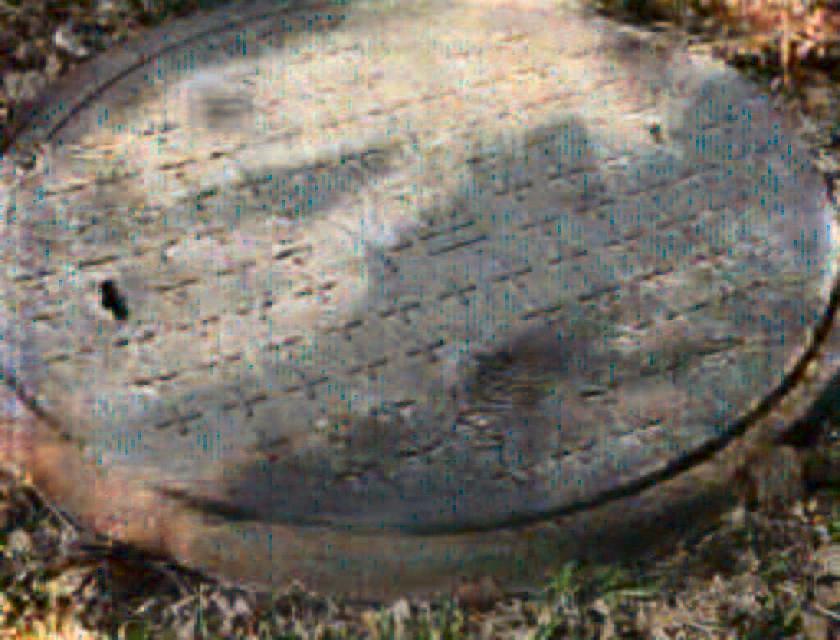}
\includegraphics[width=1.86cm]{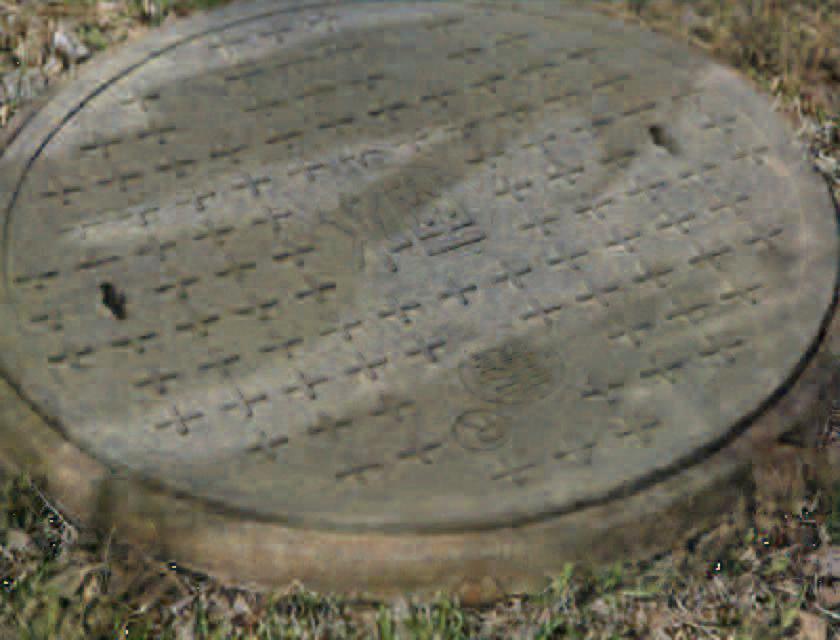}
\includegraphics[width=1.86cm]{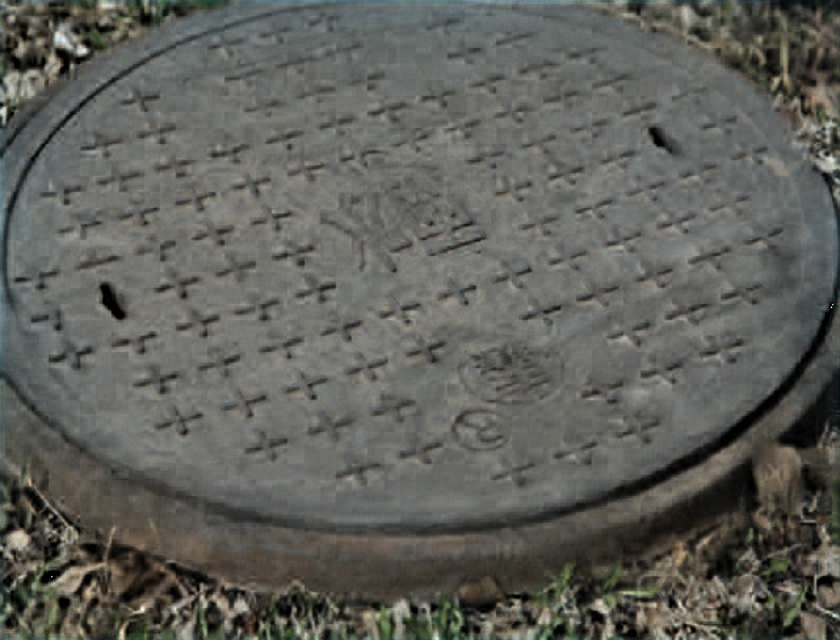}
\includegraphics[width=1.86cm]{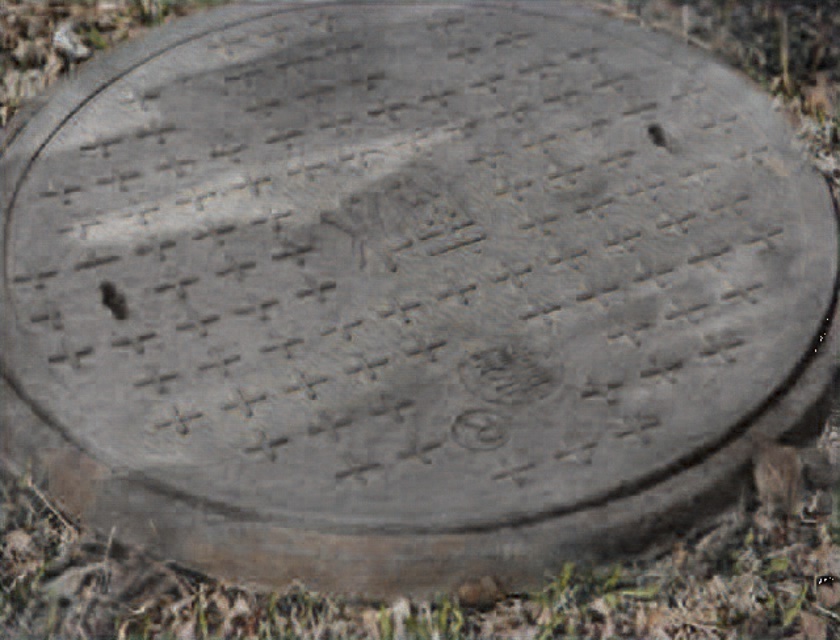}
\includegraphics[width=1.86cm]{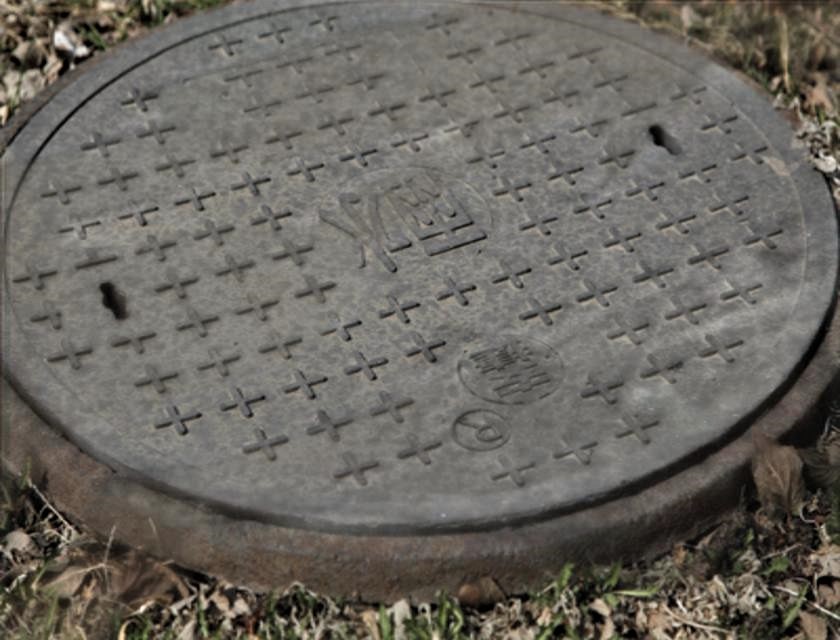}
\includegraphics[width=1.86cm]{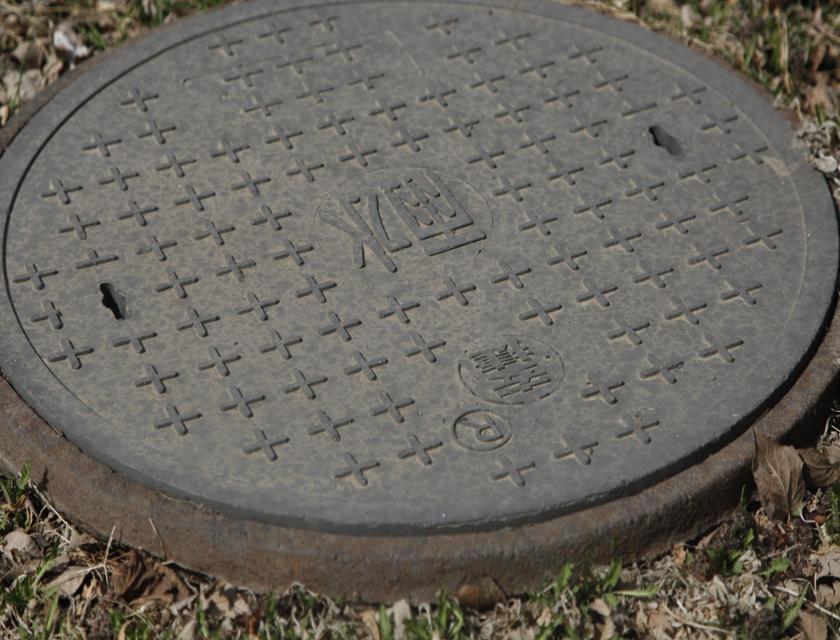}\\
\includegraphics[width=1.86cm]{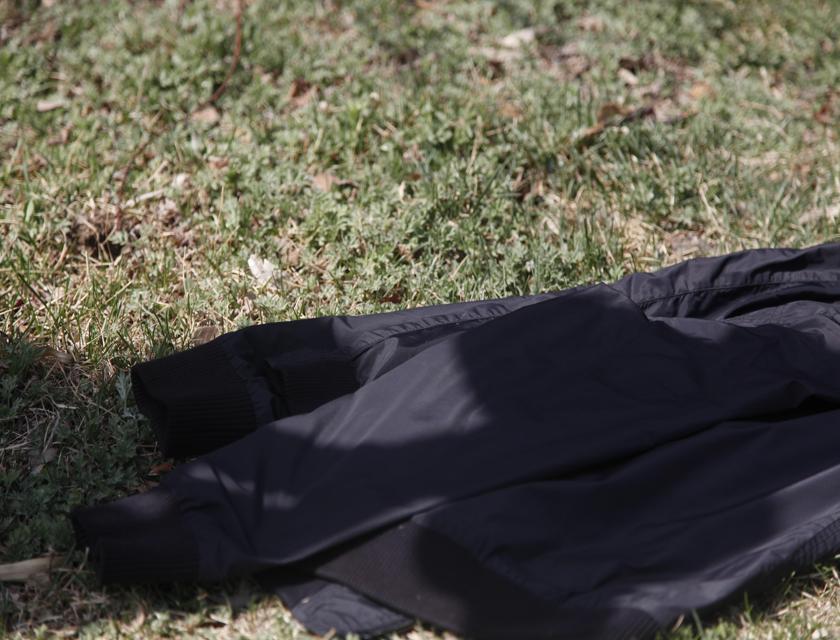}
\includegraphics[width=1.86cm]{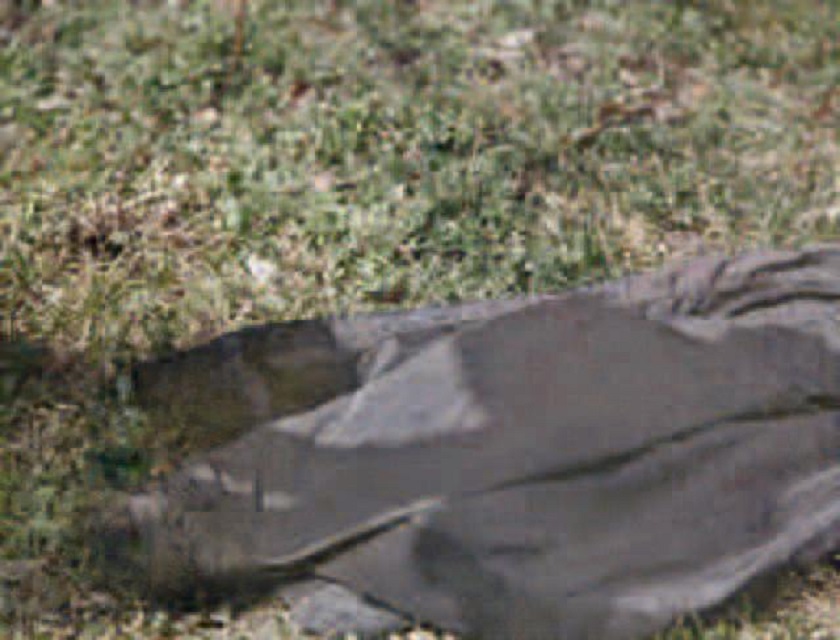}
\includegraphics[width=1.86cm]{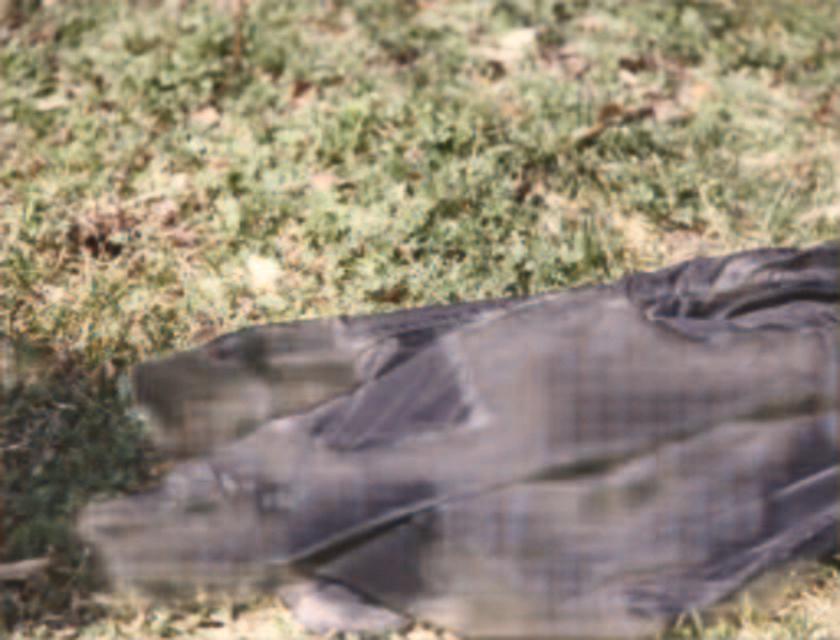}
\includegraphics[width=1.86cm]{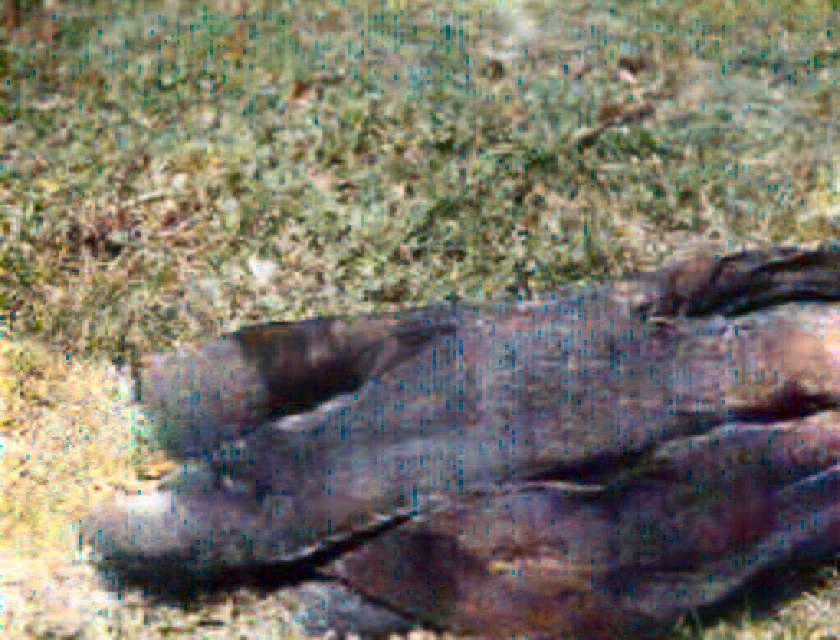}
\includegraphics[width=1.86cm]{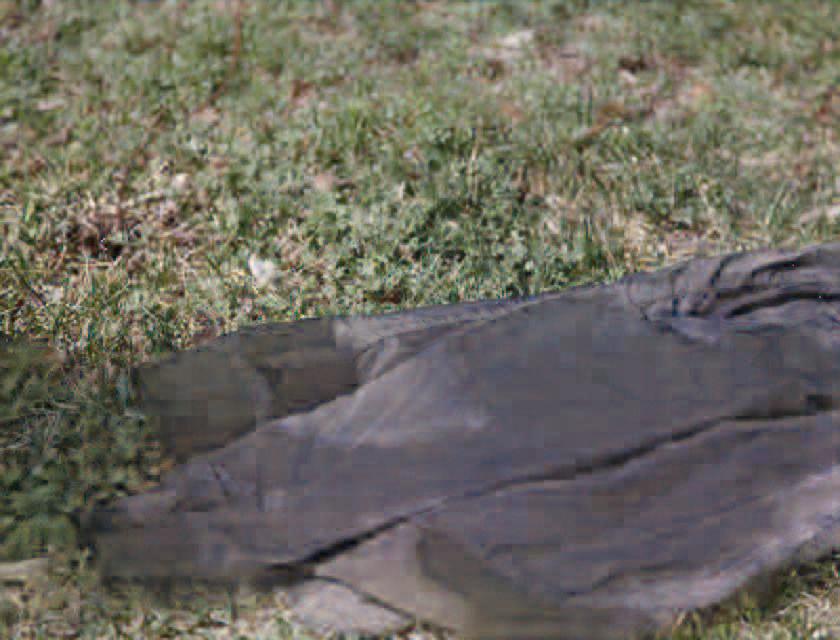}
\includegraphics[width=1.86cm]{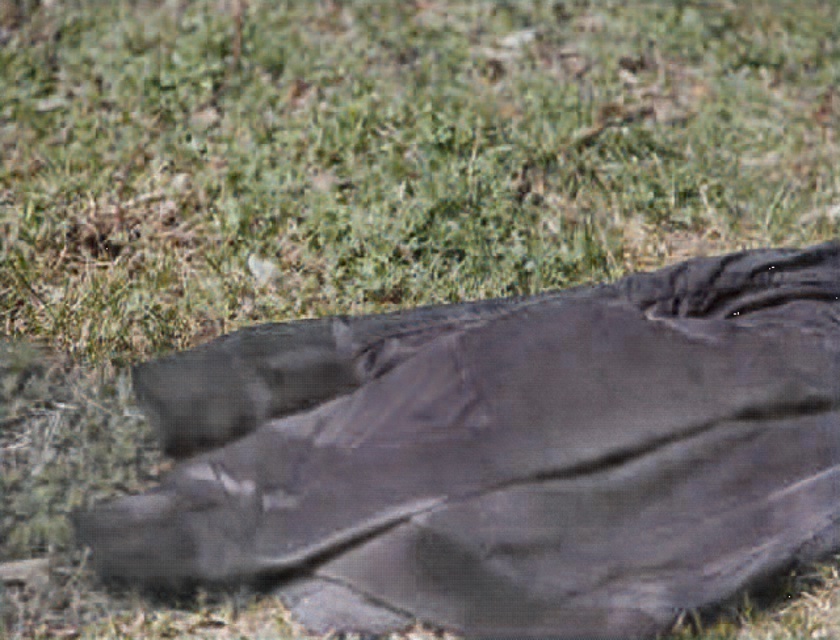}
\includegraphics[width=1.86cm]{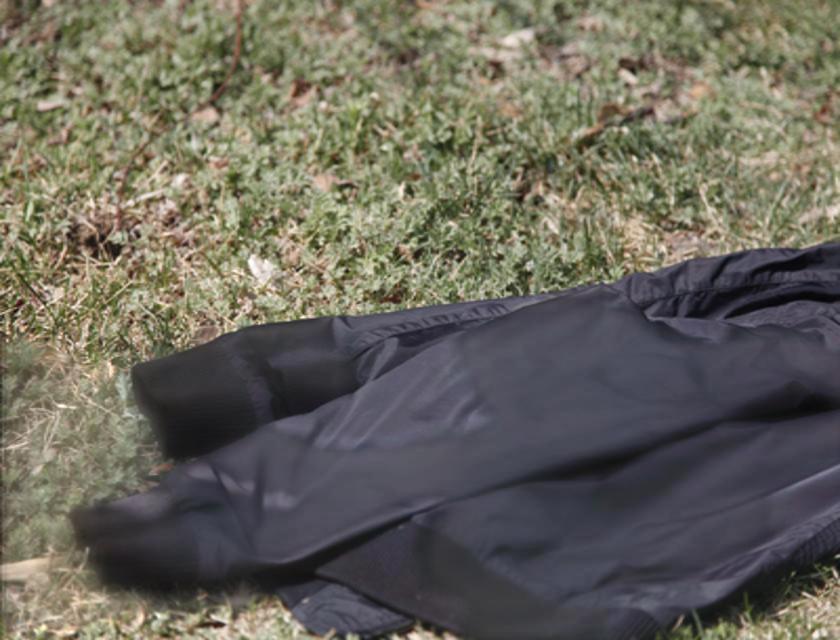}
\includegraphics[width=1.86cm]{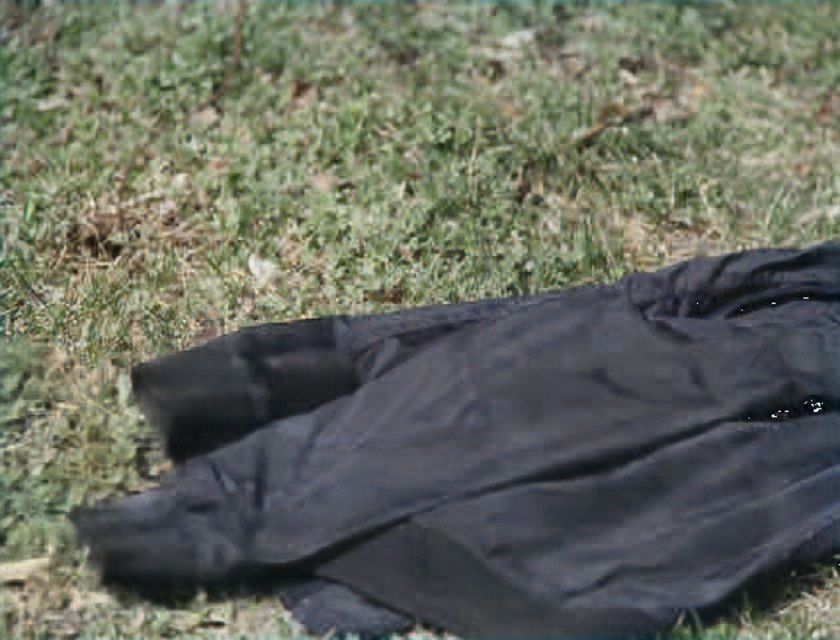}
\includegraphics[width=1.86cm]{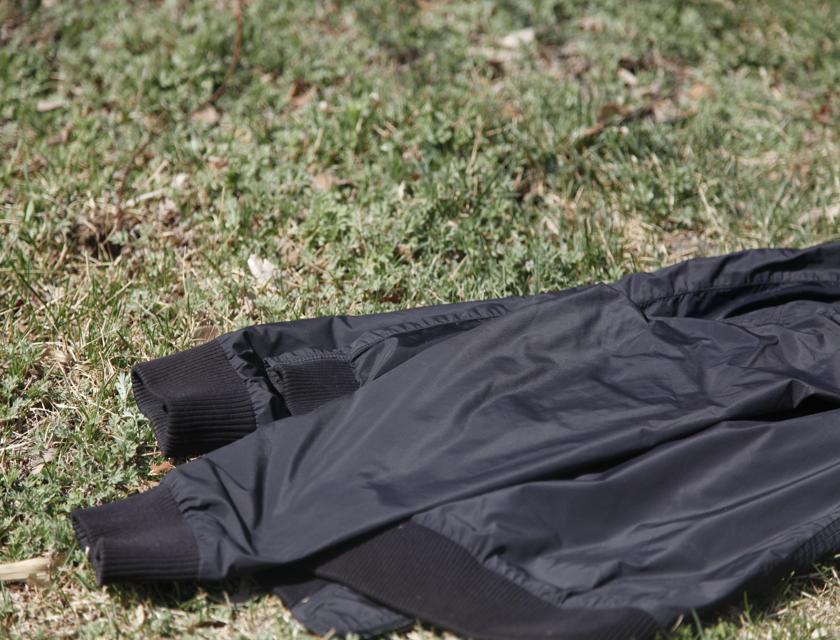}\\
\vspace{-0.15cm}
\subfigure[]{\includegraphics[width=1.86cm]{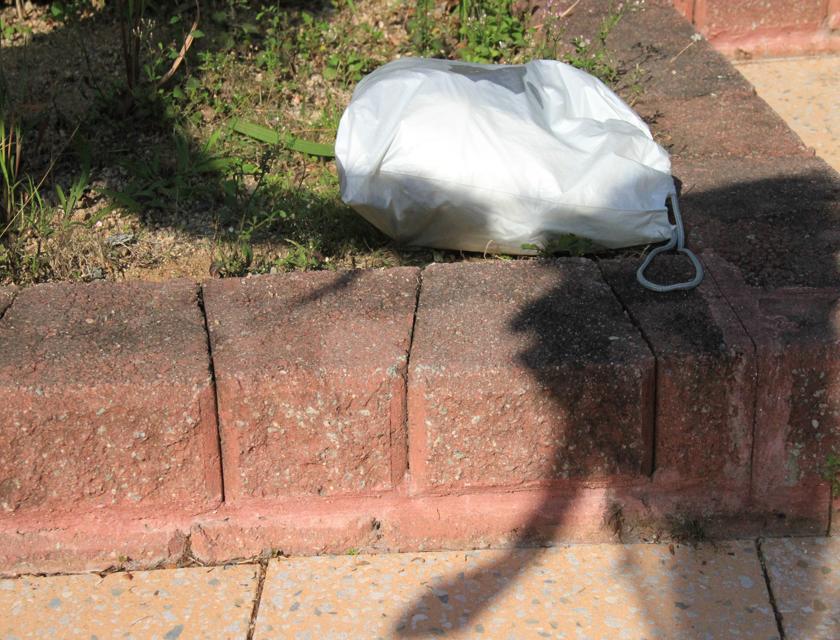}}
\subfigure[]{\includegraphics[width=1.86cm]{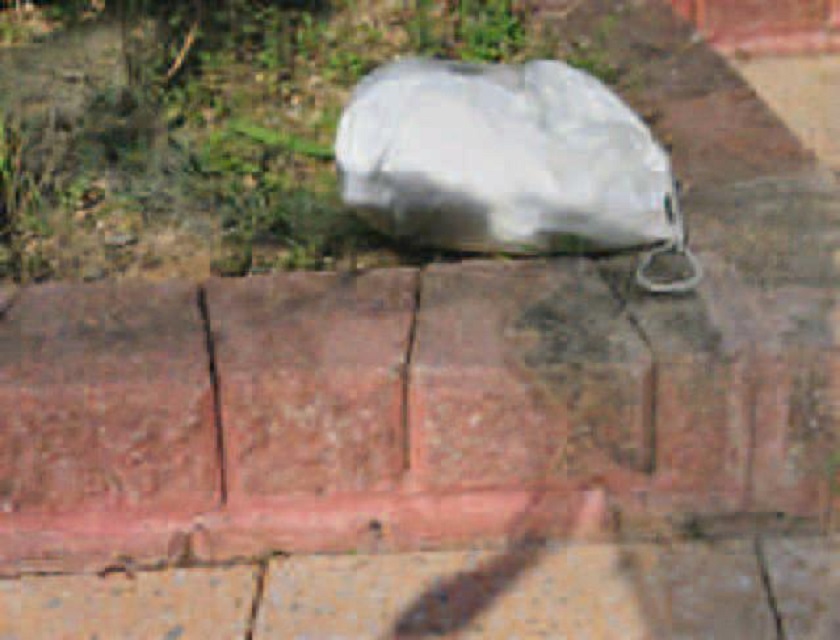}}
\subfigure[]{\includegraphics[width=1.86cm]{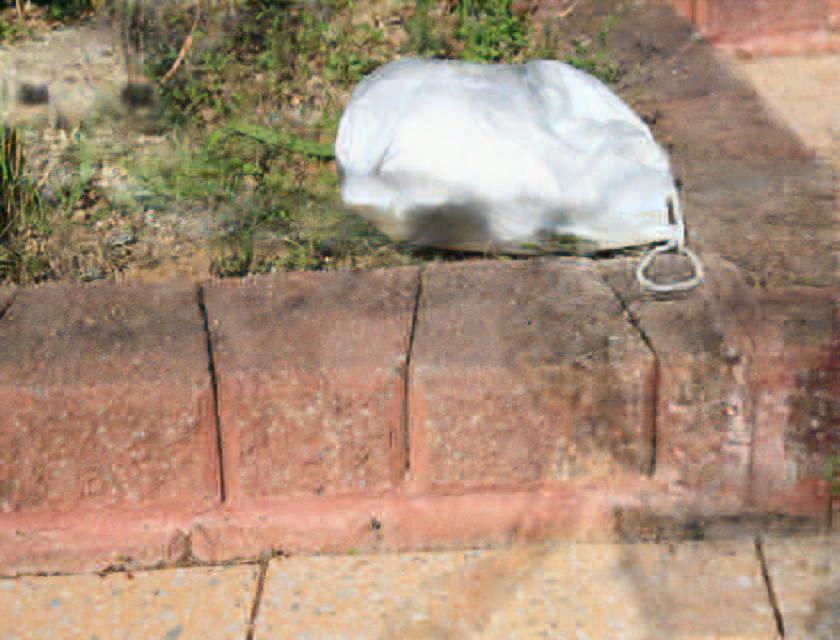}}
\subfigure[]{\includegraphics[width=1.86cm]{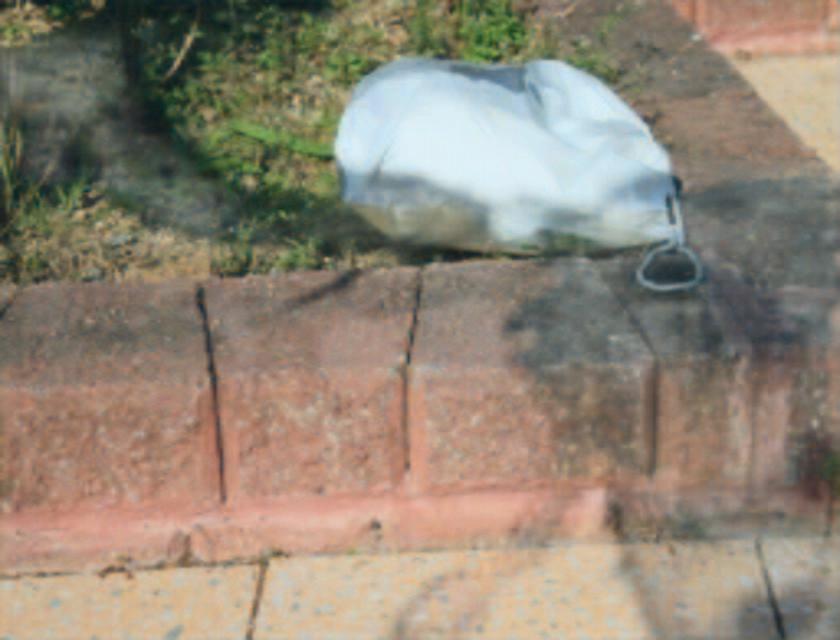}}
\subfigure[]{\includegraphics[width=1.86cm]{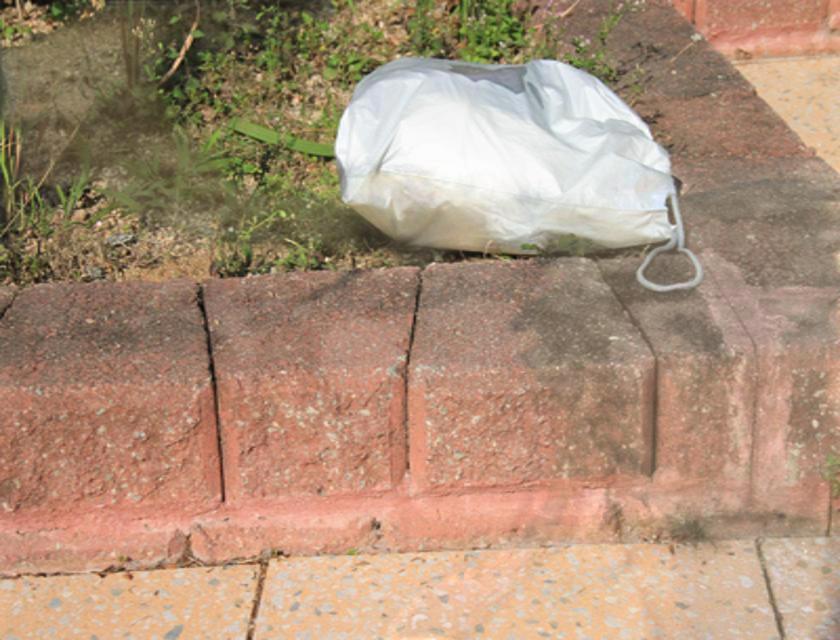}}
\subfigure[]{\includegraphics[width=1.86cm]{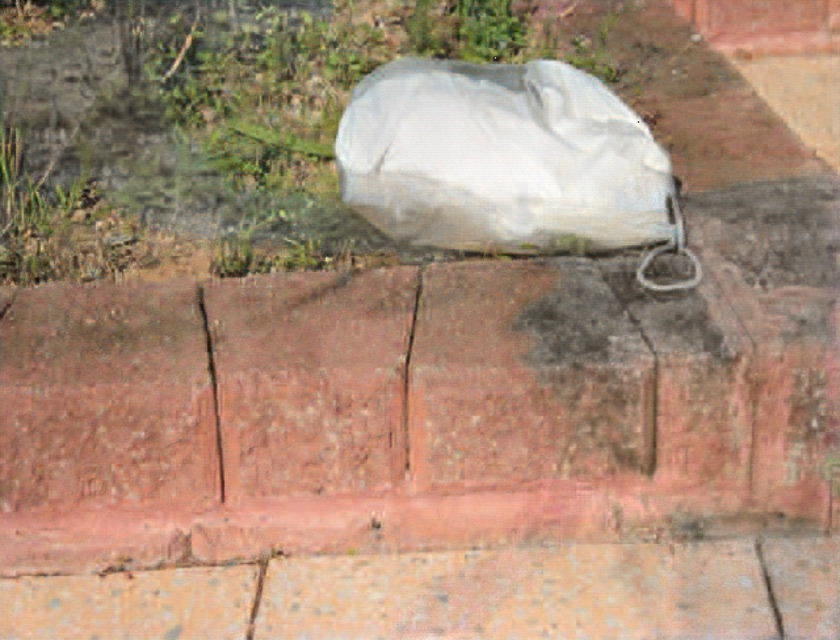}}
\subfigure[]{\includegraphics[width=1.86cm]{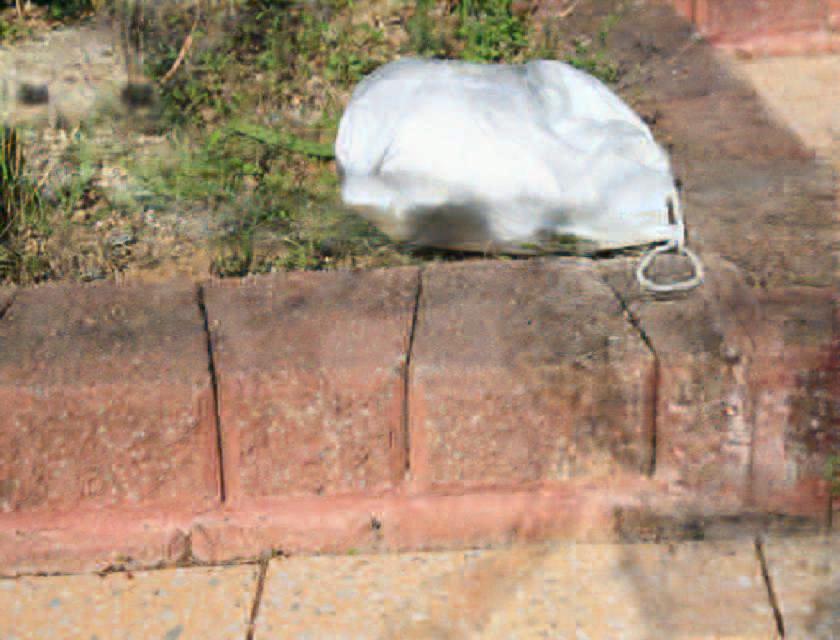}}
\subfigure[]{\includegraphics[width=1.86cm]{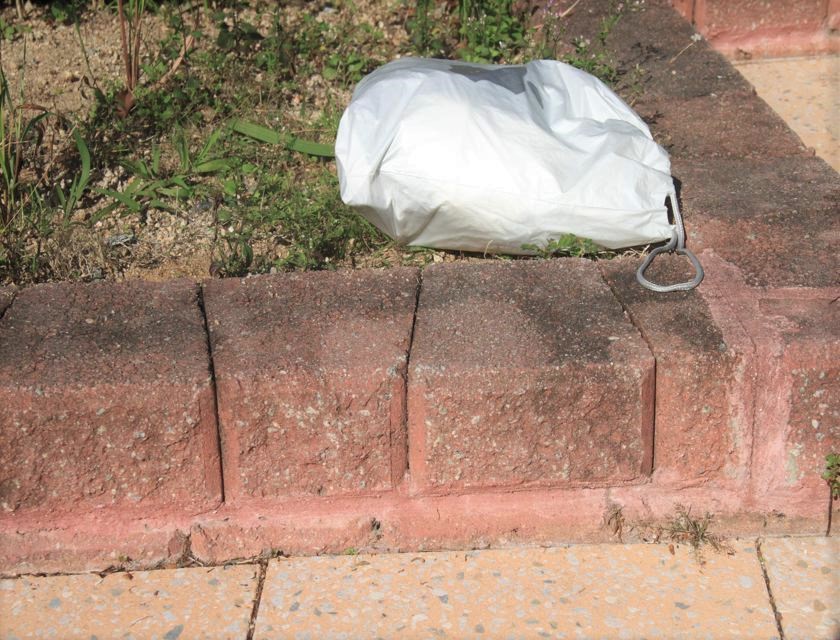}}
\subfigure[]{\includegraphics[width=1.86cm]{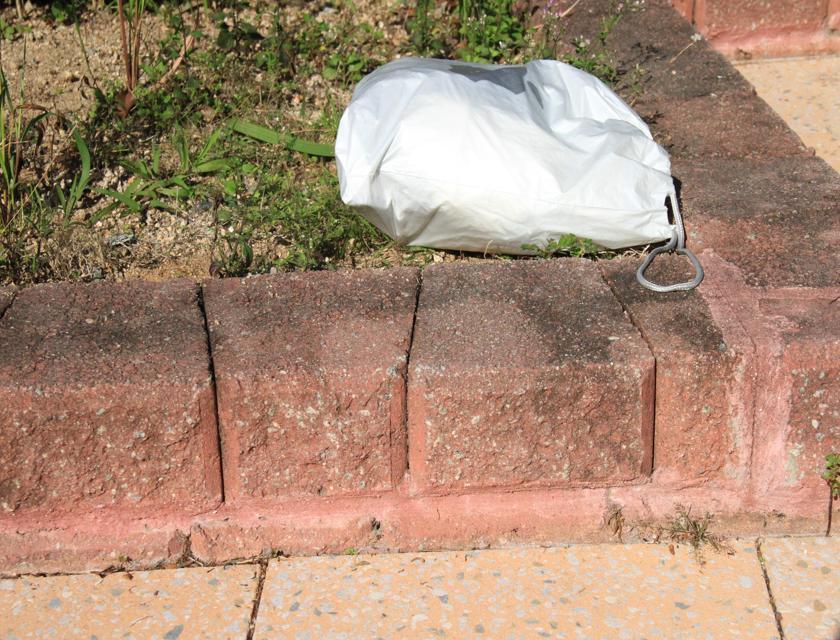}}
   \caption{Shadow removal results. From left to right are: (a)input images; shadow removal results of (b) Guo, (c) Zhang, (d) ST-CGAN, (e) DSC, (f) ARGAN, (g) RIS-GAN, (h) our CANet; and (i) the corresponding shadow-free ground truth images.
  }
\label{fig:compare}
\end{figure*}

\begin{figure*}[ht]

\centering

\includegraphics[width=2.1cm]{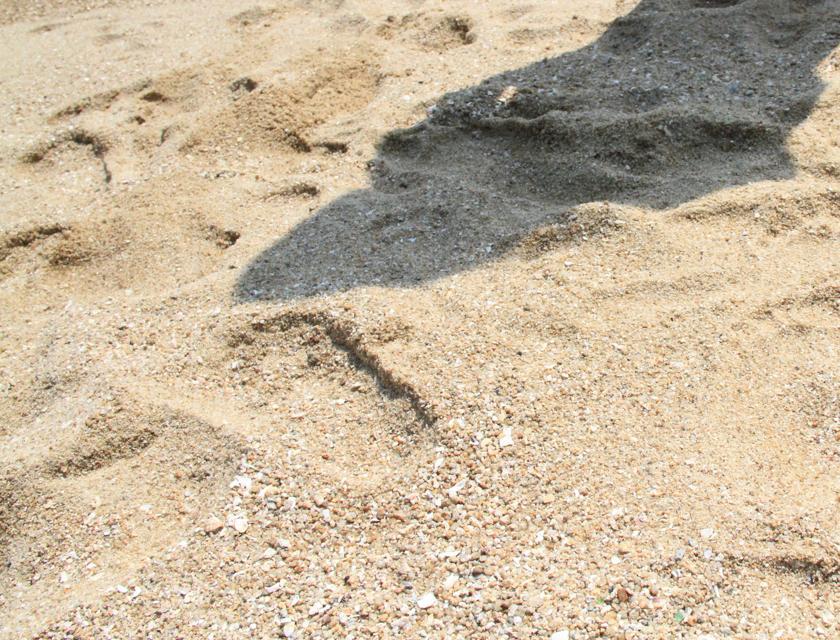}
\includegraphics[width=2.1cm]{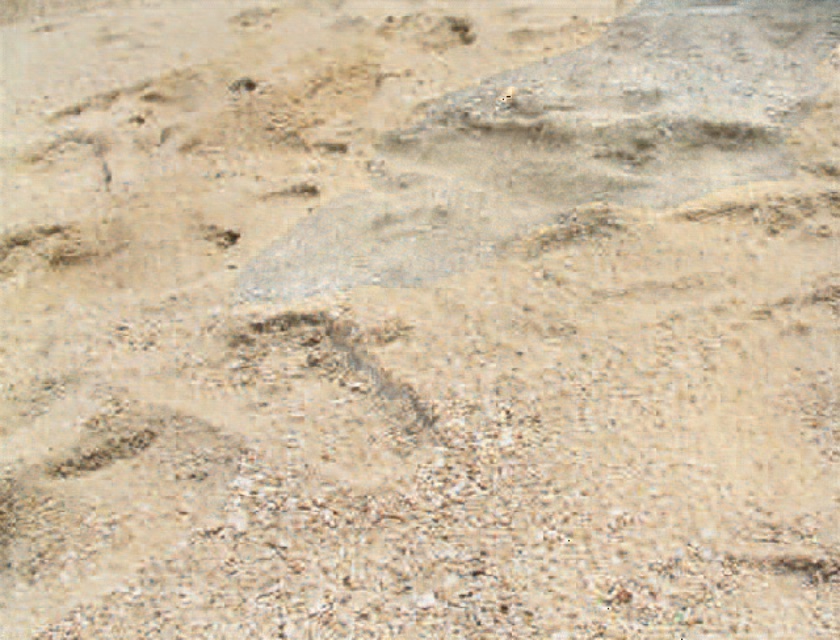}
\includegraphics[width=2.1cm]{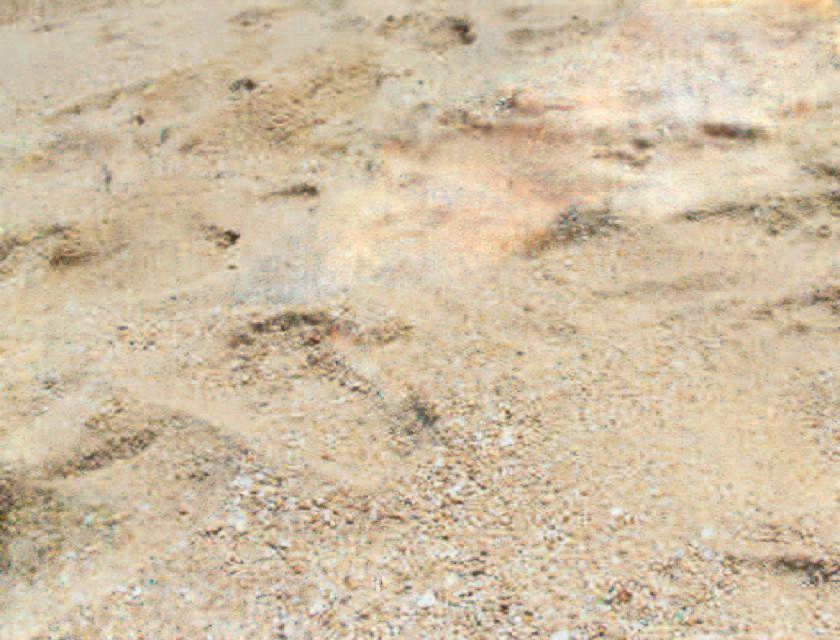}
\includegraphics[width=2.1cm]{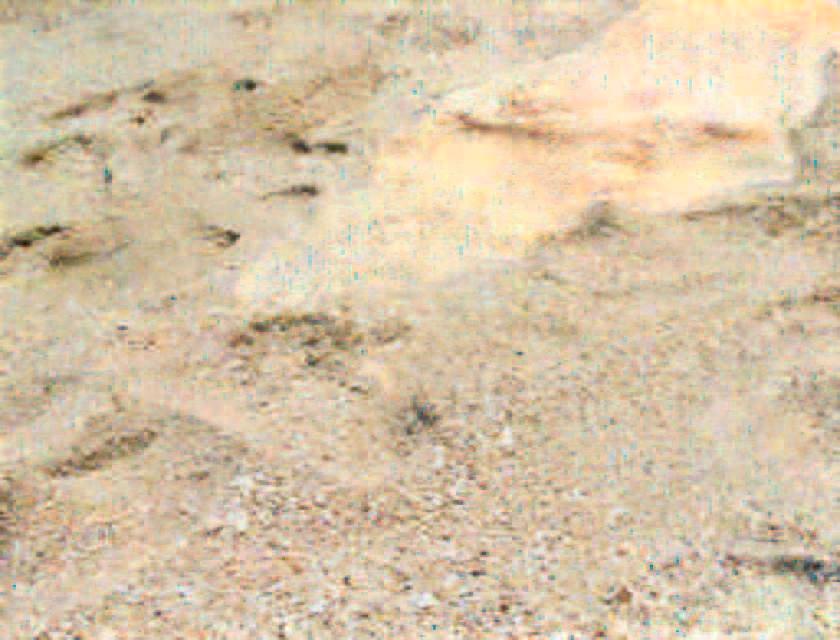}
\includegraphics[width=2.1cm]{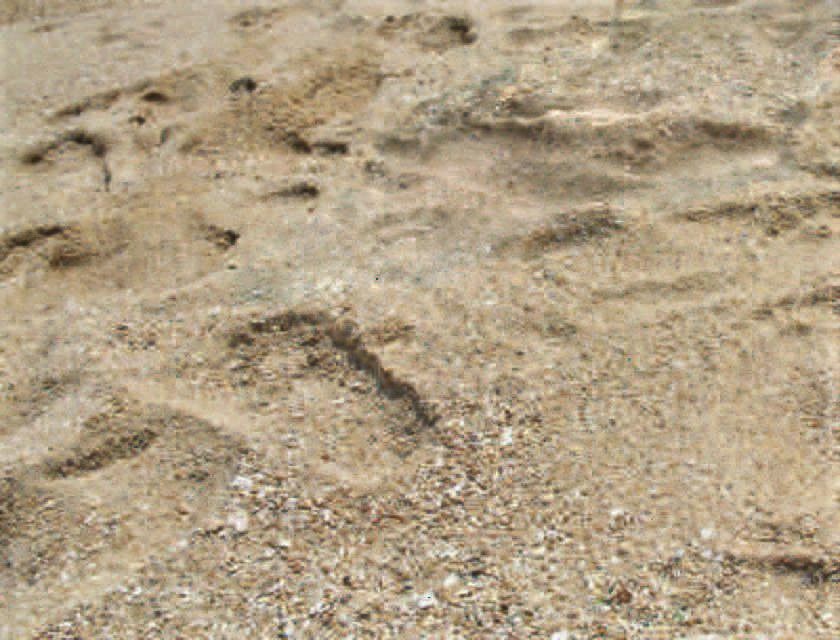}
\includegraphics[width=2.1cm]{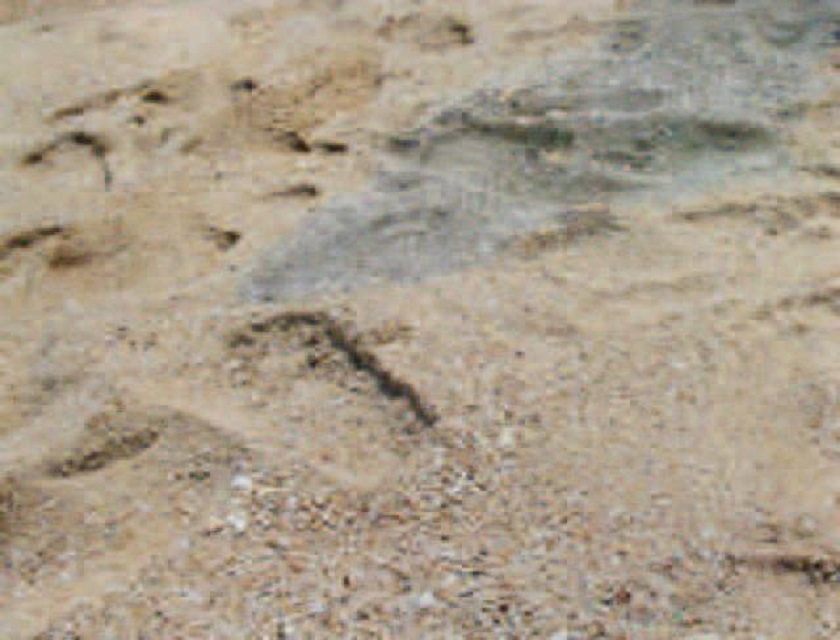}
\includegraphics[width=2.1cm]{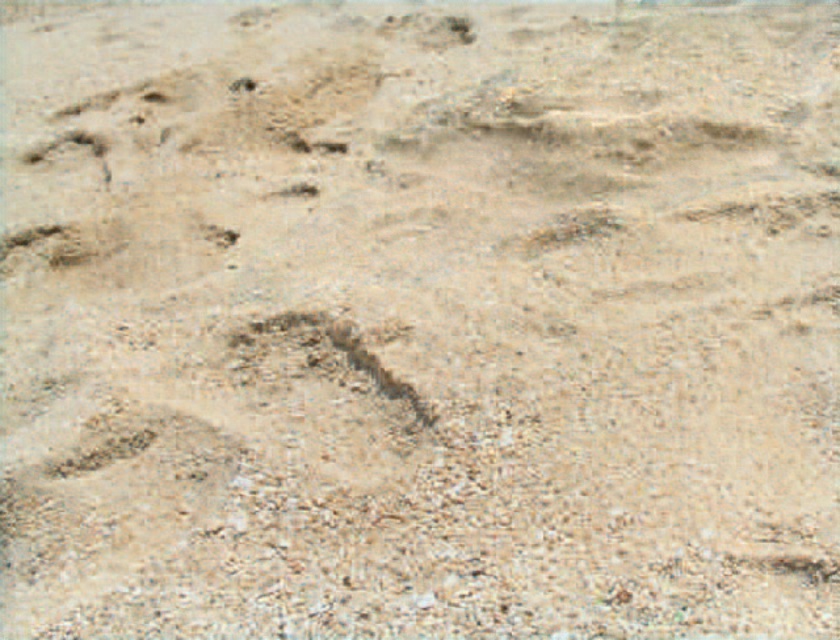}
\includegraphics[width=2.1cm]{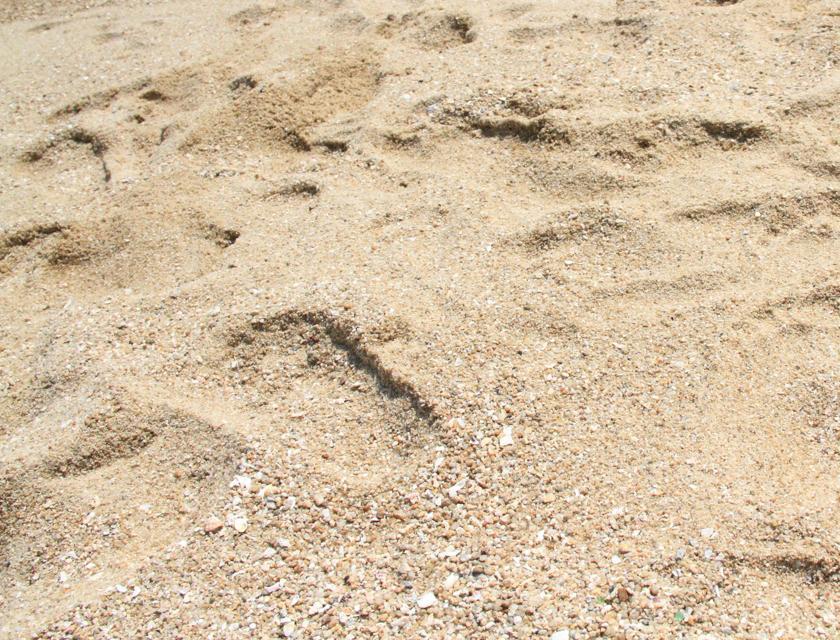}\\
\vspace{-2pt}
\subfigure[]{\includegraphics[width=2.1cm]{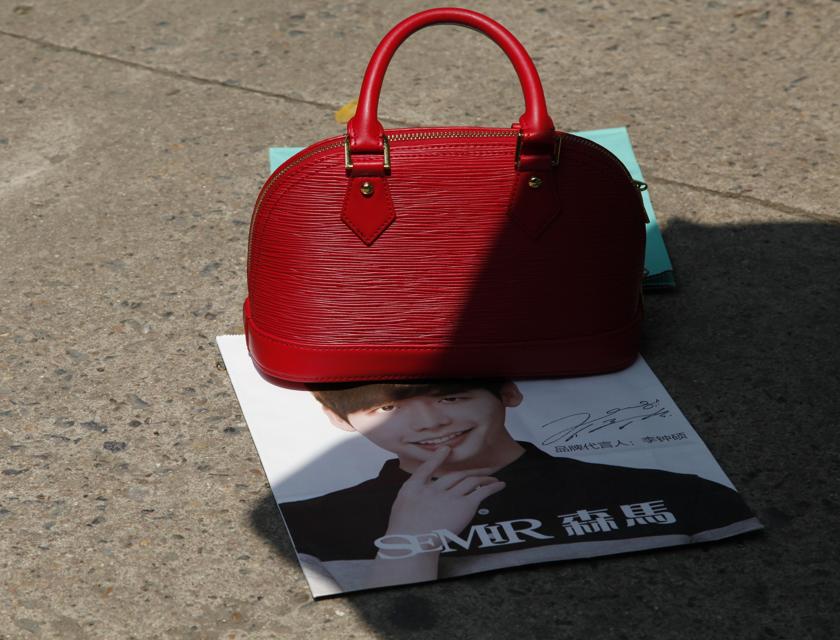}}
\subfigure[]{\includegraphics[width=2.1cm]{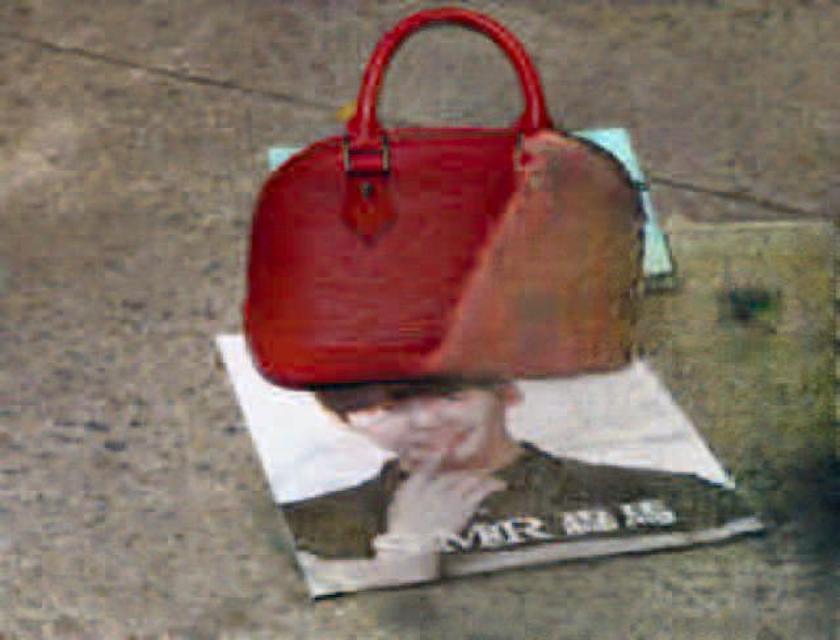}}
\subfigure[]{\includegraphics[width=2.1cm]{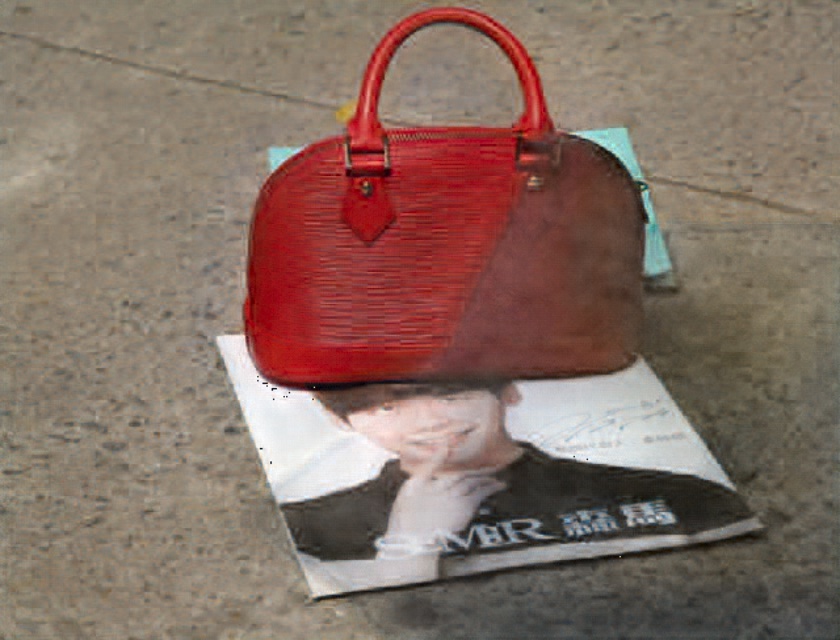}}
\subfigure[]{\includegraphics[width=2.1cm]{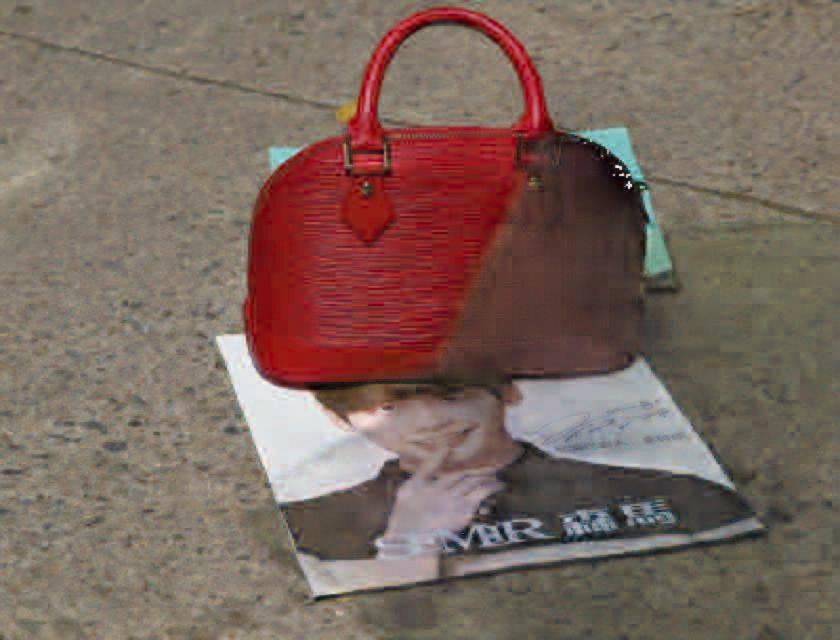}}
\subfigure[]{\includegraphics[width=2.1cm]{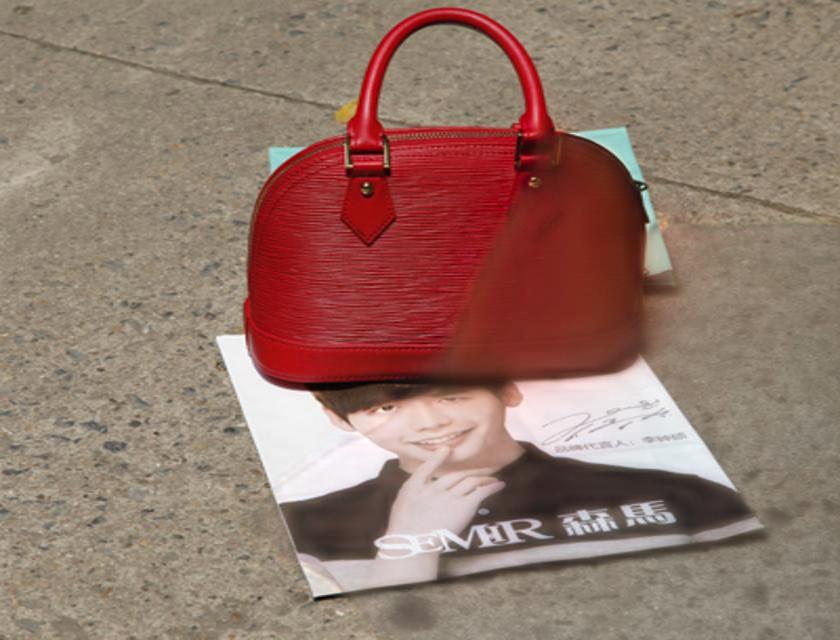}}
\subfigure[]{\includegraphics[width=2.1cm]{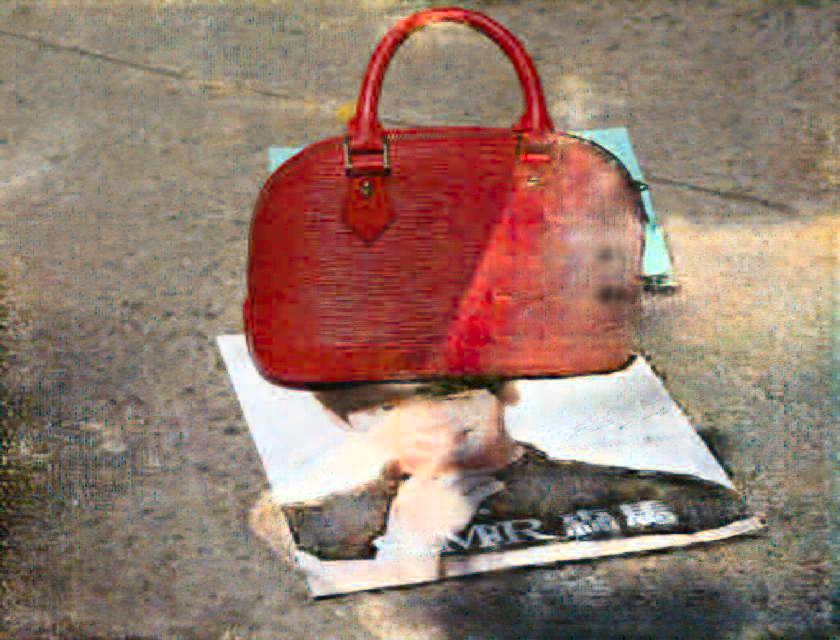}}
\subfigure[]{\includegraphics[width=2.1cm]{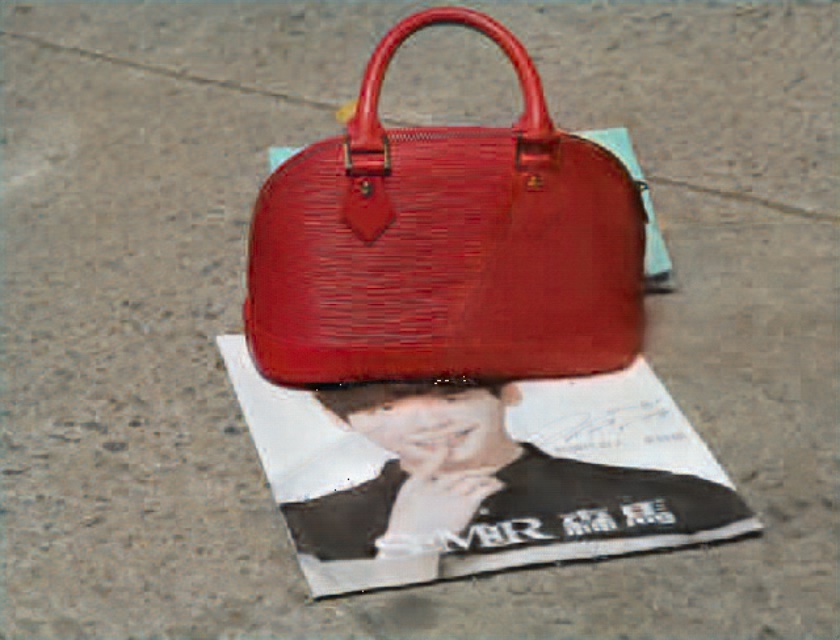}}
\subfigure[]{\includegraphics[width=2.1cm]{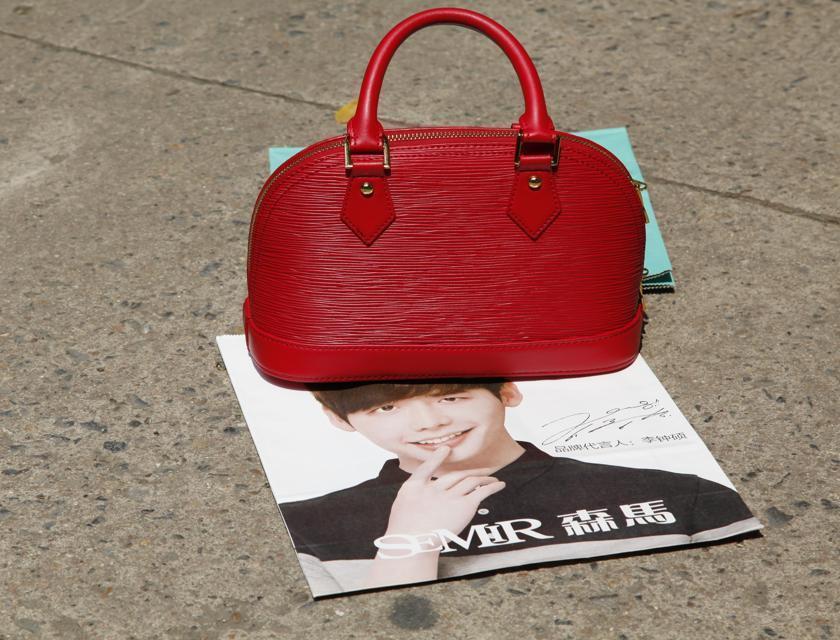}}
  \caption{Shadow removal results. From left to right are: (a)input images; shadow removal results of our (b) ``CANet w/ TM", (c) ``CANet w/ MNet", (d) ``CANet w/o CFT", (e) ``CANet w/ DRCF", (f) DenseUNet, (g) our proposed CANet, and (h) the corresponding shadow-free ground truth images.
  }
\label{fig:ablation_study}
\vspace{-0.3cm}
\end{figure*}

{\noindent\bf Qualitative Evaluation}. 
We also provide visual comparison results in Figure~\ref{fig:compare}. 
For traditional methods, due to the local information transfer, Guo~\cite{Guo_2011} can not completely remove shadows. Its results contain some artifacts, as shown in Figure~\ref{fig:compare}(b). Zhang~\cite{Zhang_2015} also cannot handle the illumination changes at the shadow boundary well, as shown in Figure~\ref{fig:compare}(c). This is due to the contextual information of the image being ignored when processing at the pixel level, causing inaccurate or wrong matching of lit block.  On the contrary, with the well-designed CPM module and CFT mechanism, our CANet can better recover the illumination consistent with surroundings and solve the boundary problems such as existing artifacts, generating more realistic shadow removal results. 

Regarding the learning-based methods, although they can handle some simple scenes well, they are still far from satisfactory for shadow images with complex scenes, resulting in some unpleasing shadow removal results. To specify,  although ST-CGAN removes most of the shadows, its results still contain some artifacts, as shown in Figure~\ref{fig:compare}(d). From Figure~\ref{fig:compare}(e-g), we can observe that DSC severely distorts the colors around the shadows and both ARGAN and RIS-GAN have a certain degree of excessive shadow removal. The main reason for such poor results is that these methods ignore the potential correlation in images, without taking the difference between different color channels into consideration. In contrast, our proposed CANet captures the underlying correlation between shadow and non-shadow regions in image and therefore can effectively avoid the color distortion in the results. As seen in Figure~\ref{fig:compare}(h), our CANet method produces more realistic and promising results than the competing methods. 

\subsection{Ablation study}
To further verify the effectiveness of these components we proposed, we design a series of variants.
First, we replace our CPM module with two patch matching methods. One is ``traditional-match" which captures contextual matching set with a traditional hand-craft descriptor and Euclidean distance, and the other one is MatchNet~\cite{matchnet}. We denote these two variants as ``CANet w/ TM" and ``CANet w/ MNet", respectively. Then we design two variants for verifying the effectiveness of the proposed CFT mechanism. One is to remove the context feature transfer mechanism completely, and the other one is to replace the contextual feature directly without considering Gaussian sampling in CFT during the feature transfer. We denote these two new variants as ``CANet w/o CFT" and ``CANet w/ DRCF", respectively. Last but not least, we also take DenseUNet to conduct one-stage shadow removal directly.

For fairness, we train these variants on the same training data. The quantitative results are summarized in Table~\ref{tab:ablation_study}. From the table, we can observ that: (1) the contextual mapping information provided by the CFT mechanism can help to improve the accuracy of shadow removal results; (2) the CPM module is essential to ensure the quality of contextual matching information; (3) the proposed CFT mechanism ensure the best performance of our CANet. 

Figure~\ref{fig:ablation_study} presents some visual results for the different variants. As we can see, without considering the contextual matching information, there may be some shadow artifacts in the results, as shown in  Figure~\ref{fig:ablation_study}(d). Replacing the contextual features directly results in unsatisfied results with discontinuous illumination and color, as shown in Figure~\ref{fig:ablation_study}(e).
Besides, Figure~\ref{fig:ablation_study}(b-c) present some artifacts due to incorrect mapping information. Clearly, we can see that our CANet is the most suitable and efficient.


\begin{table}[ht]
\centering
\caption{The quantitative shadow removal results of ablation analysis on ISTD and SRD dataset in terms of RMSE. }
\resizebox{\linewidth}{!}{
\begin{tabular}{c|c c c|c c c}
\hline
\multirow{2}{*}{Method} & \multicolumn{3}{c|}{ISTD} & \multicolumn{3}{c}{SRD} \\
& S & N & A & S & N & A\\
\hline
CANet w/ TM & 9.62 & 6.33 & 6.98 & 8.44 & 6.58 & 6.89 \\
CANet w/ MNet & 9.16 & 6.20 & 6.52 & 8.17 & 6.21 & 6.35 \\
CANet w/o CFT & 10.11 & 6.88 & 7.54 & 9.28 & 6.35 & 6.96 \\
CANet w/ DRCF & 9.15 & 6.21 & 6.56 & 8.10 & 6.11 & 6.25 \\
DenseUNet & 10.22 & 7.02 & 7.58 & 10.44 & 6.71 & 7.28 \\
CANet & {\bf 8.86} & {\bf 6.07} & {\bf 6.15} & {\bf 7.82} & {\bf 5.88} & {\bf 5.98} \\
\hline
\end{tabular}
\label{tab:ablation_study}
}
\end{table}
\vspace{-1pt}
\subsection{Discussion}

\begin{figure}[ht]
\vspace{-0.3cm}
\begin{center}
\includegraphics[width=1.32cm]{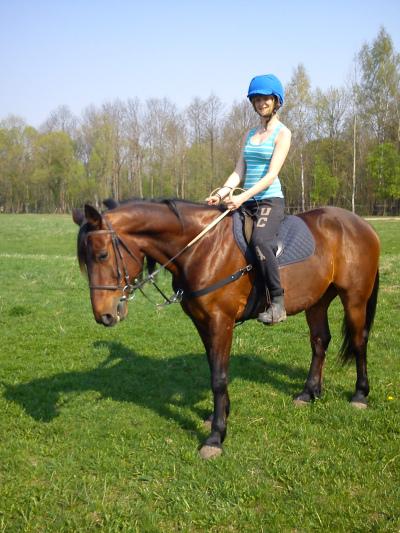}
\includegraphics[width=1.32cm]{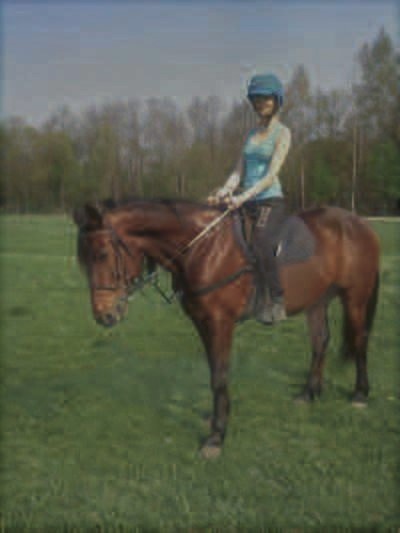}
\includegraphics[width=1.32cm]{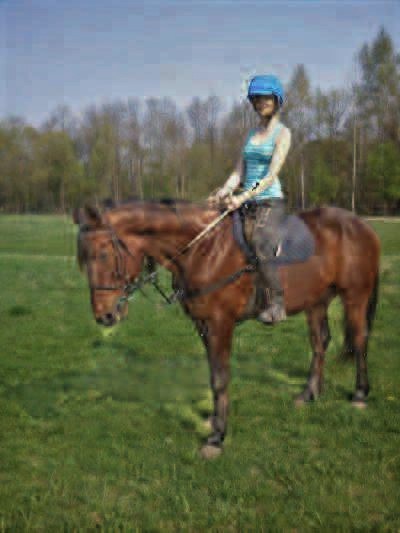}
\includegraphics[width=1.32cm]{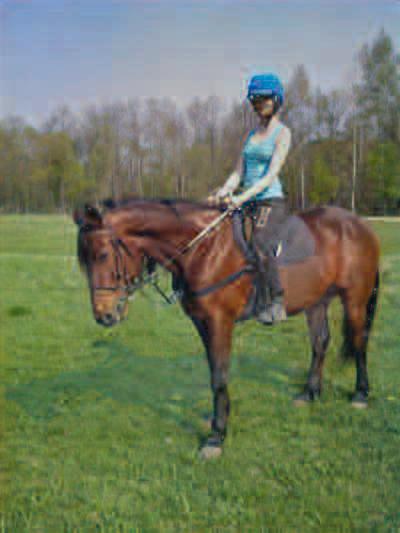}
\includegraphics[width=1.32cm]{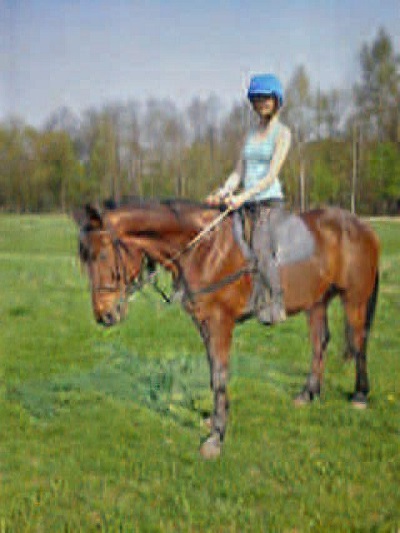}
\includegraphics[width=1.32cm]{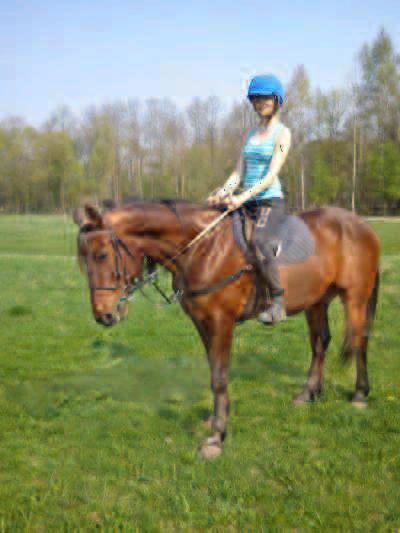}\\

\includegraphics[width=1.32cm]{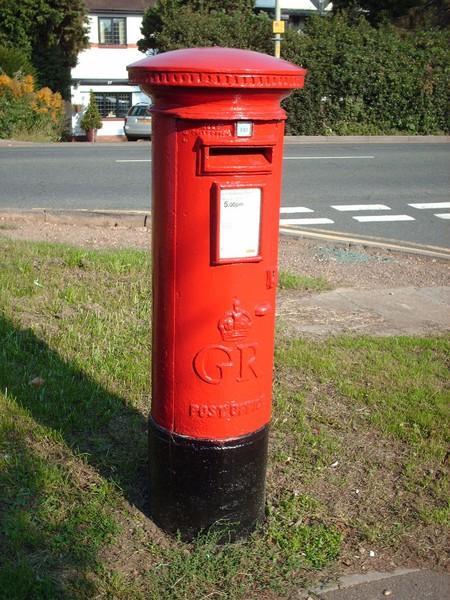}
\includegraphics[width=1.32cm]{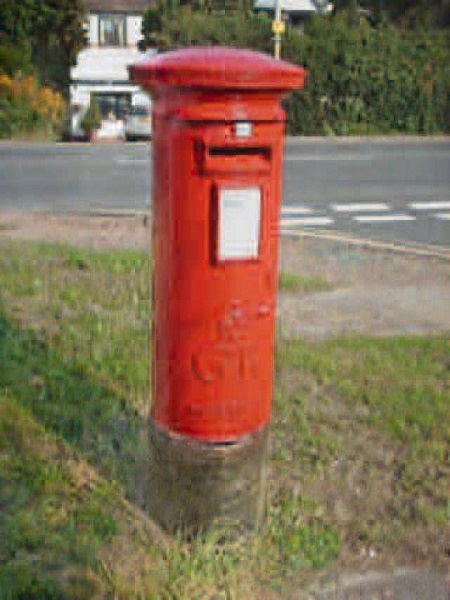}
\includegraphics[width=1.32cm]{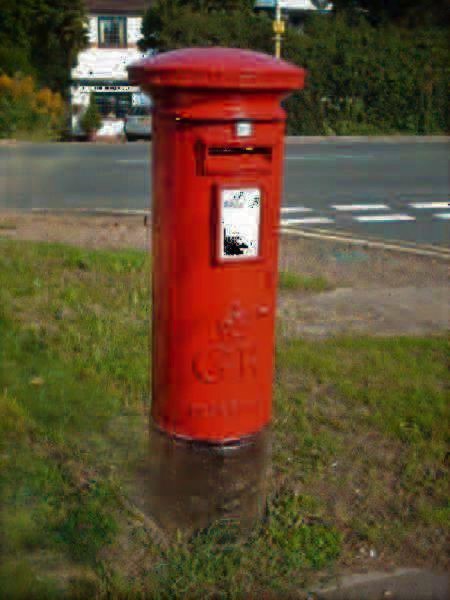}
\includegraphics[width=1.32cm]{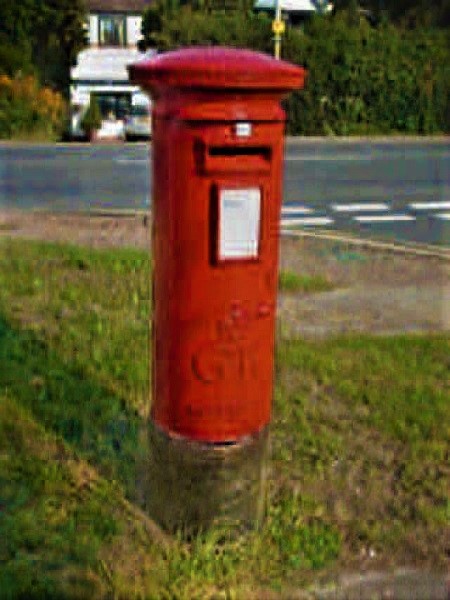}
\includegraphics[width=1.32cm]{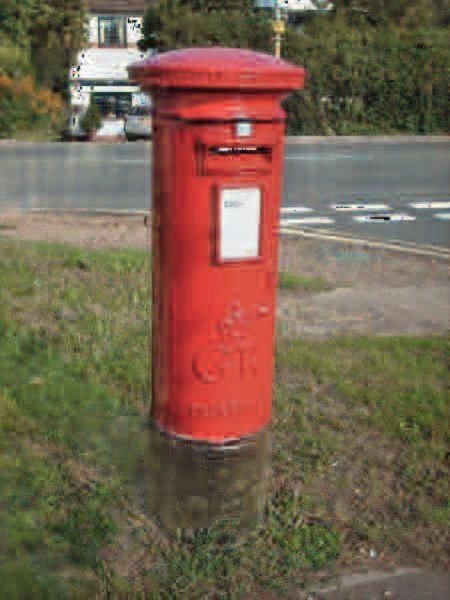}
\includegraphics[width=1.32cm]{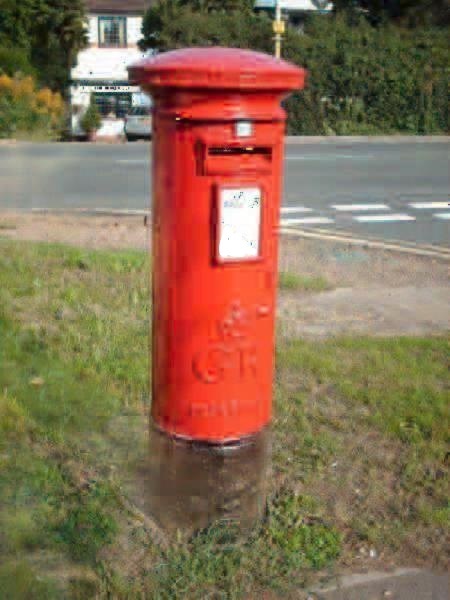}\\

\includegraphics[width=1.32cm]{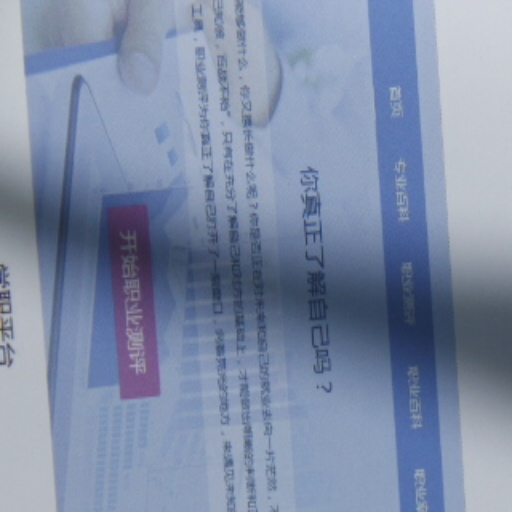}
\includegraphics[width=1.32cm]{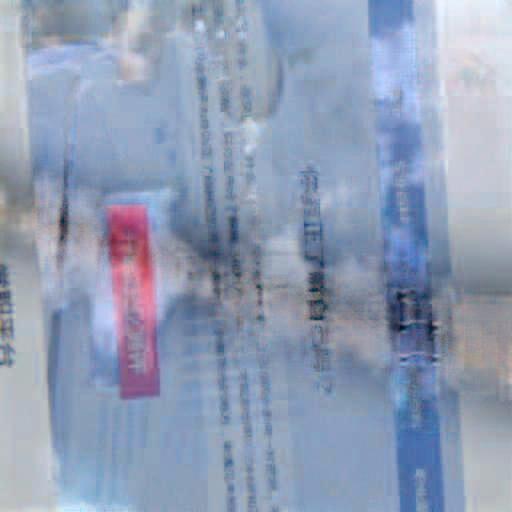}
\includegraphics[width=1.32cm]{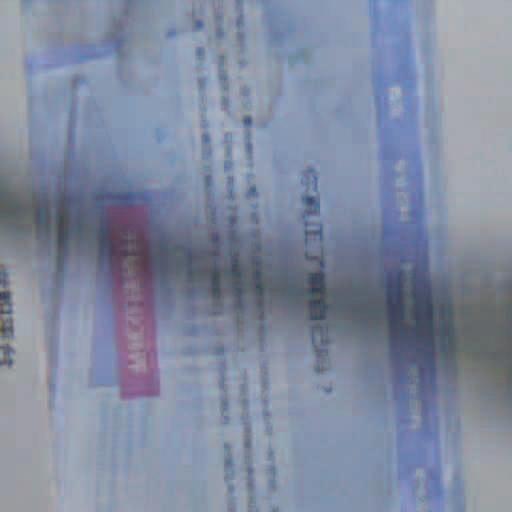}
\includegraphics[width=1.32cm]{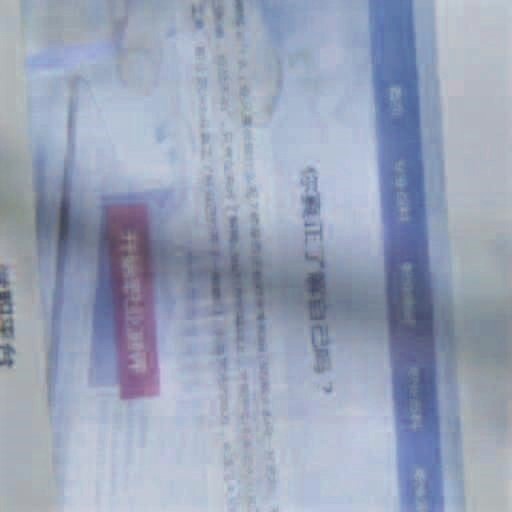}
\includegraphics[width=1.32cm]{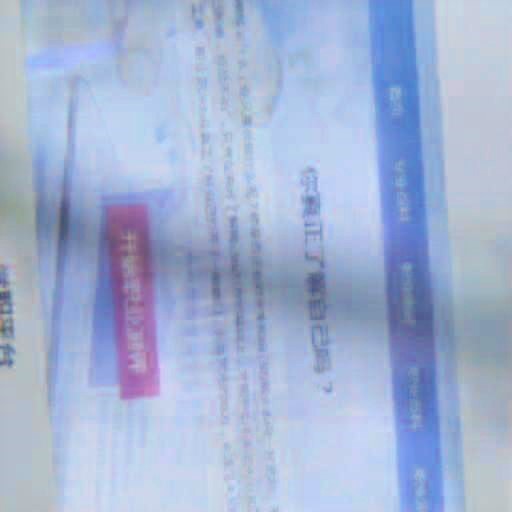}
\includegraphics[width=1.32cm]{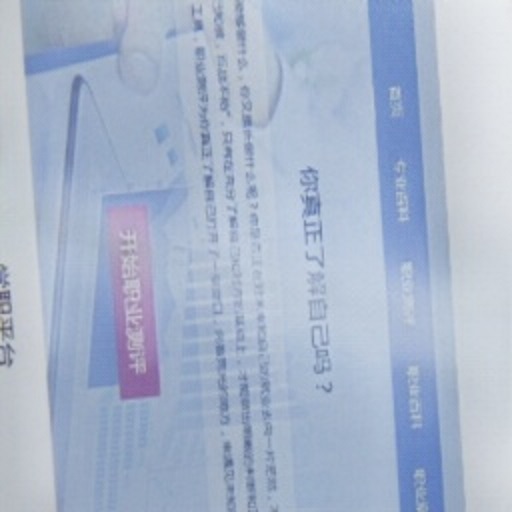}\\

\includegraphics[width=1.32cm]{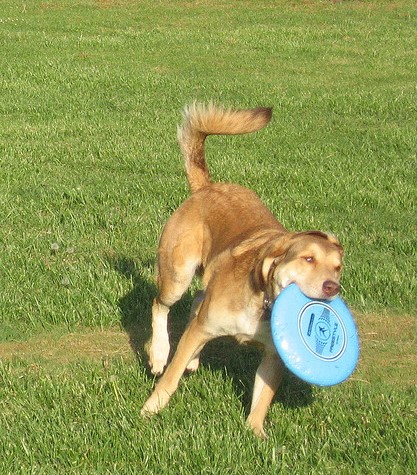}
\includegraphics[width=1.32cm]{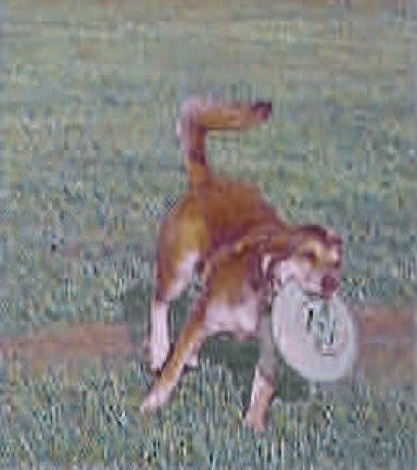}
\includegraphics[width=1.32cm]{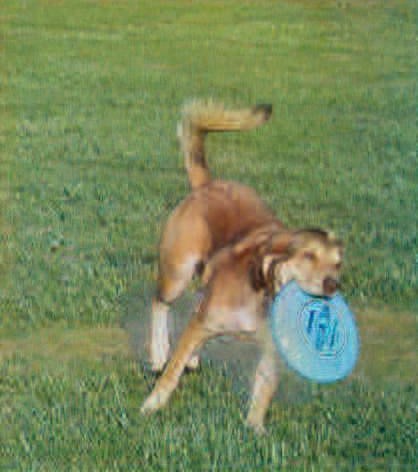}
\includegraphics[width=1.32cm]{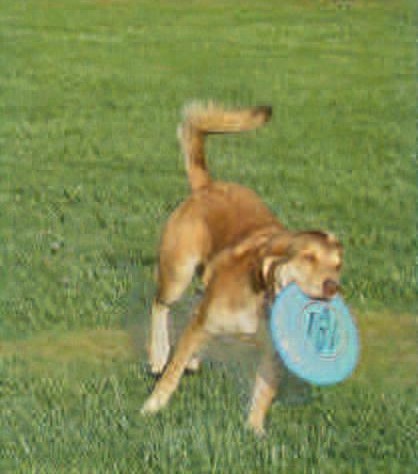}
\includegraphics[width=1.32cm]{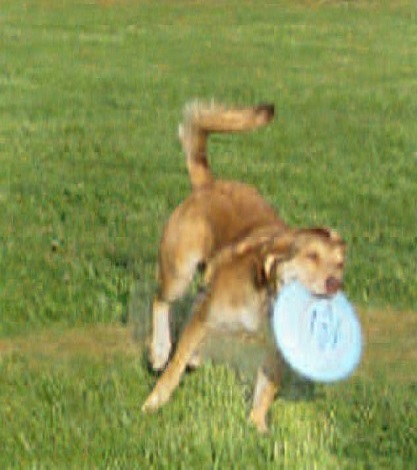}
\includegraphics[width=1.32cm]{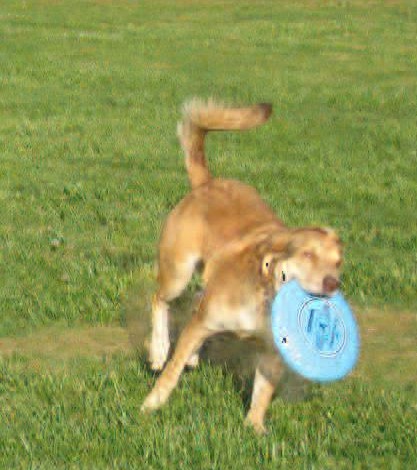}\\

\includegraphics[width=1.32cm]{COCO_train2014_000000171269input2.jpg}
\includegraphics[width=1.32cm]{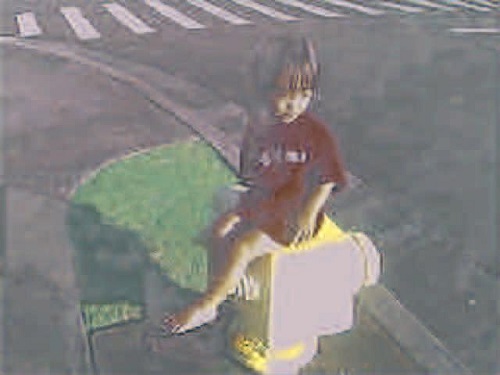}
\includegraphics[width=1.32cm]{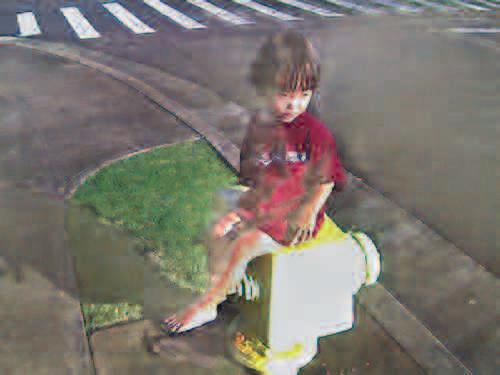}
\includegraphics[width=1.32cm]{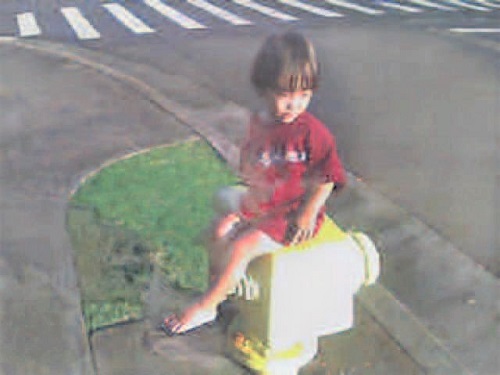}
\includegraphics[width=1.32cm]{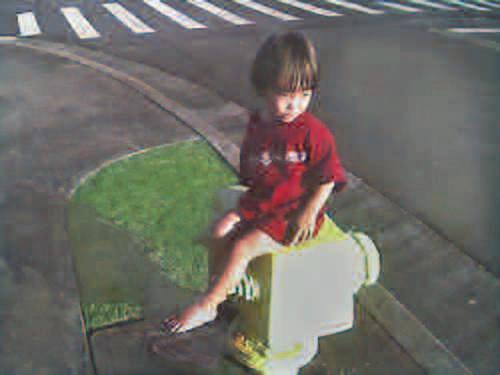}
\includegraphics[width=1.32cm]{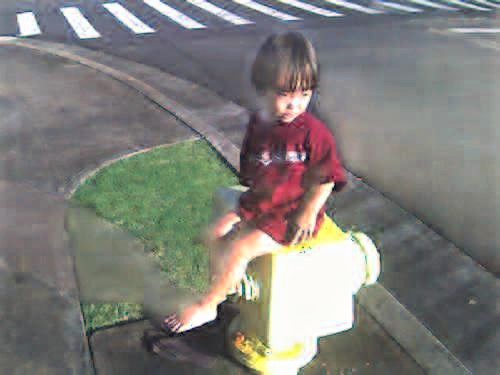}\\

\vspace{-0.15cm}
\subfigure[]{\includegraphics[width=1.32cm]{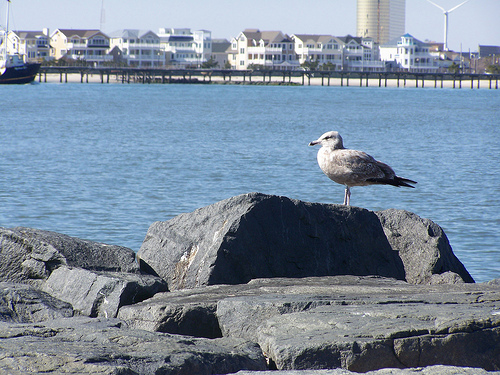}}
\subfigure[]{\includegraphics[width=1.32cm]{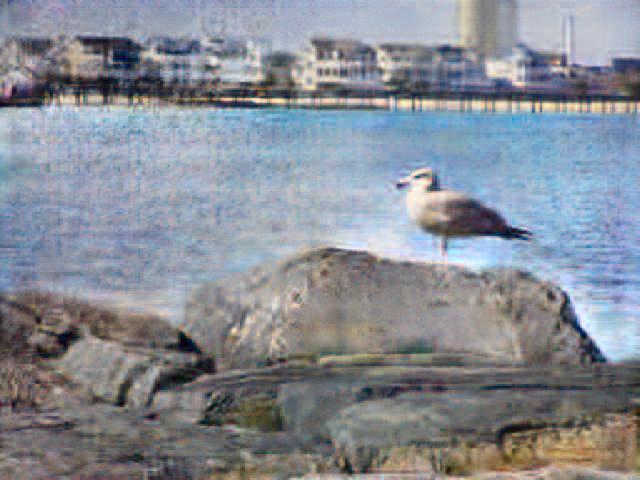}}
\subfigure[]{\includegraphics[width=1.32cm]{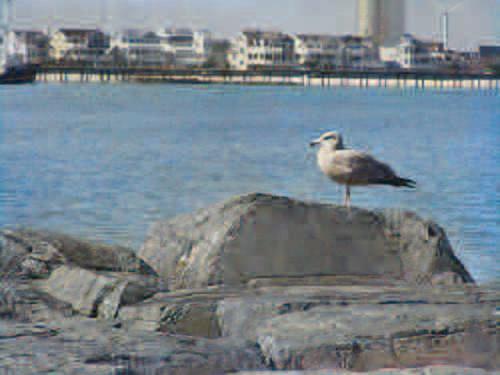}}
\subfigure[]{\includegraphics[width=1.32cm]{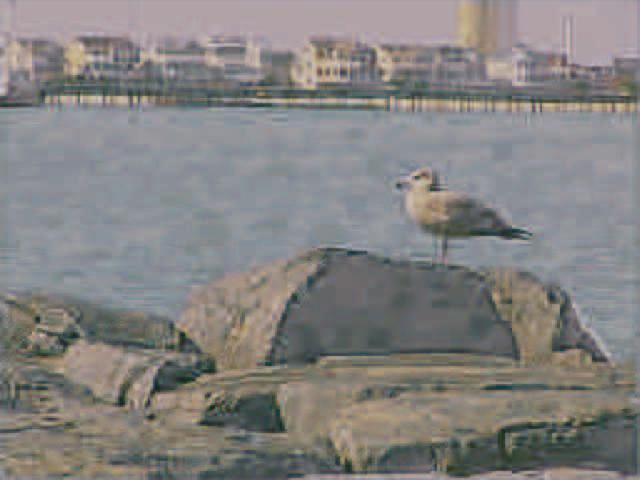}}
\subfigure[]{\includegraphics[width=1.32cm]{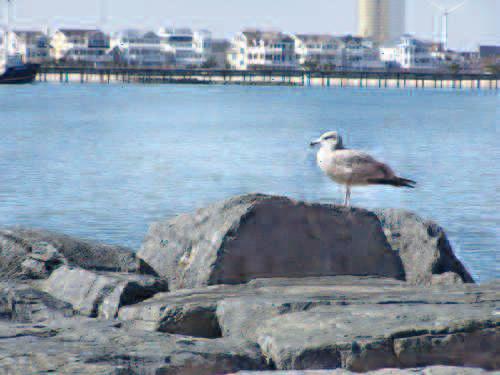}}
\subfigure[]{\includegraphics[width=1.32cm]{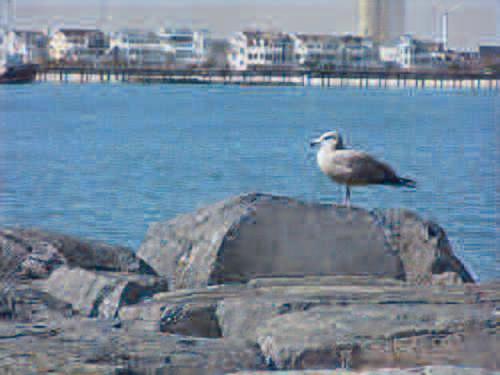}}
\end{center}
\vspace{-0.2cm}
\caption{Shadow removal results for real shadow images outside the two datasets. From left to right are: (a)input images; shadow removal results of (b)ST-CGAN, (c)DSC, (d)ARGAN, (e)RIS-GAN and (f)our CANet. }
\label{fig:robustness}
\vspace{-0.3cm}
\end{figure}

{\bfseries Robustness}.
To further verify the robustness of our method, we collect some real-world shadow images with complex scenes to conduct experiments and summarize the results in Figure \ref{fig:robustness}. Obviously, the shadow removal results generated by our CANet look more realistic compared to the other competing algorithms. These observations clearly demonstrate the robustness of our CANet to handle the complex real-world scenes.

\begin{figure}[ht]
\begin{center}
\includegraphics[width=2.7cm]{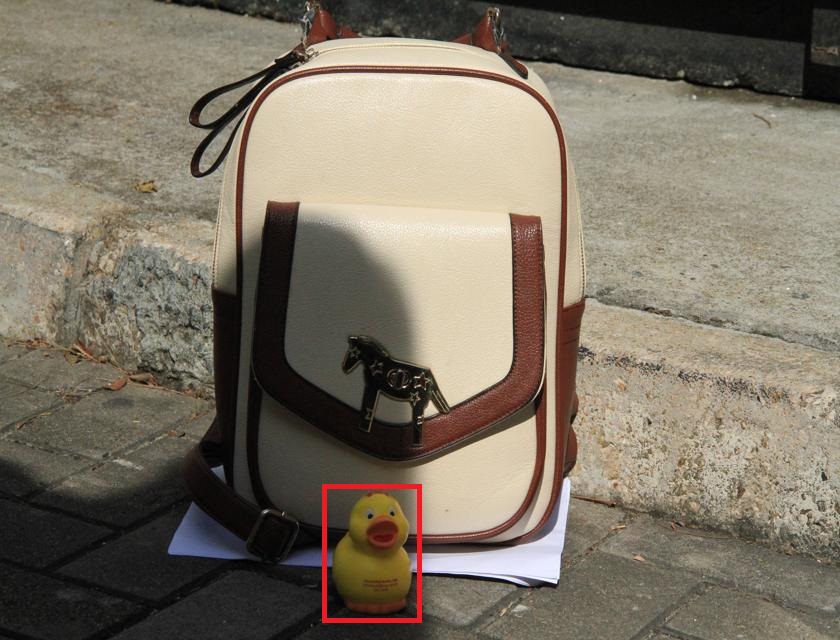}
\includegraphics[width=2.7cm]{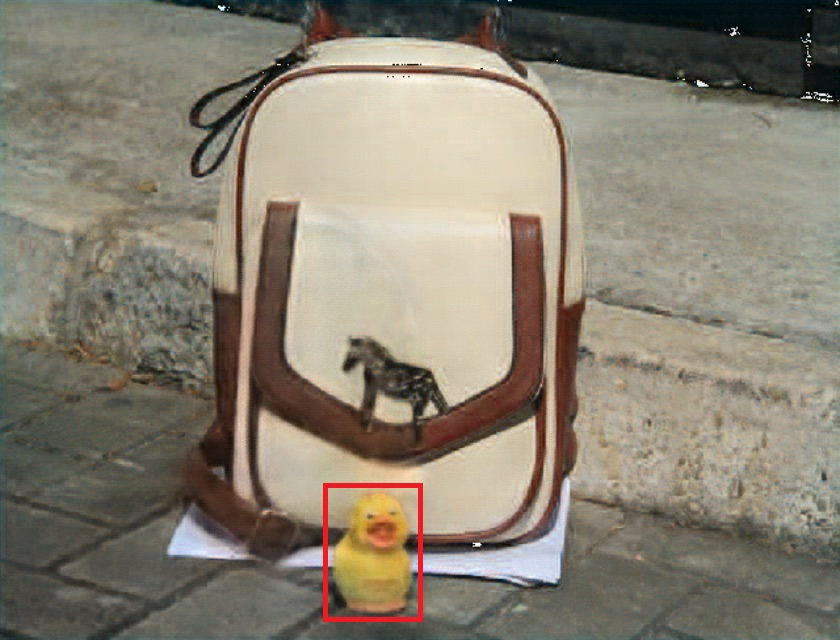}
\includegraphics[width=2.7cm]{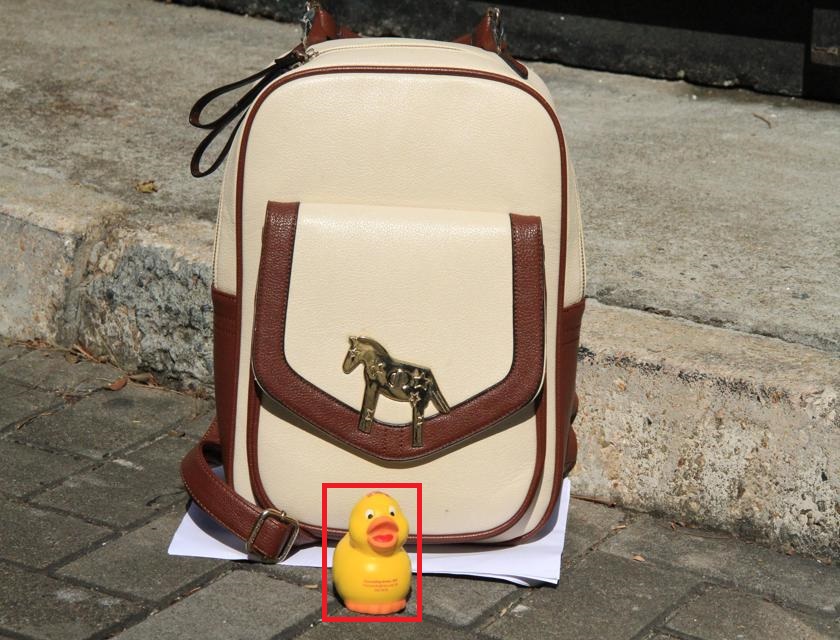}
\end{center}
   \caption{The illustration of the limitation of our method. From left to right are the input image, the result of our method and the shadow-free ground truth image. The red rectangular area in the input image does not have a matching position in the non-shadow area, which makes it difficult to recover the light in that area.}
\label{fig:limitation}
\vspace{-0.0cm}
\end{figure}

\begin{figure}[ht]
\begin{center}
\includegraphics[width=1.32cm]{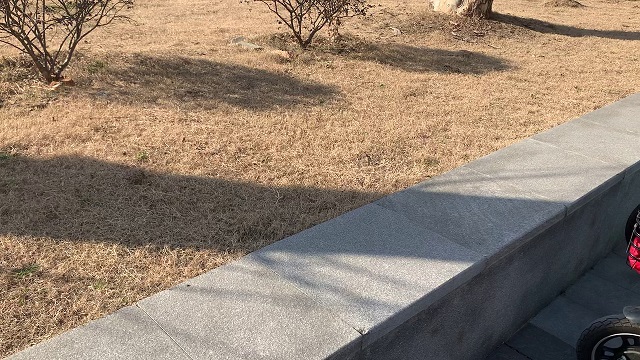}
\includegraphics[width=1.32cm]{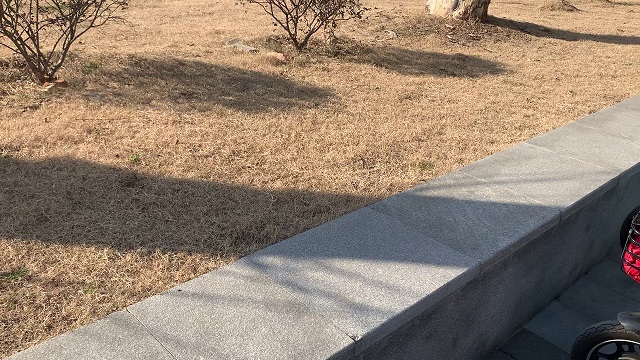}
\includegraphics[width=1.32cm]{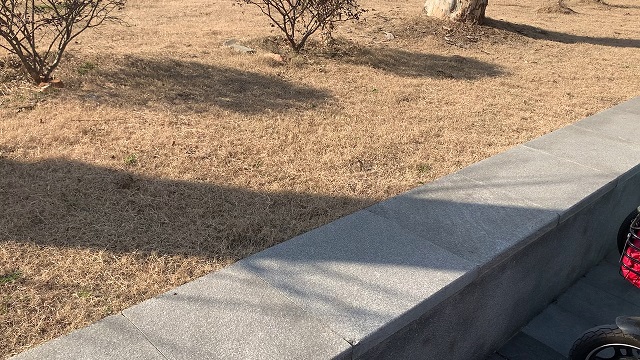}
\includegraphics[width=1.32cm]{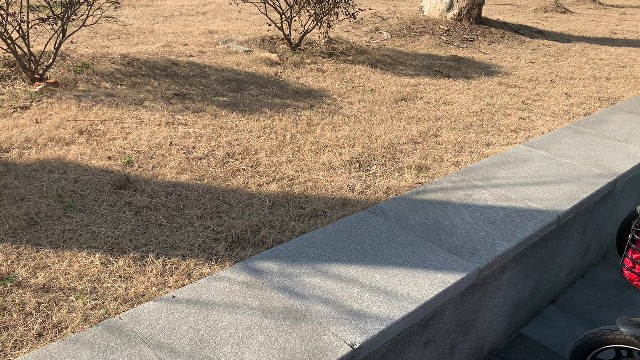}
\includegraphics[width=1.32cm]{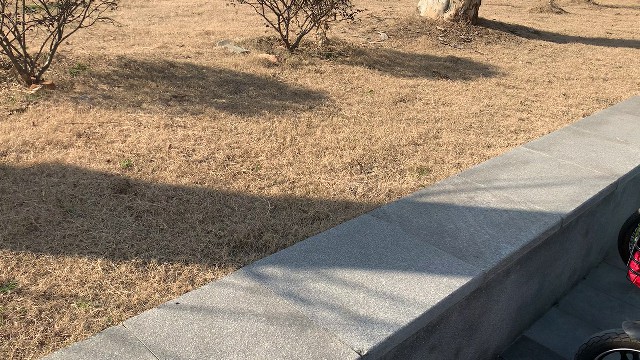}
\includegraphics[width=1.32cm]{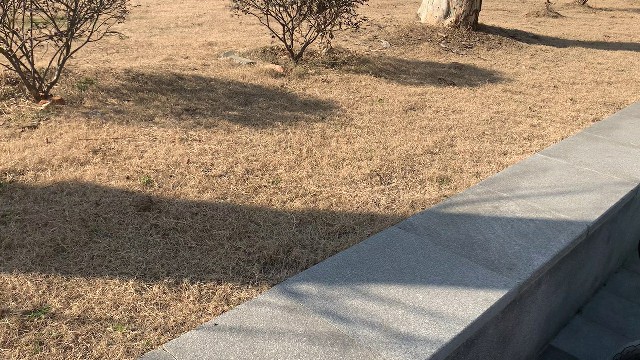}\\

\includegraphics[width=1.32cm]{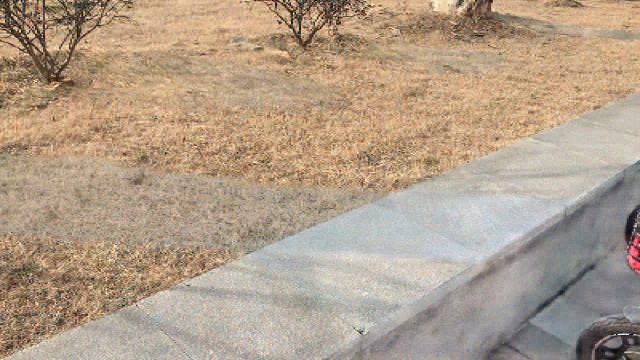}
\includegraphics[width=1.32cm]{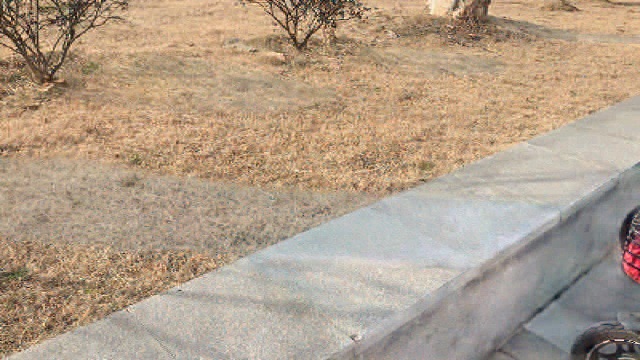}
\includegraphics[width=1.32cm]{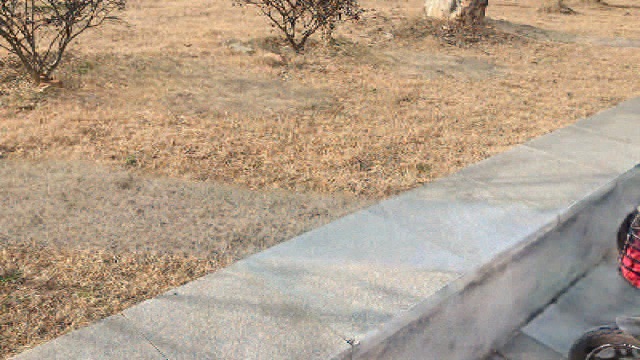}
\includegraphics[width=1.32cm]{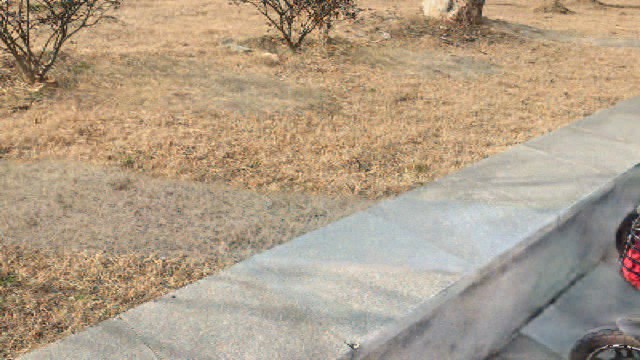}
\includegraphics[width=1.32cm]{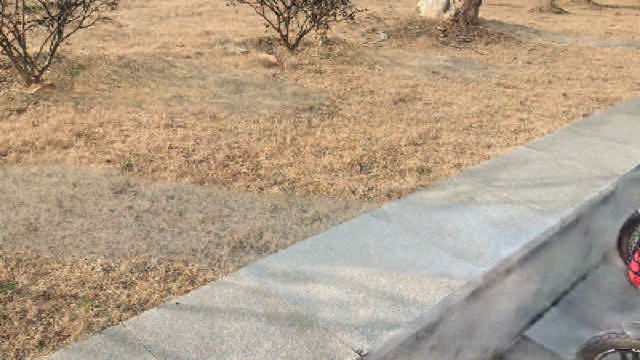}
\includegraphics[width=1.32cm]{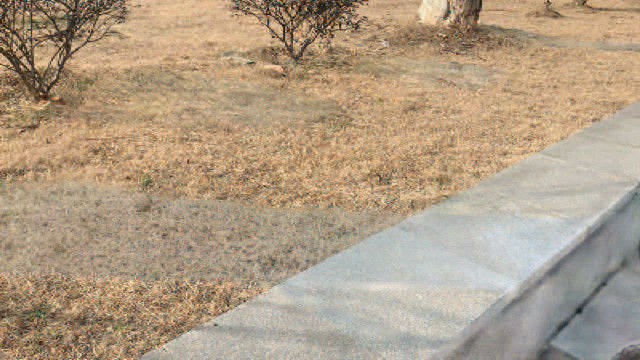}

\end{center}
\vspace{-0.1cm}
\caption{Shadow removal result in a video, a frame for every 0.5 seconds.}
\label{fig:video}
\vspace{-0.3cm}
\end{figure}

{\bfseries Extension to video-level shadow removal.}
We also apply our CANet to remove shadows in a video by processing each frame in the video separately. We take one frame result for every 0.5 seconds and visualize the results in Figure~\ref{fig:video}. From the results, we can see that our CANet can remove shadow well at frame-level, but the continuity of the video is not ensured, which we take as a part of our future work.  

{\bfseries Running time.}
It firstly takes around 10 hours to train our CPM module, then takes 32 hours to train our CANet on the ISTD and SRD training set. After training, only 1.8 seconds on average is required to process an $400\times400$ image. 

{\bfseries Limitation.}
Our proposed CANet can effectively remove shadows in images. However, it still has some limitations. 
(1) For some shadow images, if there is no strong contextual correlation between shadow and non-shadow regions, our CANet will fail to recover the illumination consistent with surroundings, as shown in Figure~\ref{fig:limitation}.
(2) Since the environmental luminosity and camera exposure may vary during photo shooting, training pairs may have inconsistent colors and luminosity~\cite{DSC}, causing our data-driven CANet produces shadow-removal results with color inconsistency.

\section{Conclusion}
In this paper, we have proposed a novel two-stage context aware network CANet for shadow removal. At stage-I, we design a contextual patch matching module (CPM) to search potential match pairs for the contextual feature transfer mechanism (CFT).  At stage-II, we apply an encoder-decoder to refine the results of stage-I to generate the final high-quality shadow-removal results. The extensive experiment results have strongly confirmed the effectiveness and superiority of our method. Our framework can be extended to handle more computer vision tasks such as highlight removal~\cite{fu-2019-specul-highl, fu-2021-multi-task}, which we take as the future work.

\section*{Acknowledgement}
This work was partly supported by the Key Technological Innovation Projects of Hubei Province (2018AAA062), NSFC (NO. 61972298), and CAAI-Huawei MindSpore Open Fund.

{\small
\bibliographystyle{ieee_fullname}
\bibliography{CANet}
}

\newpage
\appendixpage

\begin{appendix}
 \begin{figure*}[ht]
     \vspace{-0.3cm}
     \centering
     \includegraphics[width=1\linewidth]{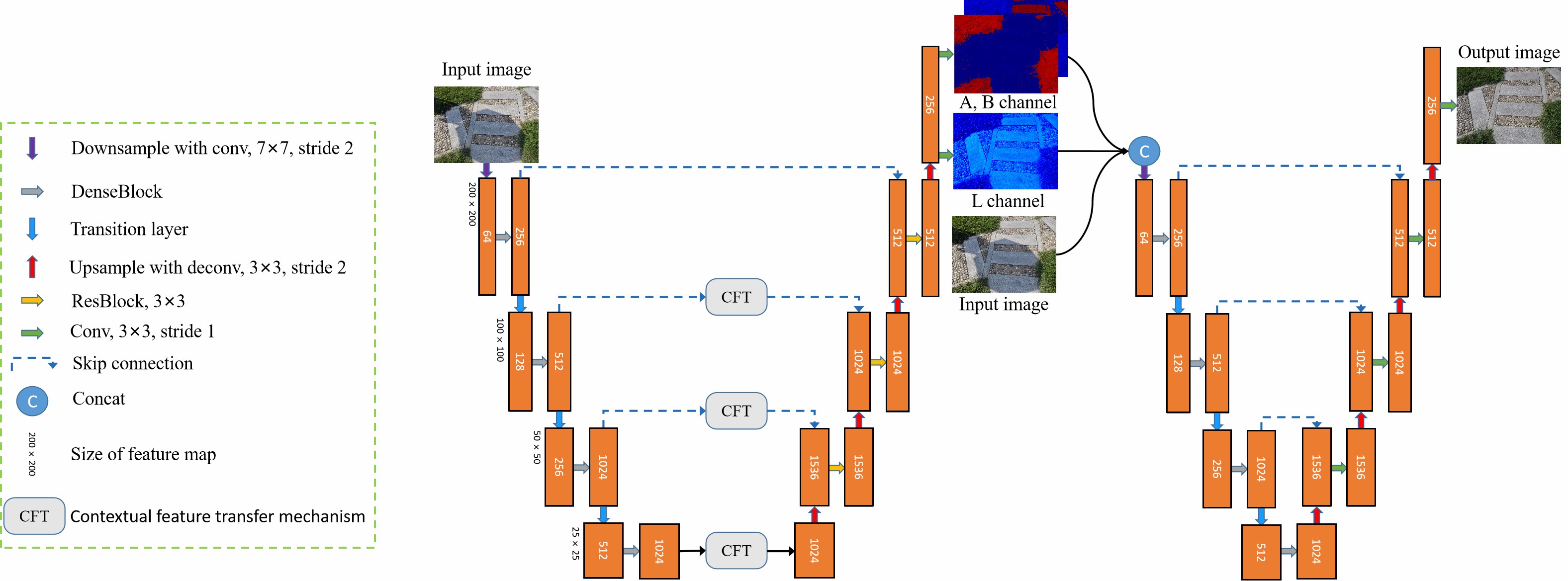}
     \caption{The network architecture of our CANet.}
     \label{fig:network}
 \end{figure*}
\section{Details about Network Architectures}
{\bf CPM module.}
The architecture of our CPM module is shown in Table \ref{table:cpm}, which divides into three parts: feature extractor, pair type classifier, and correlation degree regressor. The feature extractor is used to extract 256-dimensional feature representations for input patch pairs. It is designed with 4 convolutional layers, 3 residual blocks, and a bottleneck layer. The obtained representations are feed into the classifier and regressor separately. Both the classification head and the regression head contain 3 fully connected layers, and classification head also have a softmax layer.

\begin{table}[ht]
\caption{The architecture of our CPM module. It contains a feature extractor, a pair type classifier, and a correlation degree regressor.}
\centering
\resizebox{\linewidth}{!}{
\begin{tabular}{c|c|c|c}
\hline
& \textbf{Layer} & \textbf{Output Size} & \textbf{Operation} \\
\hline
\multirow{8}{*}{\begin{tabular}[c]{@{}c@{}}Feature\\ extractor\end{tabular}} & Conv1 & $16 \times 16 \times 64 $ & Conv($3 \times 3$ stride 2) \\
& Res1 & $16 \times 16 \times 64 $ & Res-blocks($3 \times 3$) \\
& Conv2 & $8 \times 8 \times 96 $ & Conv($3 \times 3$ stride 2) \\
& Res2 & $8 \times 8 \times 96 $ & Res-blocks($3 \times 3$) \\
& Conv3 & $4 \times 4 \times 96 $ & Conv($3 \times 3$ stride 2) \\
& Res3 & $4 \times 4 \times 96 $ & Res-blocks($3 \times 3$) \\
& Conv4 & $4 \times 4 \times 64 $ & Conv($3 \times 3$ stride 1) \\
& Bottleneck & 256  & FC \\
\hline
\multirow{4}{*}{Classifier} & FC1 & 256 & FC \\
& FC2 & 128 & FC \\
& FC3 & 3 & FC \\
& Softmax & 3 & Softmax \\
\hline
\multirow{4}{*}{Regressor} & FC1 & 256 & FC \\
& FC2 & 128 & FC \\
& FC3 & 1 & FC \\
\hline
\end{tabular}
}
\label{table:cpm}
\end{table}

{\bf CANet.}
 In Figure \ref{fig:network}, we illustrate the detailed network architecture of our proposed CANet. Each orange rectangles in the network is the feature map of the corresponding layer, and the number in the rectangles is their channel number. Note that the ``DenseBlock", ``Transition layer" are followed as the original version of DenseNet \cite{Densenet}. 
 
\label{appendix}
\section{Better Understanding for CPM Module}
\subsection{Effectiveness of Light-unaware Images}
\begin{figure}[ht]
\vspace{-0.3cm}
\begin{center}
\includegraphics[width=4.1cm]{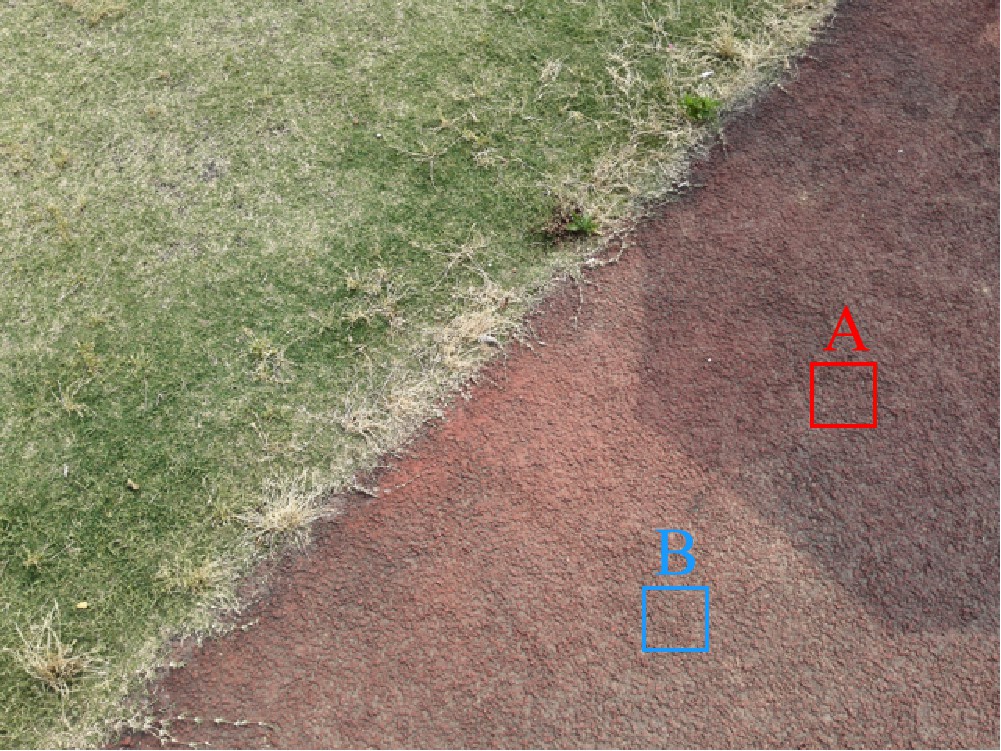}
\includegraphics[width=4.1cm]{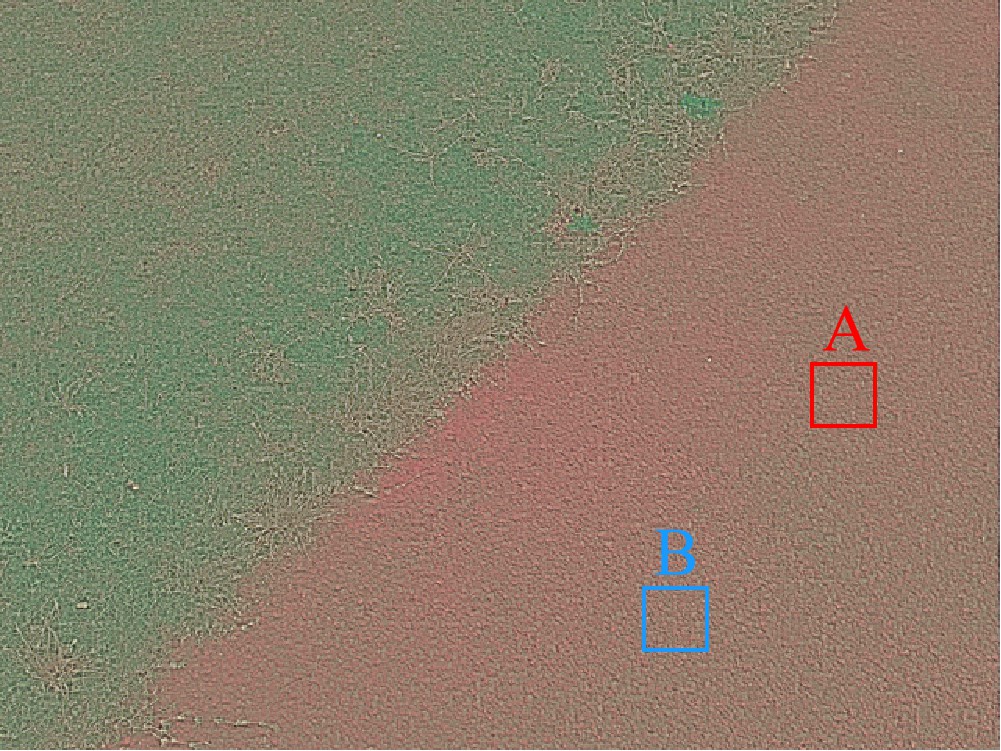}\\

\includegraphics[width=4.1cm]{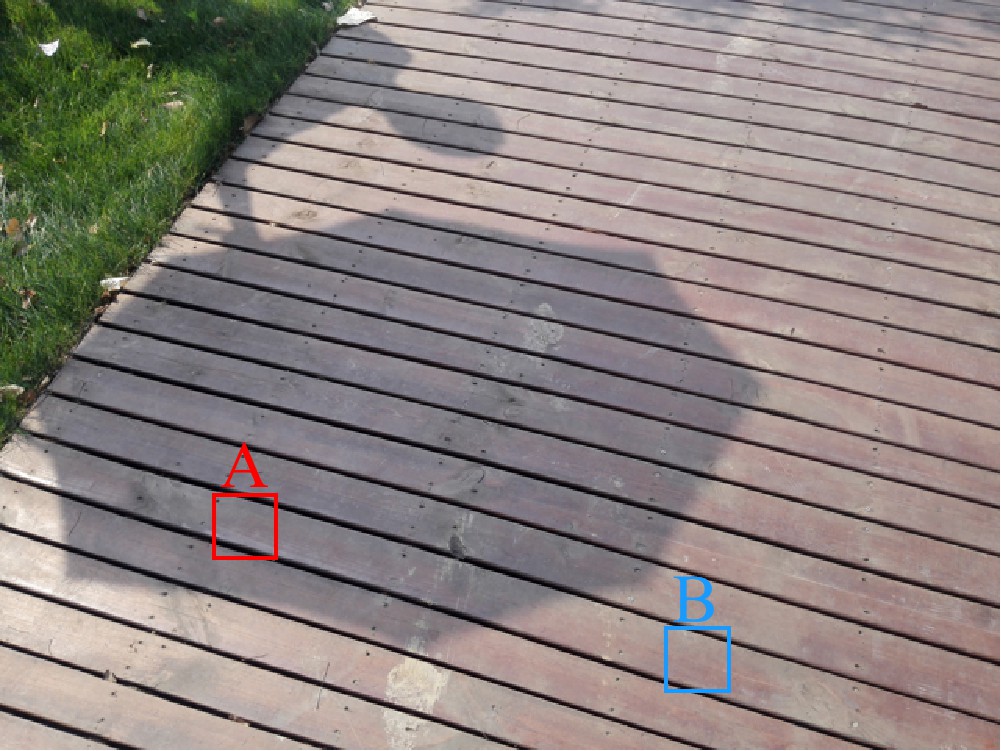}
\includegraphics[width=4.1cm]{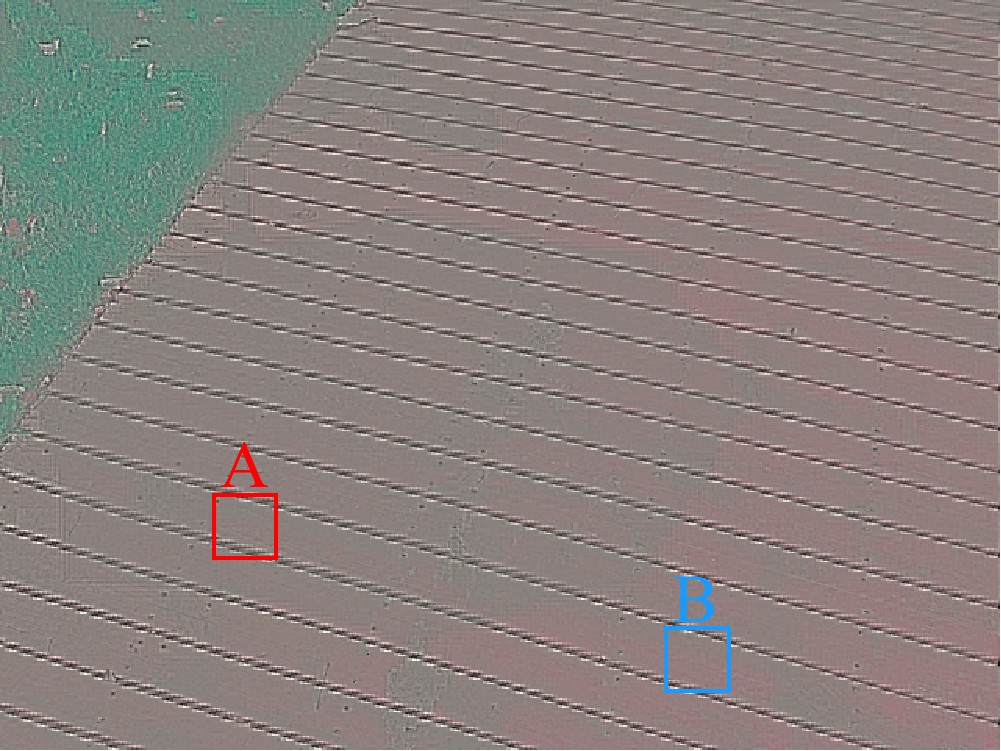}\\

\end{center}
\vspace{-0.2cm}
\caption{From left to right are: input shadow image; and the proposed light-unaware image, which can eliminate the difference between region A and B caused by shadow.}
\label{fig:figure2}
\end{figure}

As shown in Figure \ref{fig:figure2}, with the supplementary light-unaware image, we can largely eliminate the difference between shadow regions and non-shadow regions, which effectively avoids the matching errors caused by shadows.

\begin{figure}[ht]
\vspace{-0.3cm}
\begin{center}
\includegraphics[width=2.7cm]{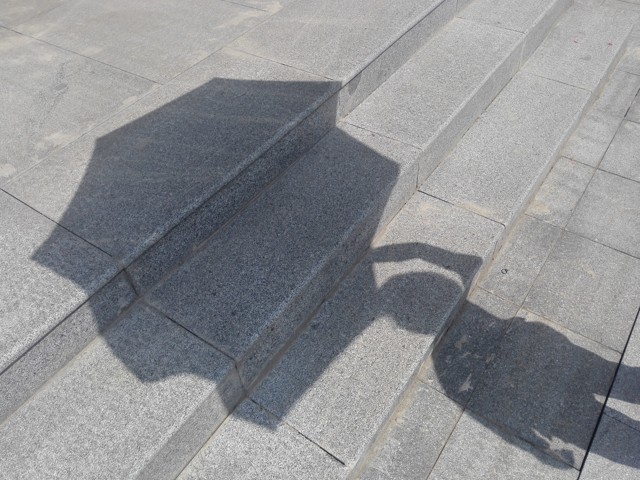}
\includegraphics[width=2.7cm]{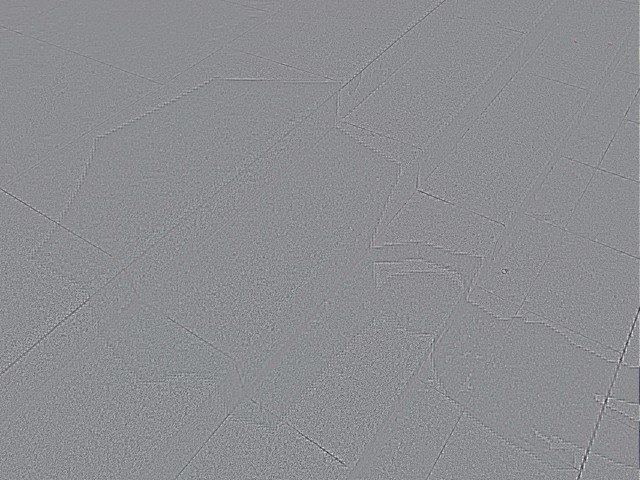}
\includegraphics[width=2.7cm]{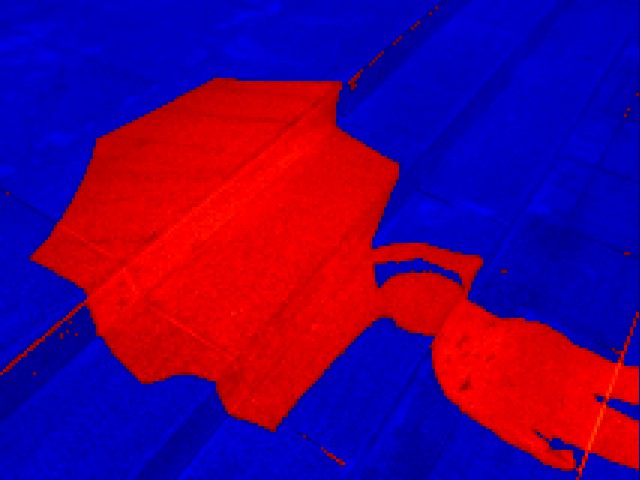}\\

\vspace{-0.15cm}
\subfigure[]{\includegraphics[width=2.7cm]{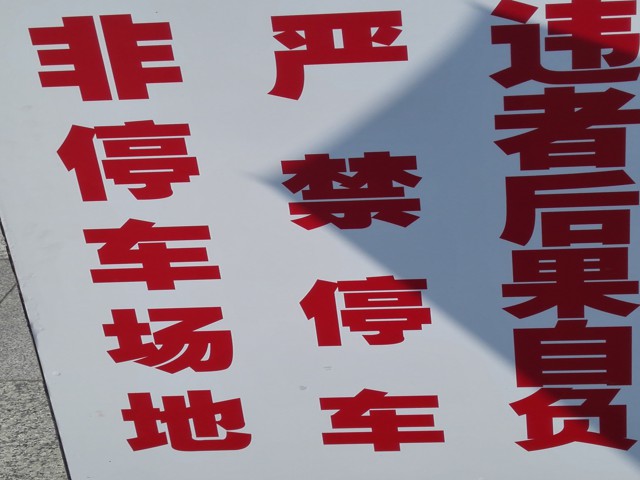}}
\subfigure[]{\includegraphics[width=2.7cm]{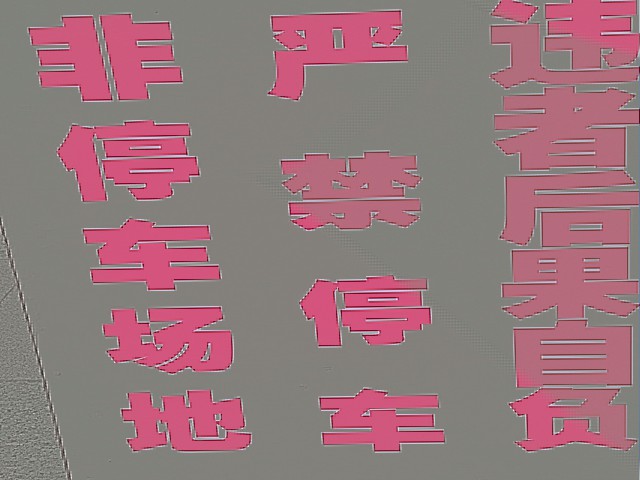}}
\subfigure[]{\includegraphics[width=2.7cm]{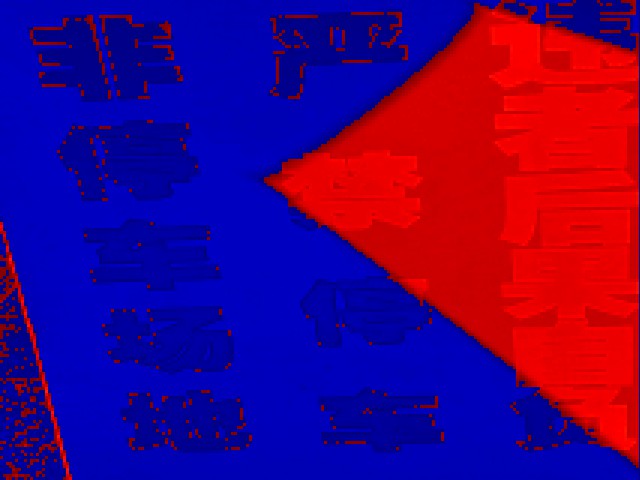}}\\

\end{center}
\vspace{-0.2cm}
\caption{The illustration of the difference between shadow image and light-unaware image, from left to right are: (a)input shadow image; (b)proposed light-unaware image and (c)the difference between them.}
\label{fig:figure3}
\end{figure}

Also, from Figure \ref{fig:figure3}, we can observe that there is a larger difference between shadow image and light-unaware image in shadow regions while a smaller difference in the non-shadow region. It suggests that the shadow image and light-unaware image pair can provide an indication to distinguish shadow patches from non-shadow patches, which can be used to perform our pair type classifier.

\subsection{Large-scale Training Dataset for CPM} 
\begin{figure}
    \vspace{-0.3cm}
    \centering
    \includegraphics[width=1\linewidth]{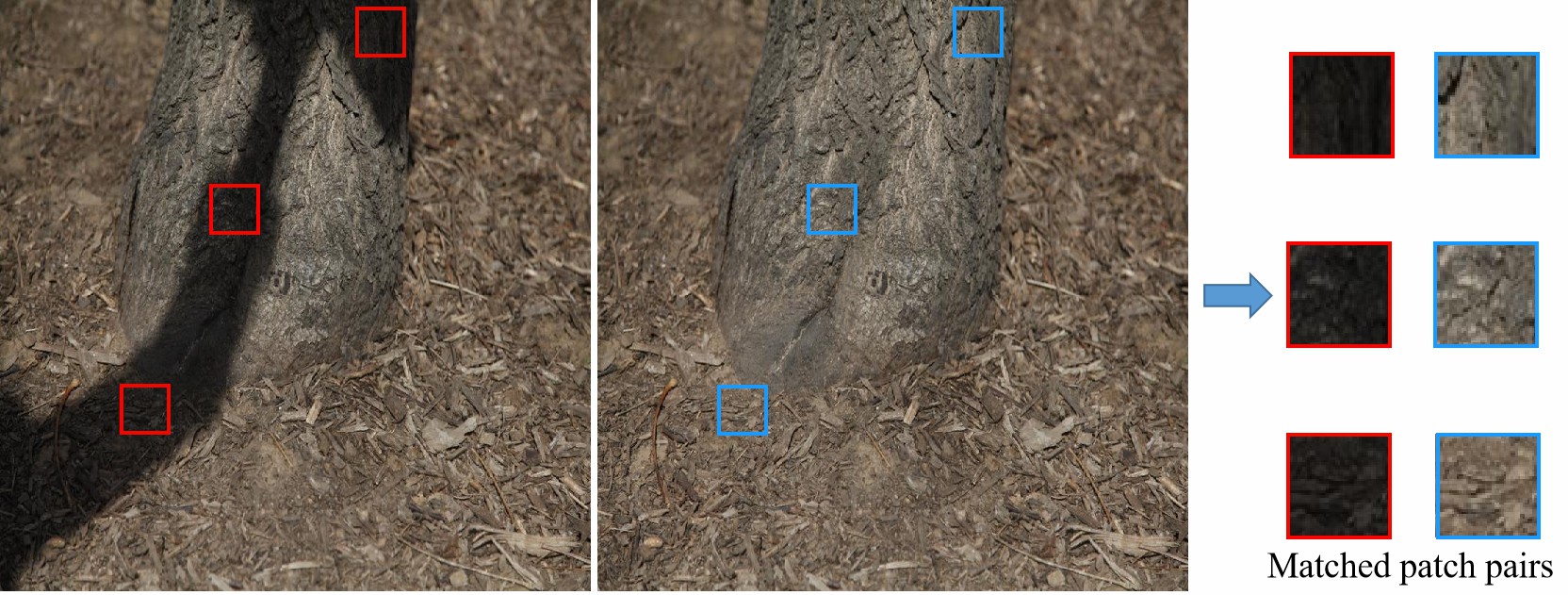}
    \caption{The illustration of our first way to collect matched patch pairs.}
    \label{fig:dataset1}
\end{figure}

To train our CPM module, we collect a large-scale training collection from the existing shadow benchmark datasets: ISTD \cite{Stc-gan} and SRD \cite{Deshadownet}. The collected training dataset contains more than 360,000 and 600,000 
patch pairs separately (50$\%$ match pairs and 50$\%$ non-match pairs). These patch pairs are collected from two ways: (1) we select a shadow patch in the shadow image and a matched non-shadow patch in the shadow-free image, which has the same position as the shadow patch, as illustrated in Figure \ref{fig:dataset1}; (2) we select a shadow patch from shadow regions and find another matched patch from non-shadow regions in the shadow image. Specially, we randomly select two patches from shadow and non-shadow regions in the shadow image and calculate the cosine similarity between the two patches in the corresponding shadow-free image. We choose the pairs with cosine similarity higher than 0.95 as the matching pair and less than 0.6 as the non-match pair, as shown in Figure \ref{fig:dataset1}. Due to the lack of shadow mask ground-truth in SRD dataset \cite{Densenet}, we firstly use the results of the latest shadow detection method DSD \cite{dsd}, and then manually choose the correct results as the approximate ground-truth during the process of dataset collecting.

\begin{figure}
    \vspace{-0.3cm}
    \centering
    \includegraphics[width=1\linewidth]{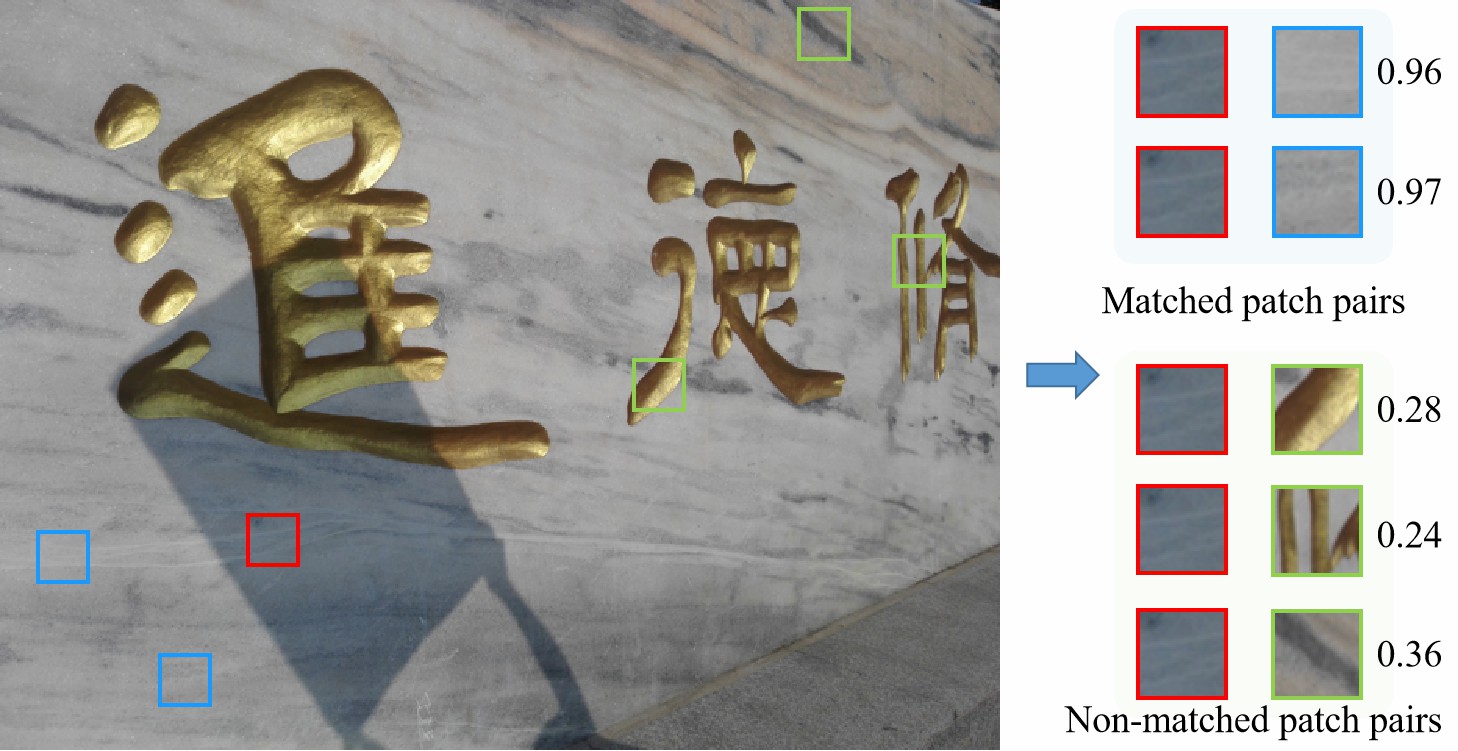}
    \caption{The illustration of our second way to collect matched and non-matched patch pairs.}
    \label{fig:dataset2}
\end{figure}

\section{More Visual Shadow Removal Results}
We provide more visual shadow removal comparison results in Figure \ref{fig:compare}. Here, we compare our CANet with six state-of-the-art methods, {\em i.e.}, Guo ~\cite{Guo_2011}, Zhang ~\cite{Zhang_2015}, ST-CGAN ~\cite{Stc-gan}, ARGAN ~\cite{ARGAN}, DSC ~\cite{DSC} and RIS-GAN ~\cite{RIS-GAN}.

\begin{figure*}[ht]

\centering

\includegraphics[width=1.86cm]{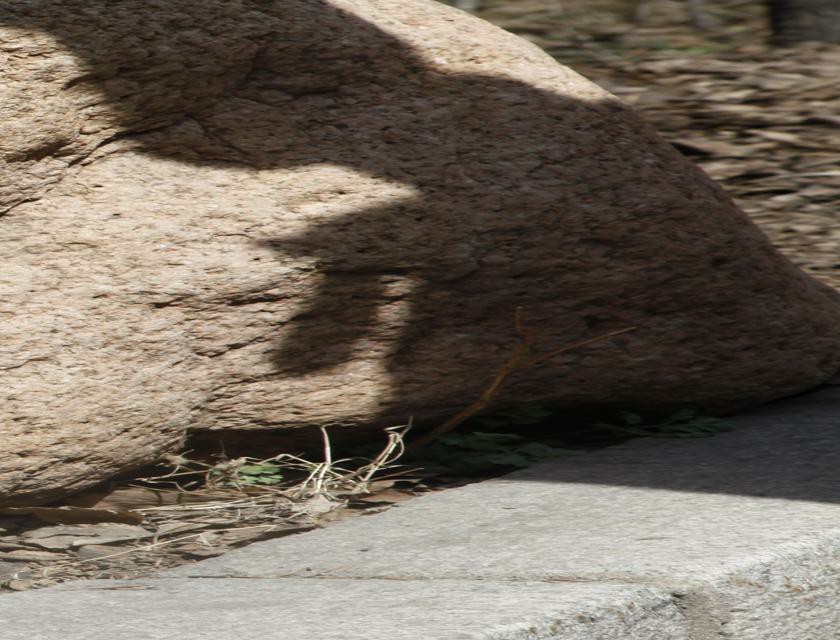}
\includegraphics[width=1.86cm]{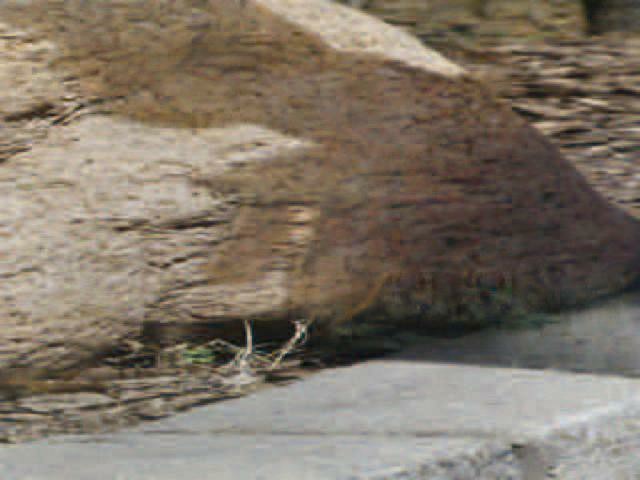}
\includegraphics[width=1.86cm]{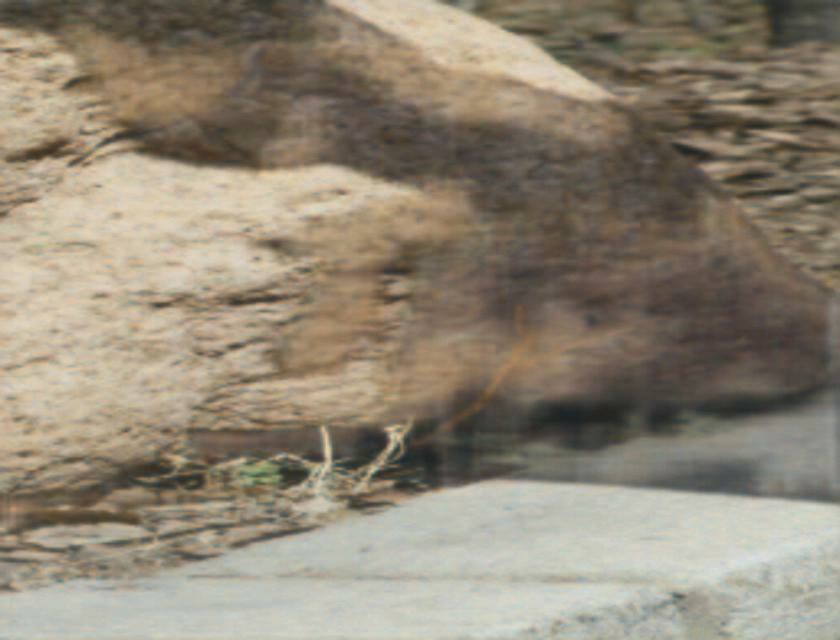}
\includegraphics[width=1.86cm]{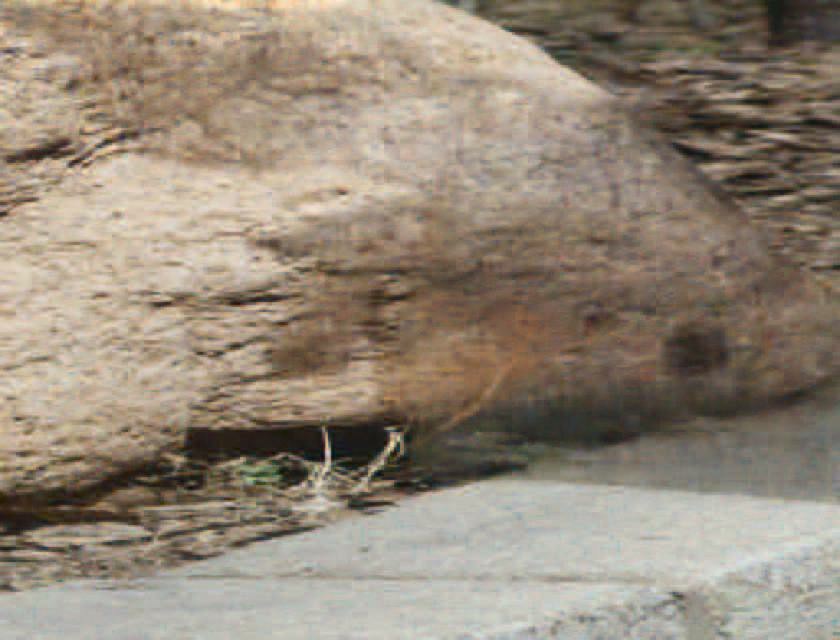}
\includegraphics[width=1.86cm]{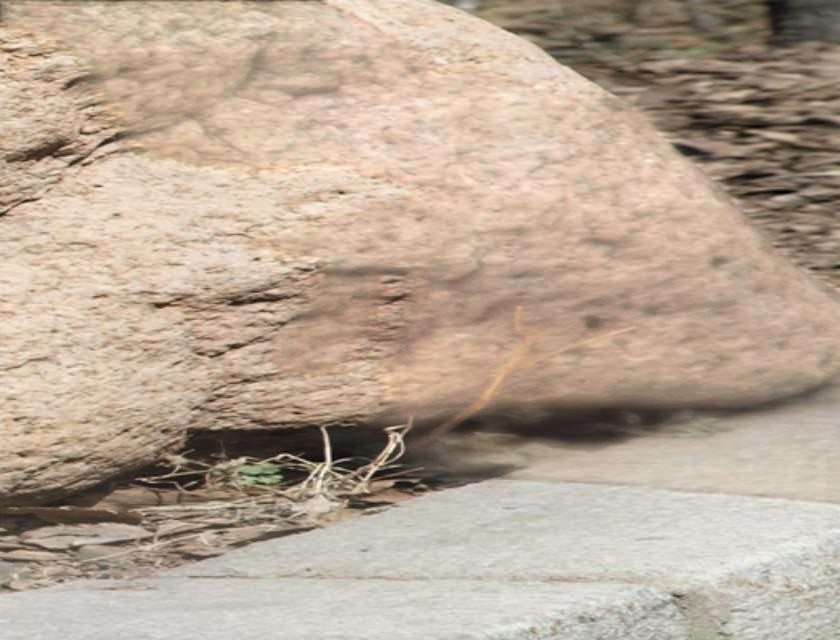}
\includegraphics[width=1.86cm]{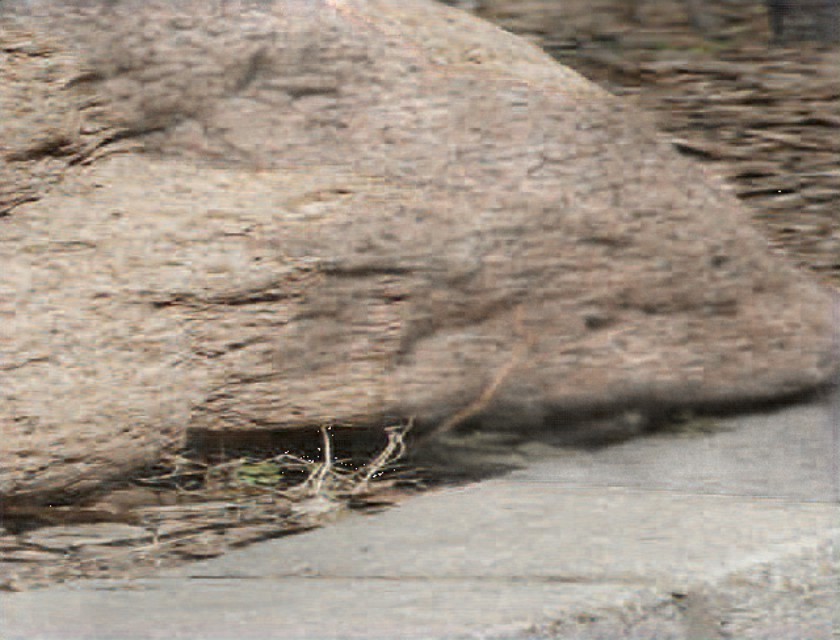}
\includegraphics[width=1.86cm]{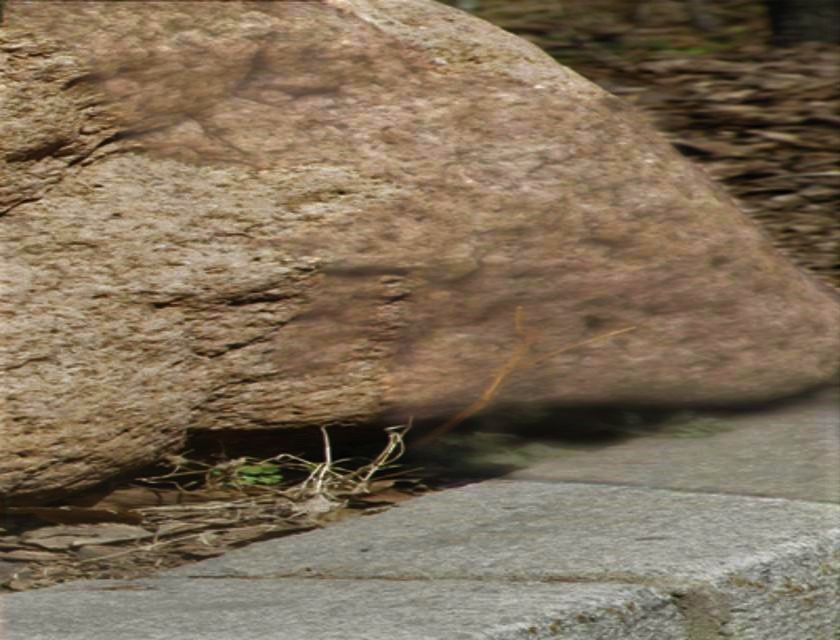}
\includegraphics[width=1.86cm]{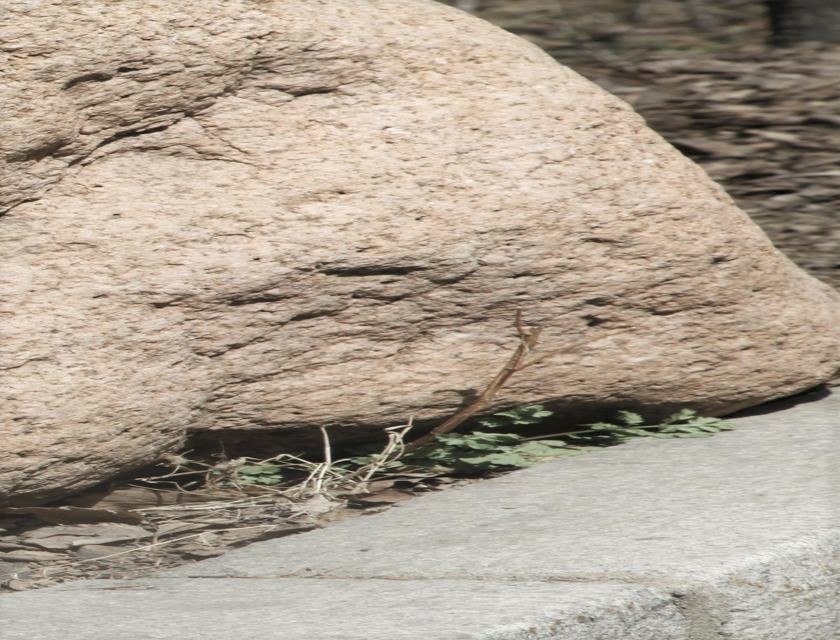}
\includegraphics[width=1.86cm]{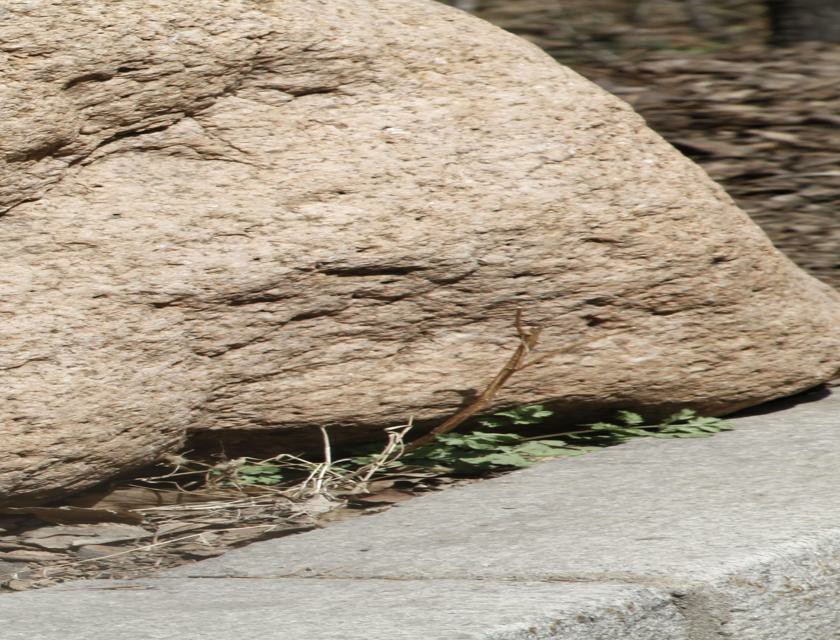}\\
\includegraphics[width=1.86cm]{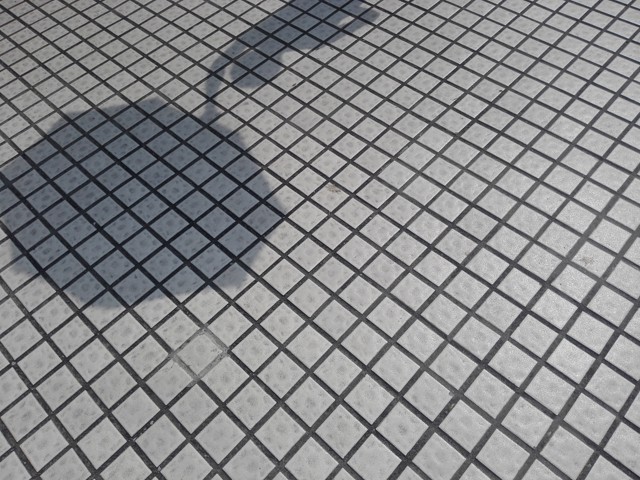}
\includegraphics[width=1.86cm]{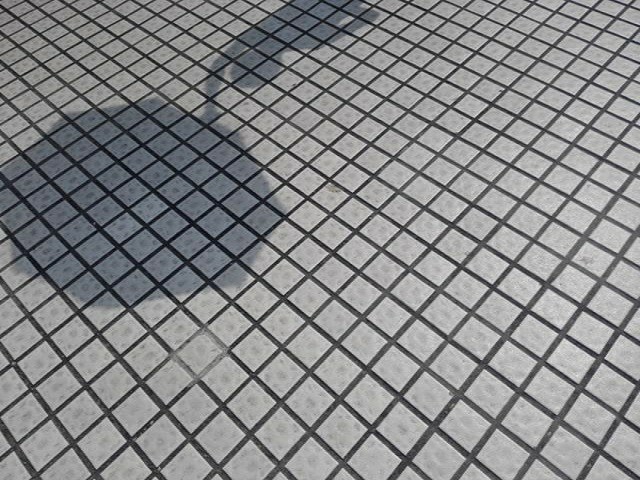}
\includegraphics[width=1.86cm]{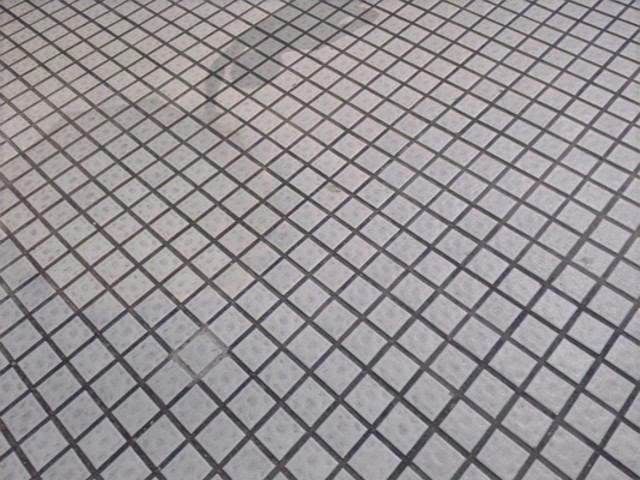}
\includegraphics[width=1.86cm]{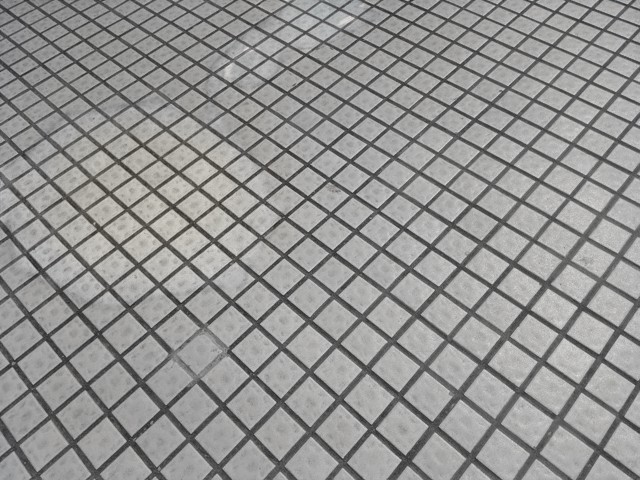}
\includegraphics[width=1.86cm]{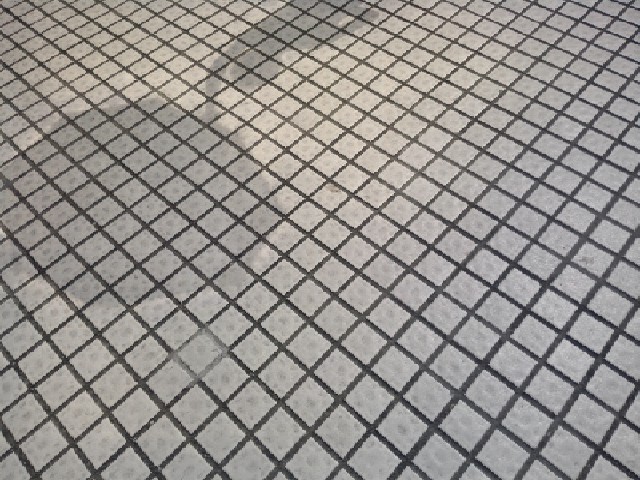}
\includegraphics[width=1.86cm]{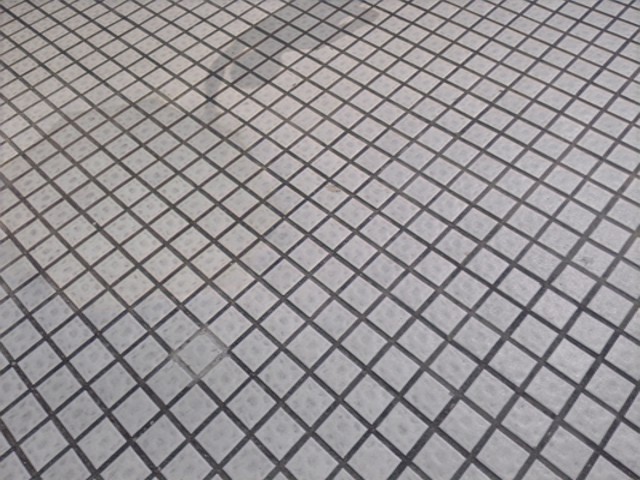}
\includegraphics[width=1.86cm]{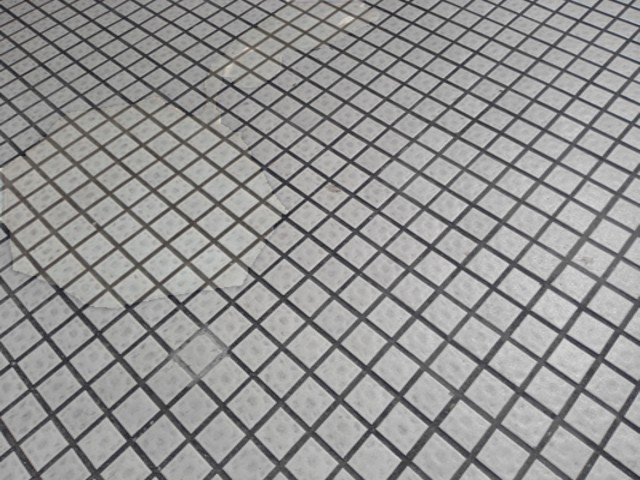}
\includegraphics[width=1.86cm]{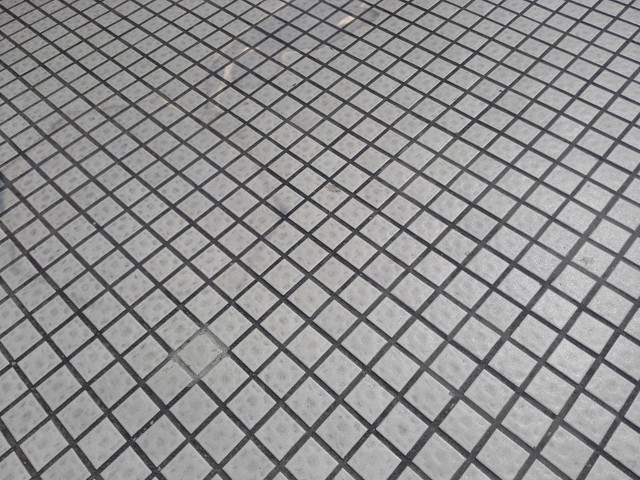}
\includegraphics[width=1.86cm]{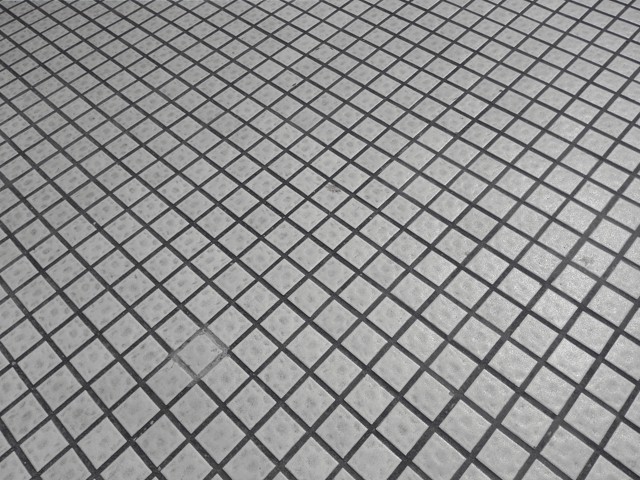}\\
\includegraphics[width=1.86cm]{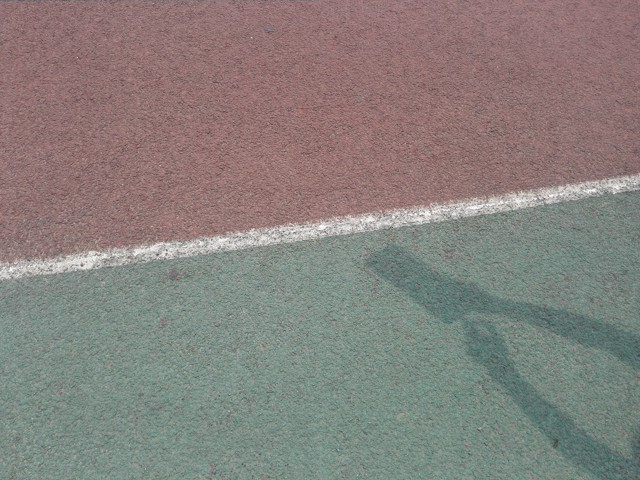}
\includegraphics[width=1.86cm]{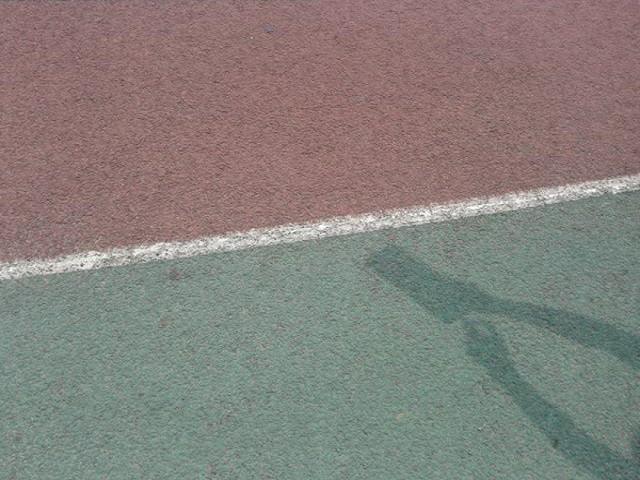}
\includegraphics[width=1.86cm]{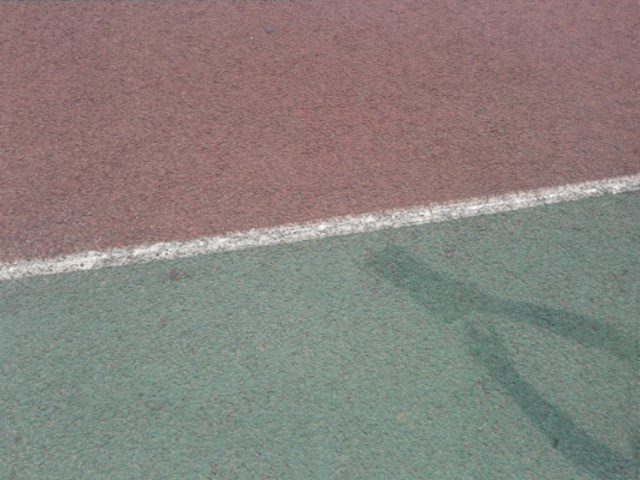}
\includegraphics[width=1.86cm]{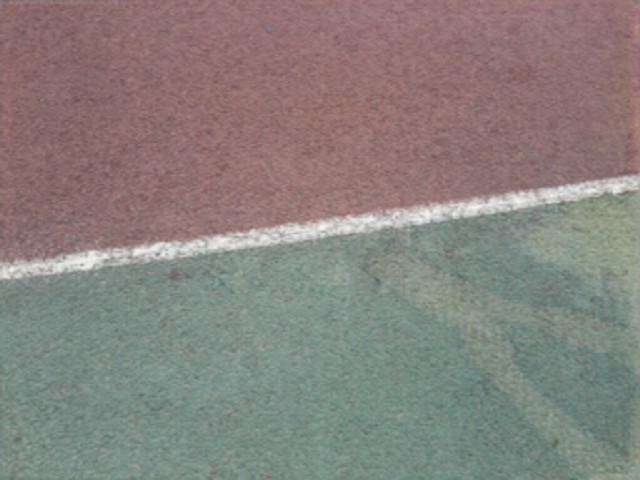}
\includegraphics[width=1.86cm]{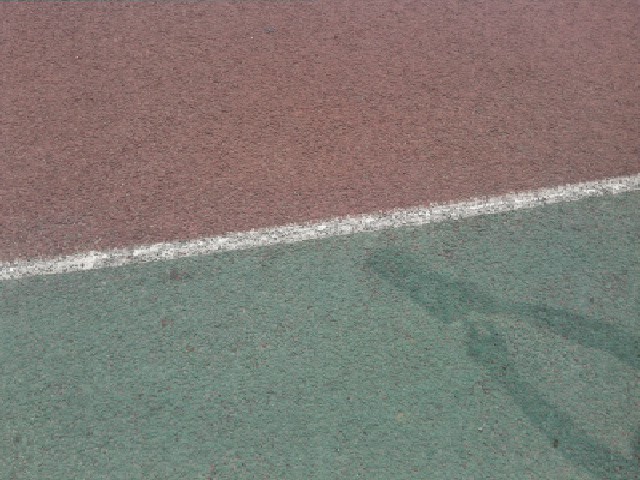}
\includegraphics[width=1.86cm]{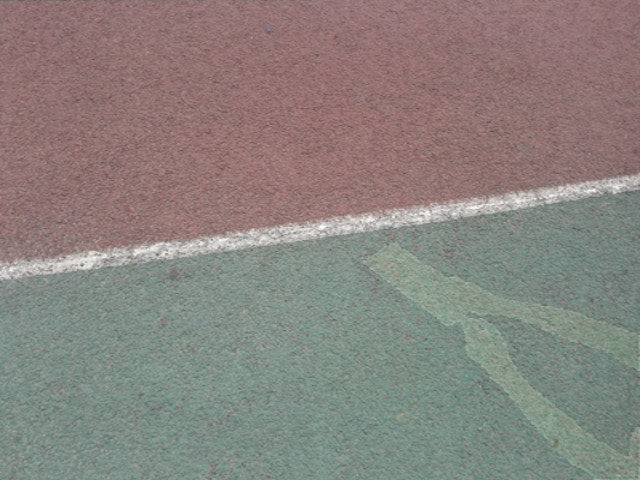}
\includegraphics[width=1.86cm]{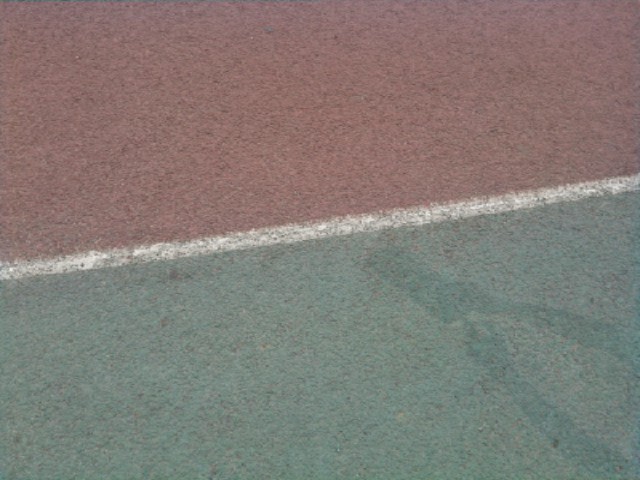}
\includegraphics[width=1.86cm]{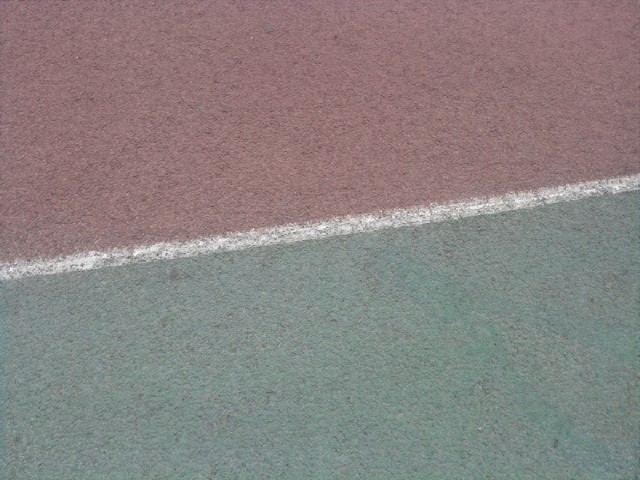}
\includegraphics[width=1.86cm]{sumpplemental_new/90-1canet.jpg}\\
\includegraphics[width=1.86cm]{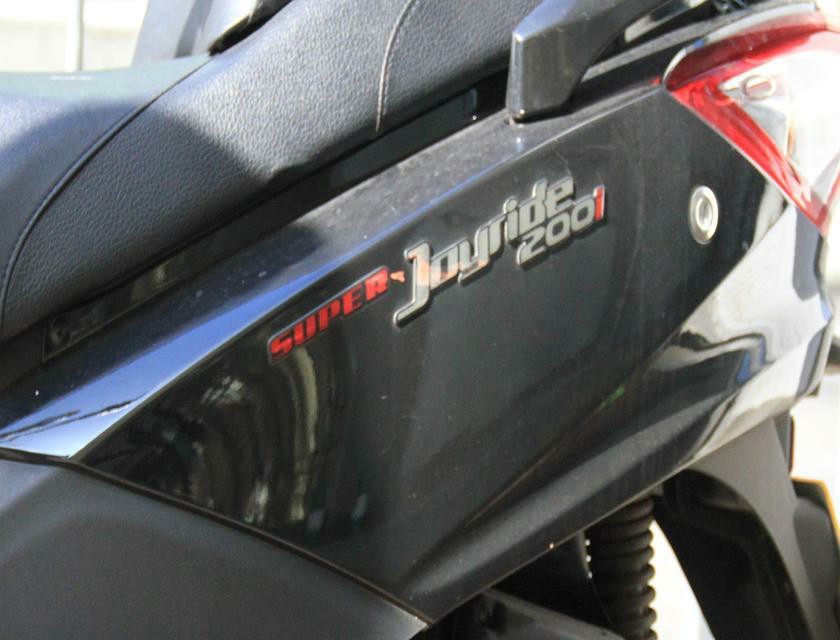}
\includegraphics[width=1.86cm]{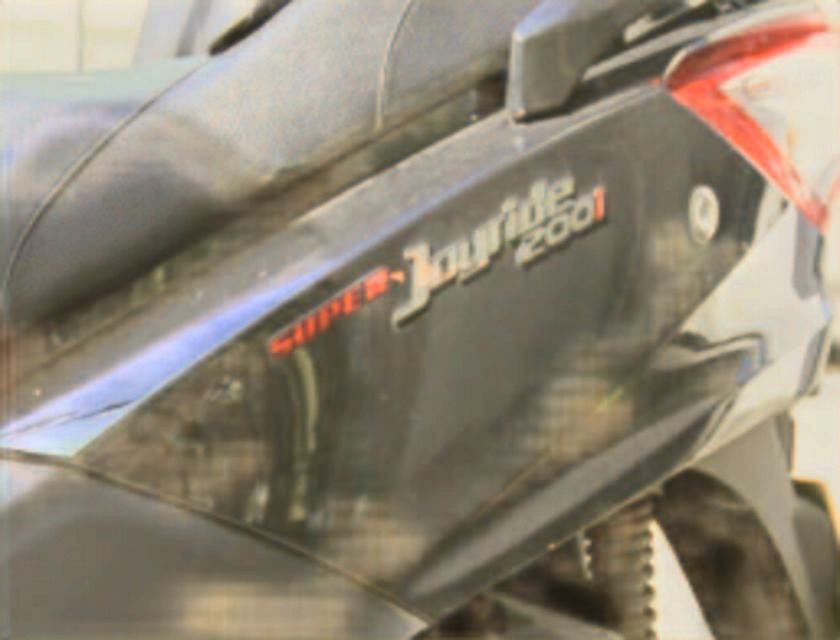}
\includegraphics[width=1.86cm]{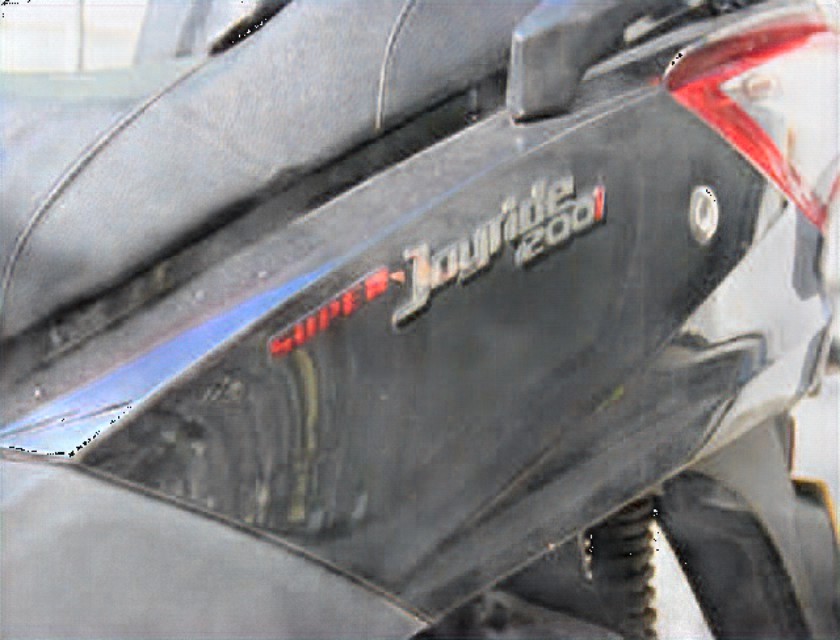}
\includegraphics[width=1.86cm]{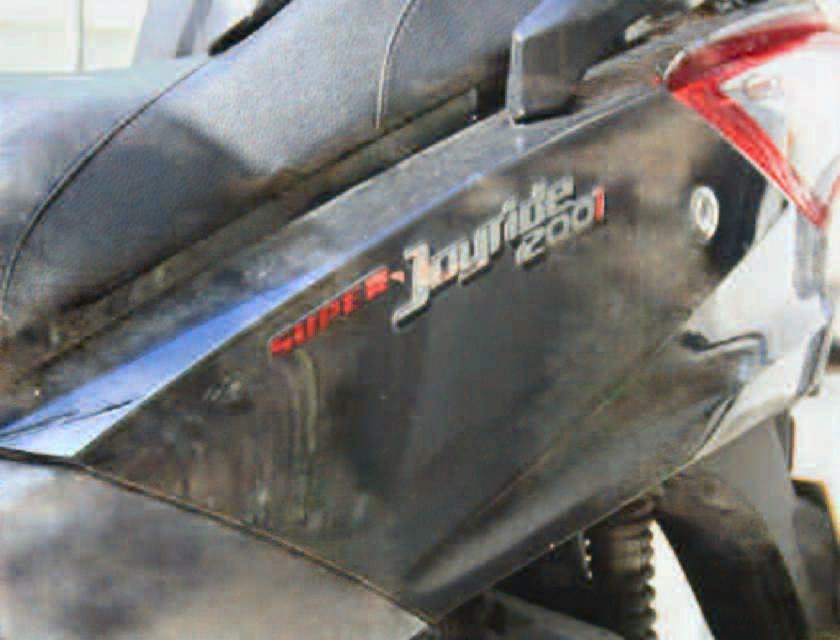}
\includegraphics[width=1.86cm]{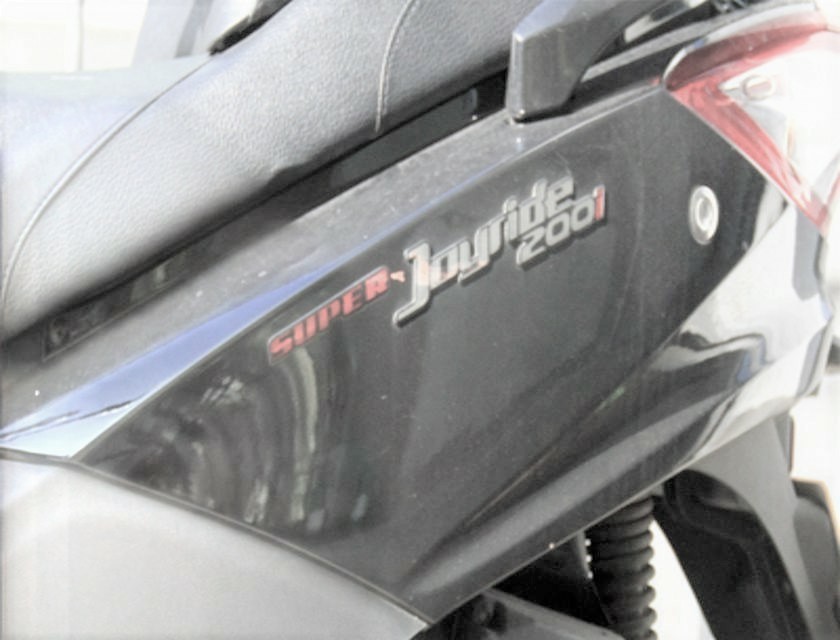}
\includegraphics[width=1.86cm]{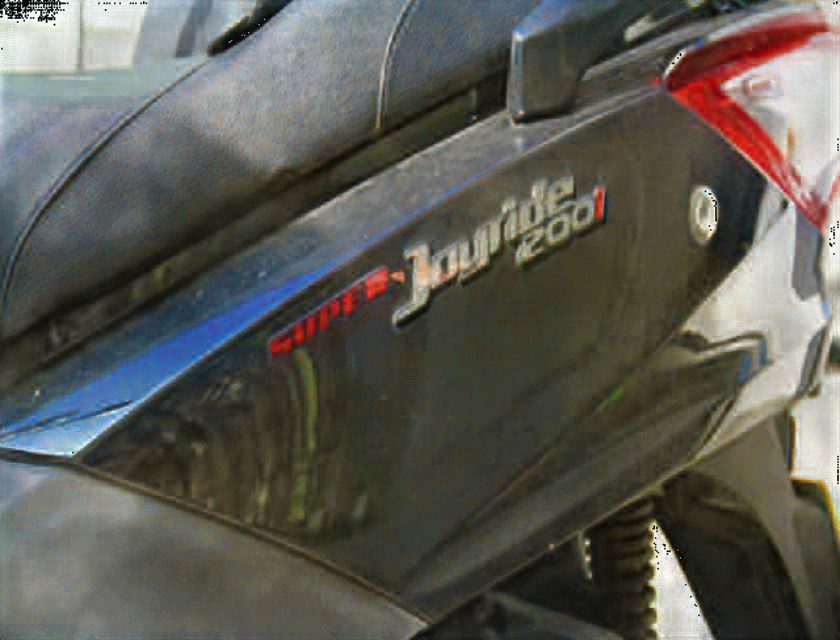}
\includegraphics[width=1.86cm]{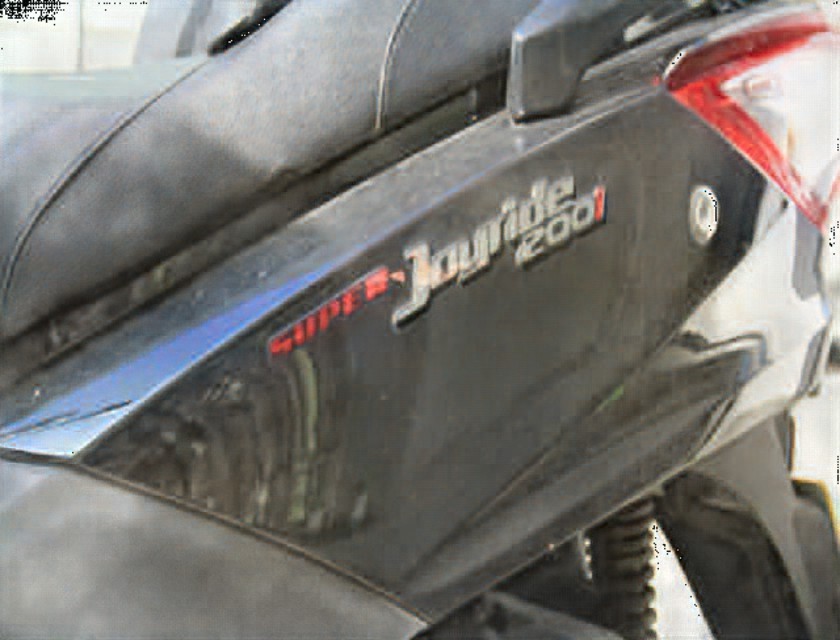}
\includegraphics[width=1.86cm]{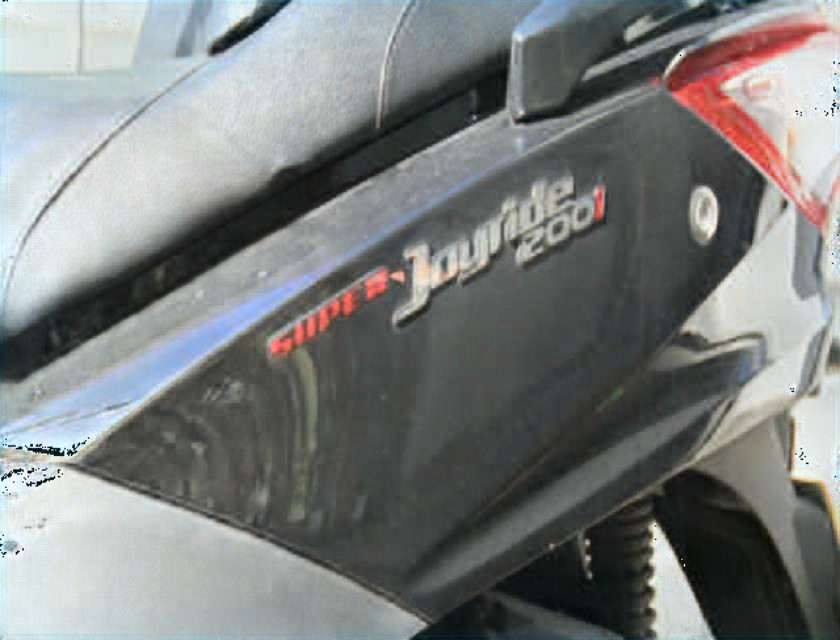}
\includegraphics[width=1.86cm]{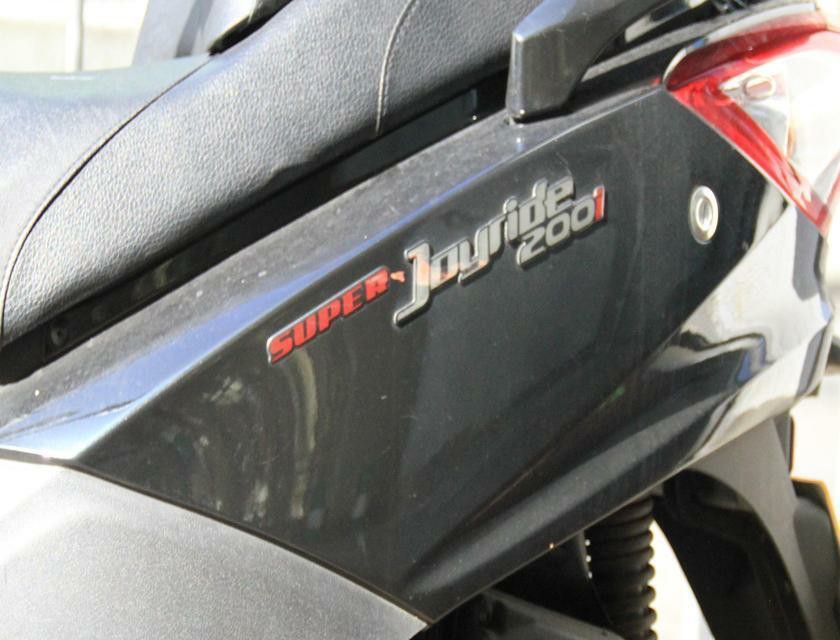}\\
\includegraphics[width=1.86cm]{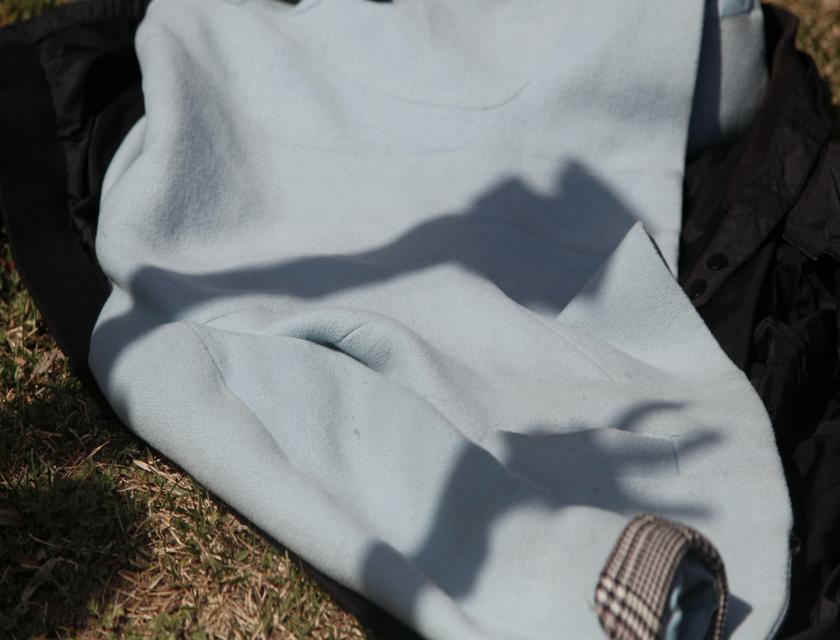}
\includegraphics[width=1.86cm]{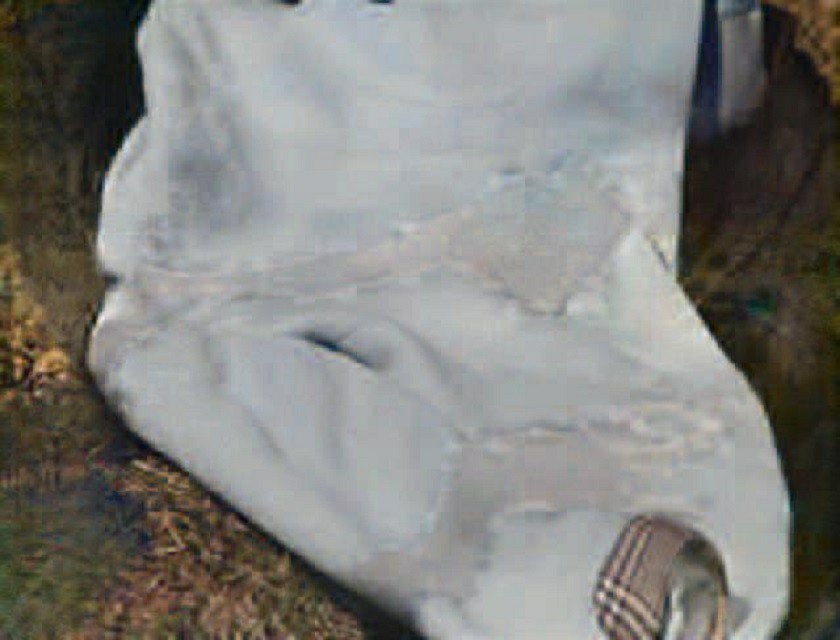}
\includegraphics[width=1.86cm]{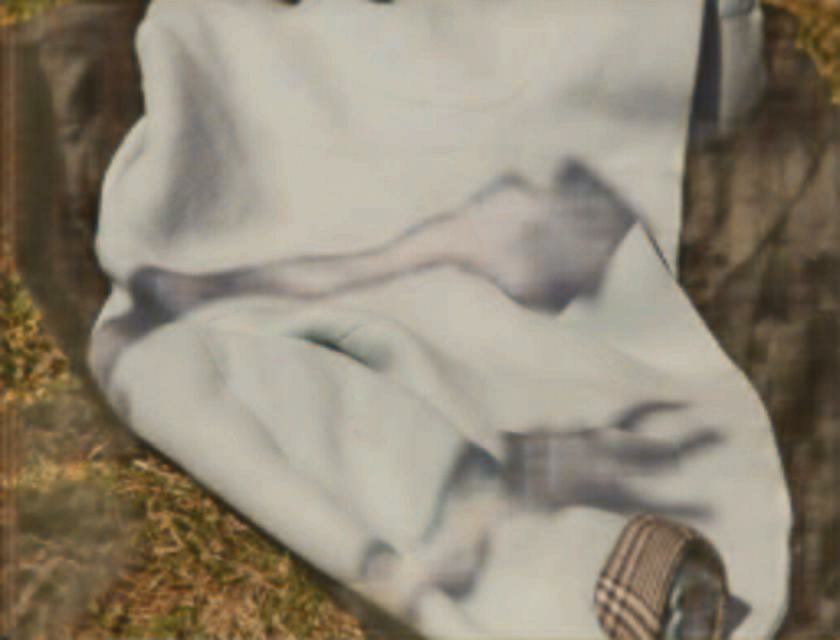}
\includegraphics[width=1.86cm]{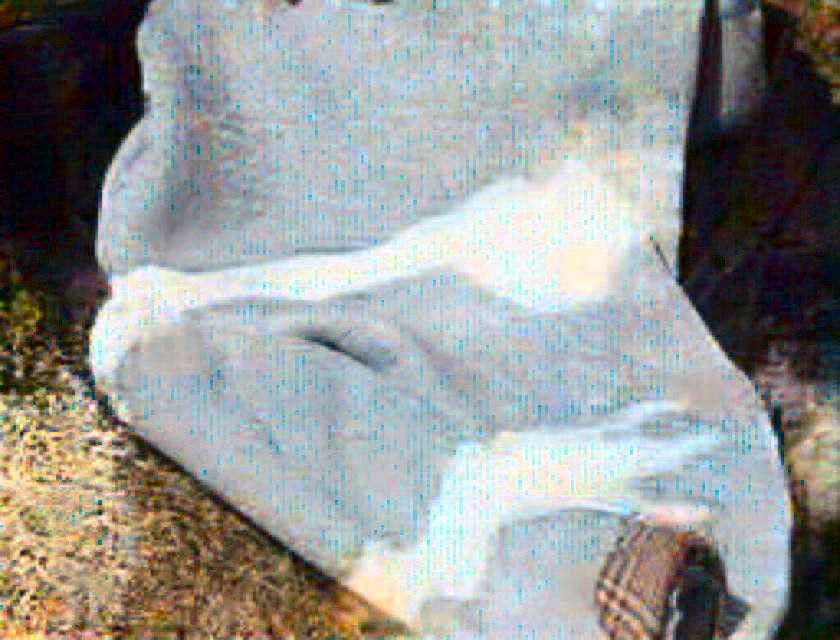}
\includegraphics[width=1.86cm]{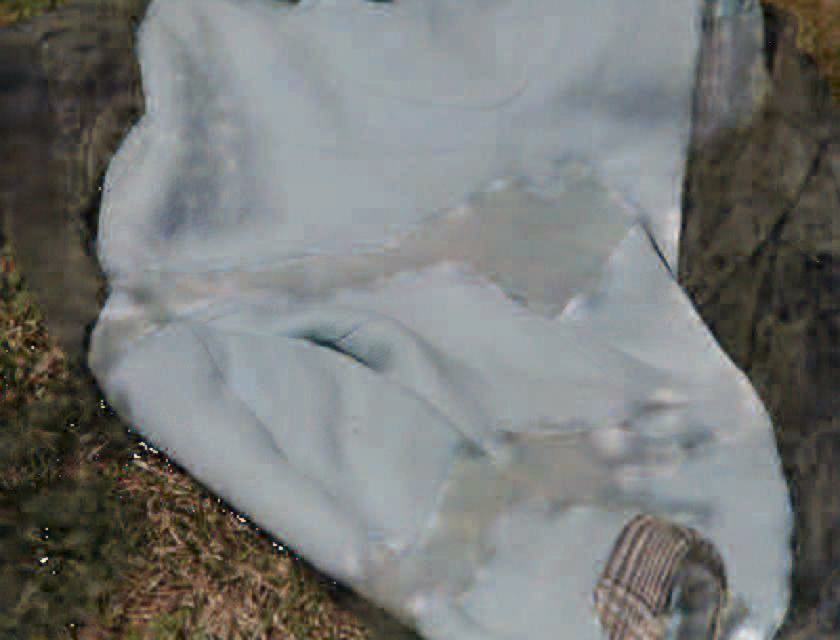}
\includegraphics[width=1.86cm]{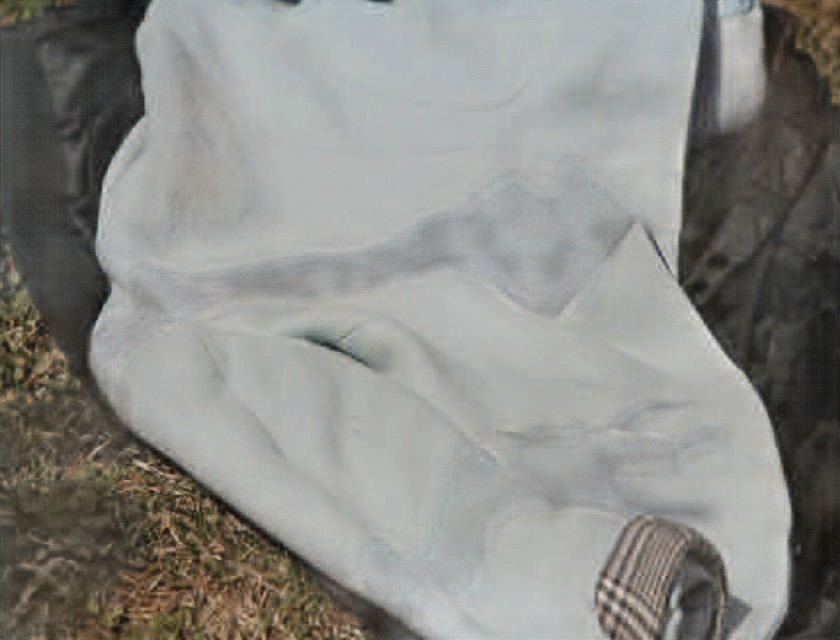}
\includegraphics[width=1.86cm]{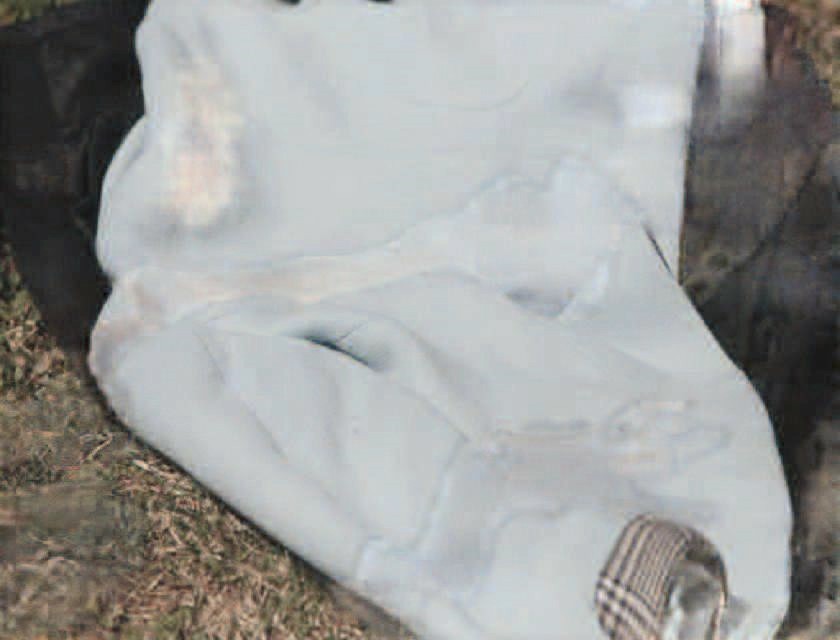}
\includegraphics[width=1.86cm]{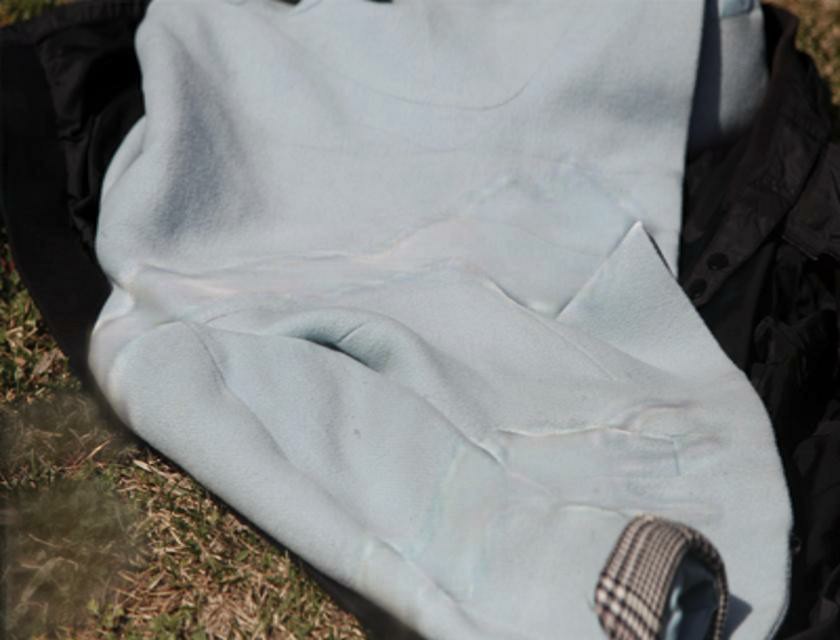}
\includegraphics[width=1.86cm]{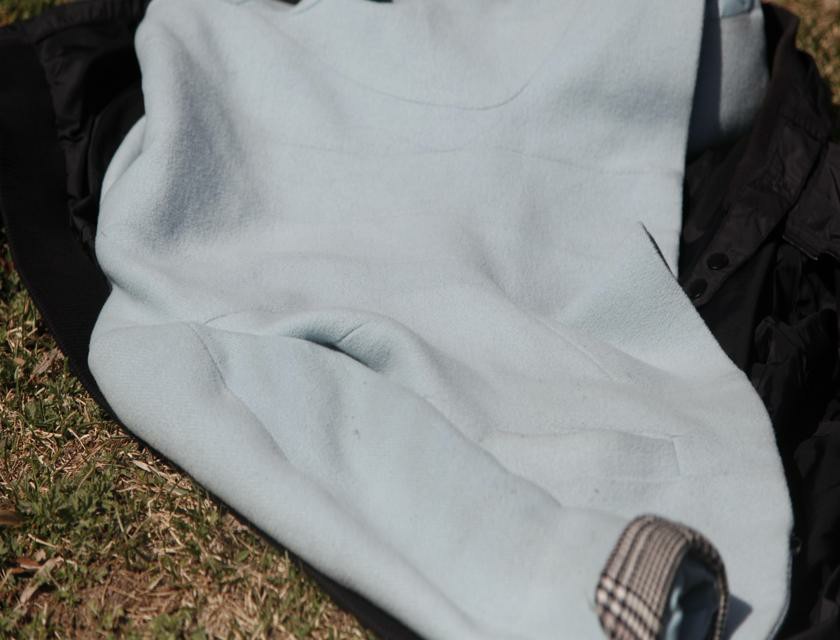}\\
\includegraphics[width=1.86cm]{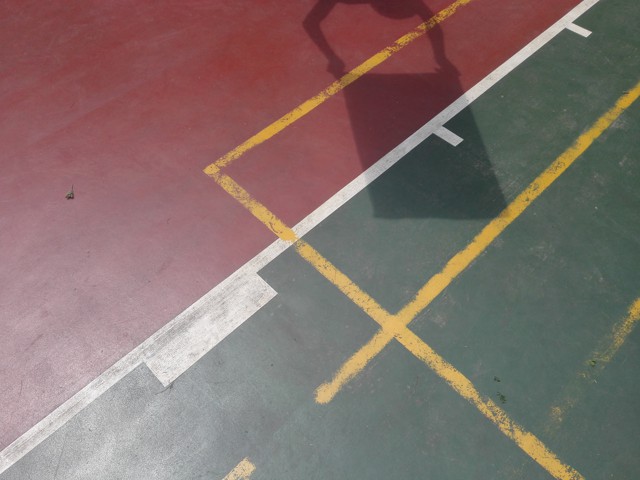}
\includegraphics[width=1.86cm]{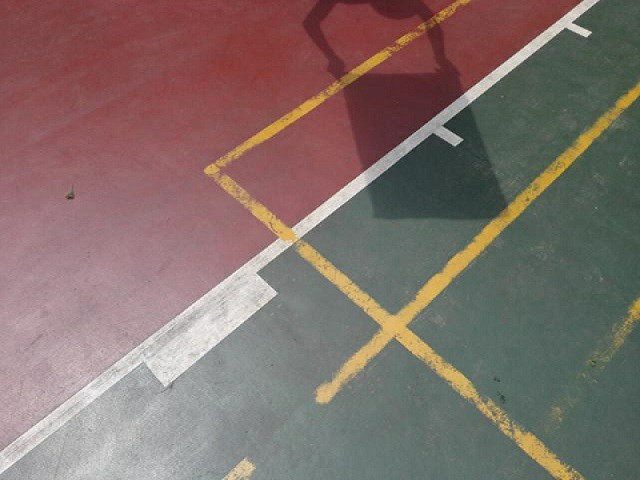}
\includegraphics[width=1.86cm]{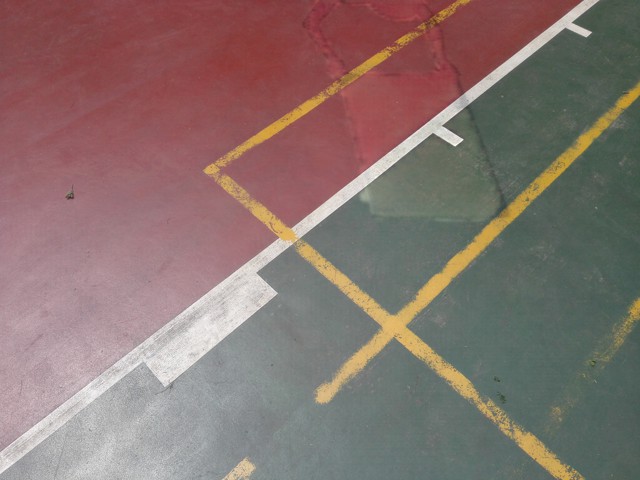}
\includegraphics[width=1.86cm]{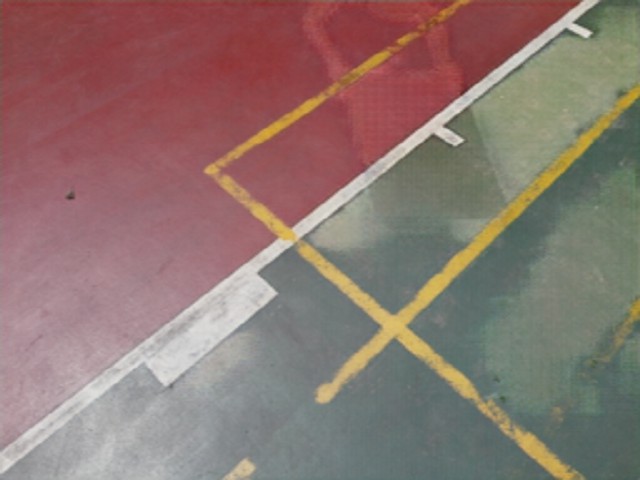}
\includegraphics[width=1.86cm]{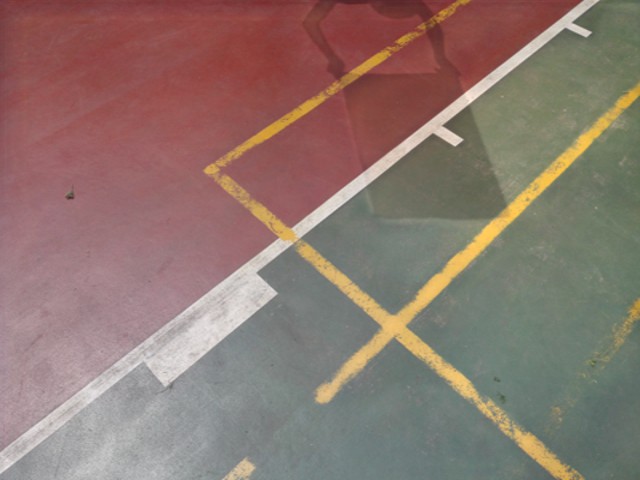}
\includegraphics[width=1.86cm]{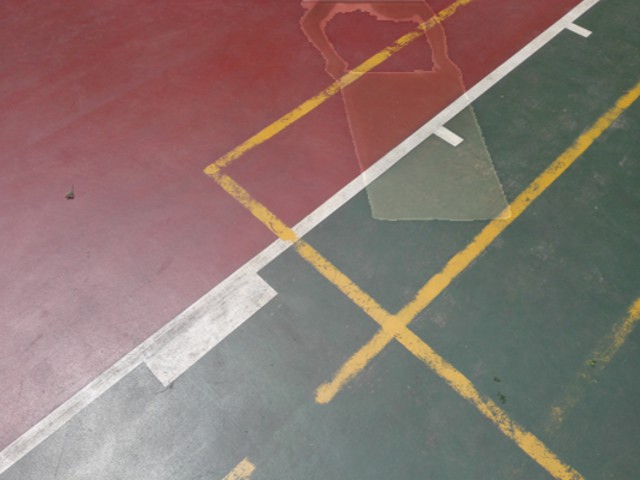}
\includegraphics[width=1.86cm]{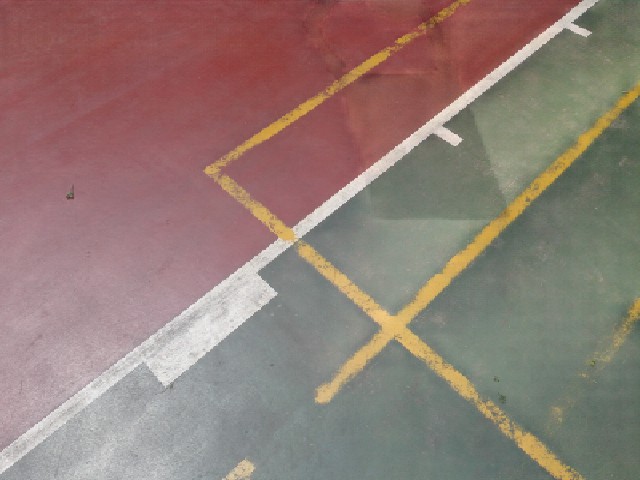}
\includegraphics[width=1.86cm]{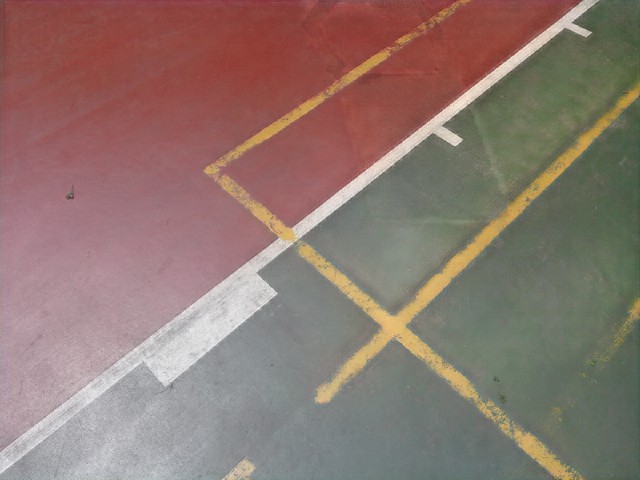}
\includegraphics[width=1.86cm]{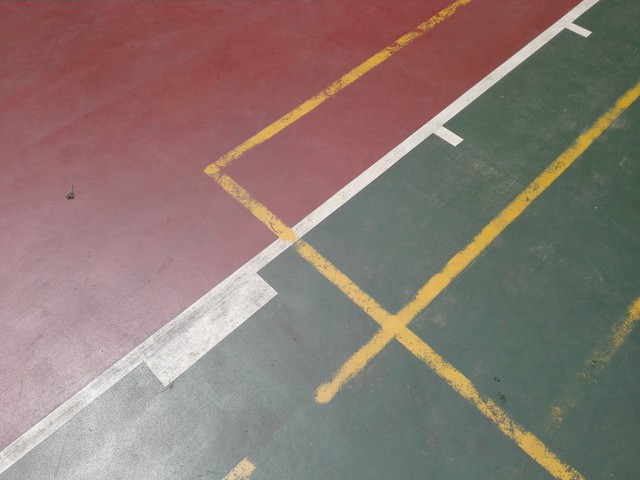}\\
\includegraphics[width=1.86cm]{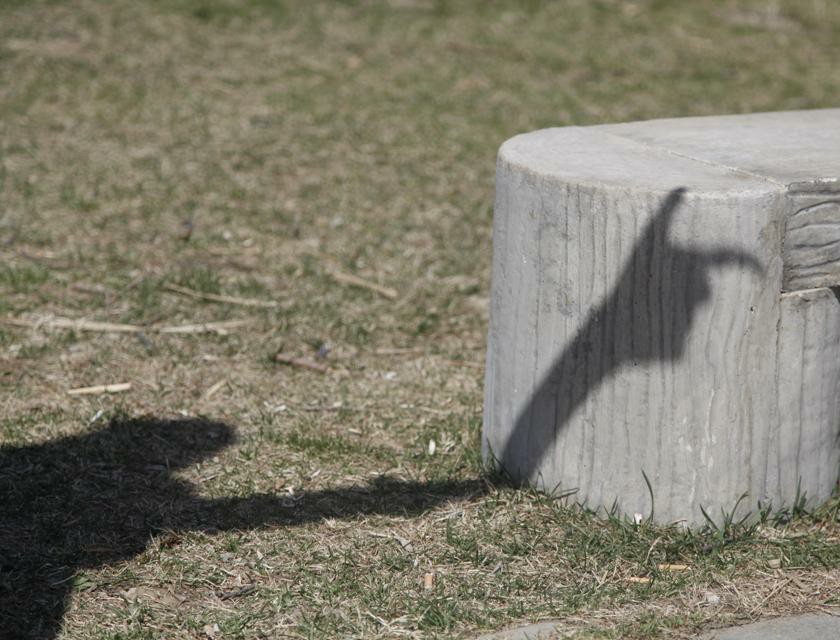}
\includegraphics[width=1.86cm]{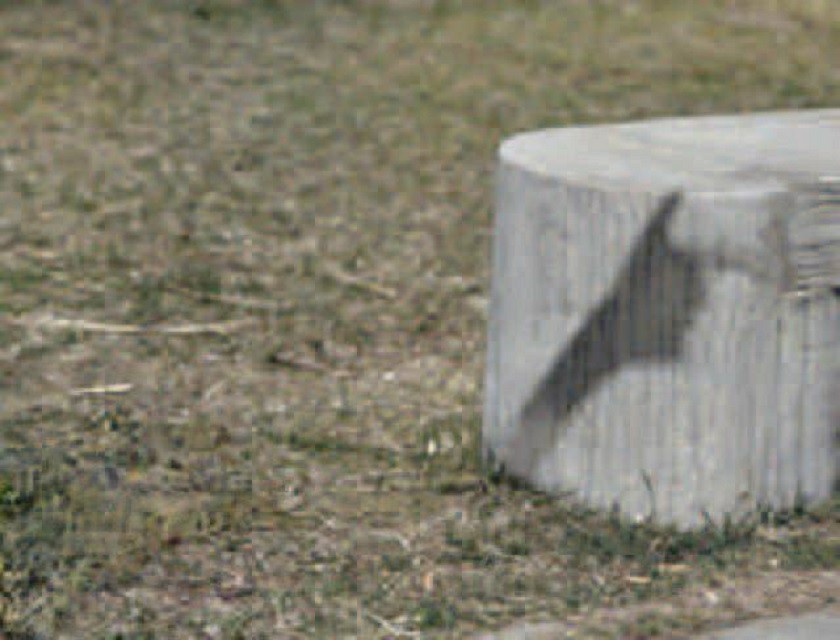}
\includegraphics[width=1.86cm]{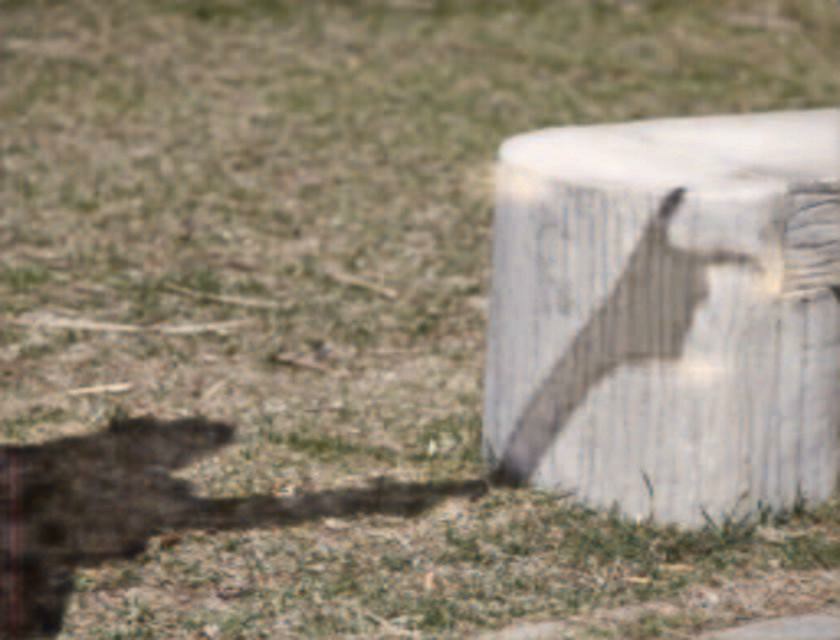}
\includegraphics[width=1.86cm]{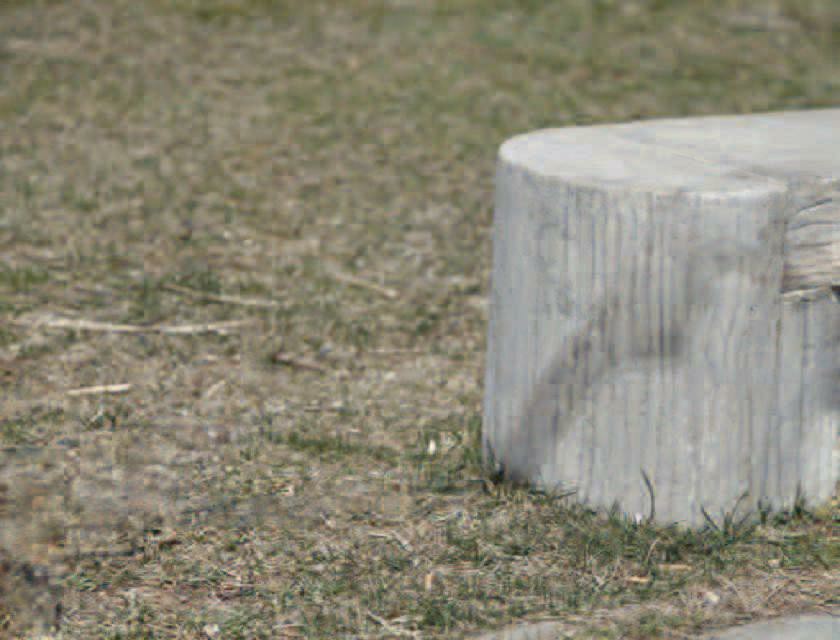}
\includegraphics[width=1.86cm]{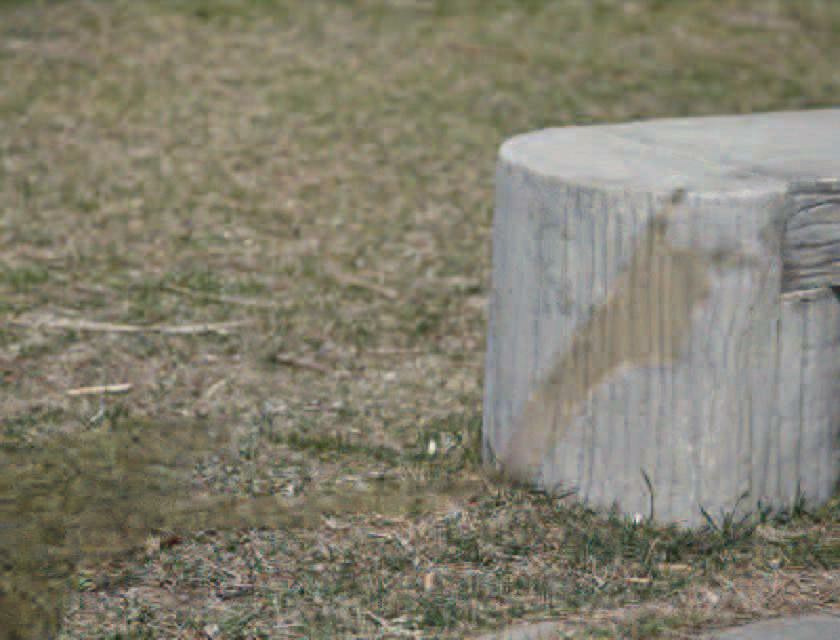}
\includegraphics[width=1.86cm]{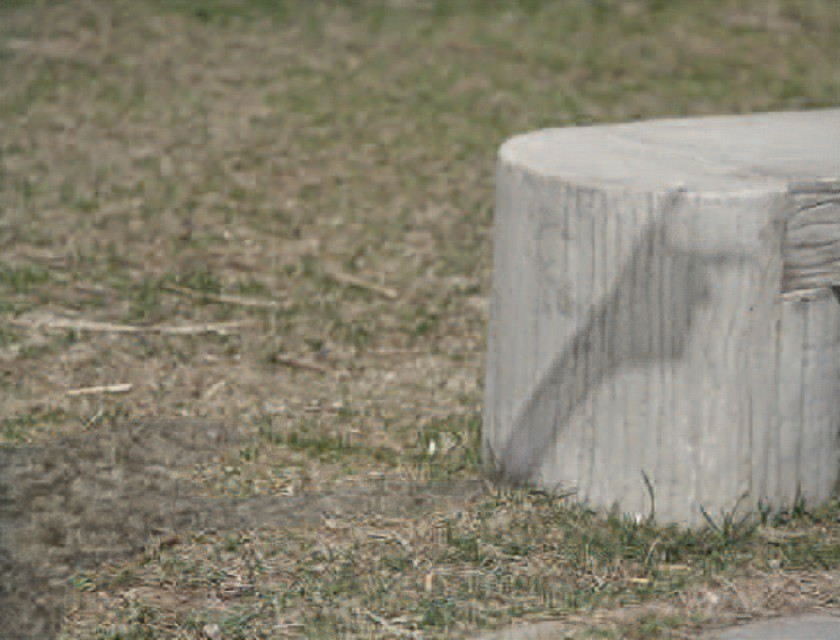}
\includegraphics[width=1.86cm]{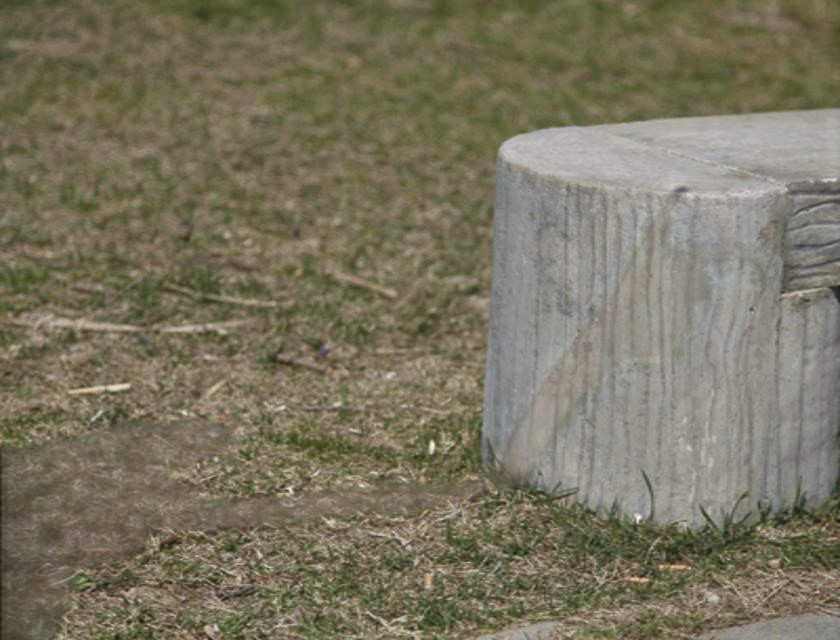}
\includegraphics[width=1.86cm]{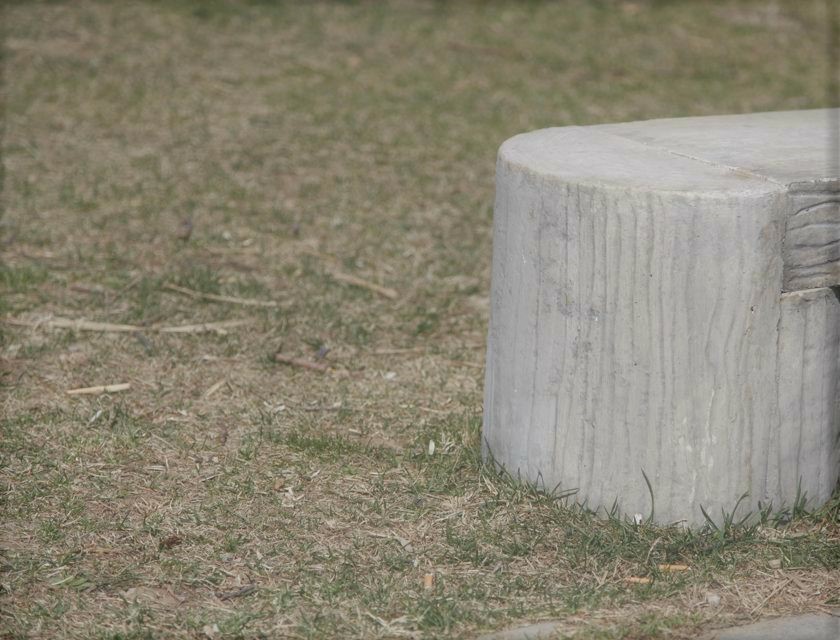}
\includegraphics[width=1.86cm]{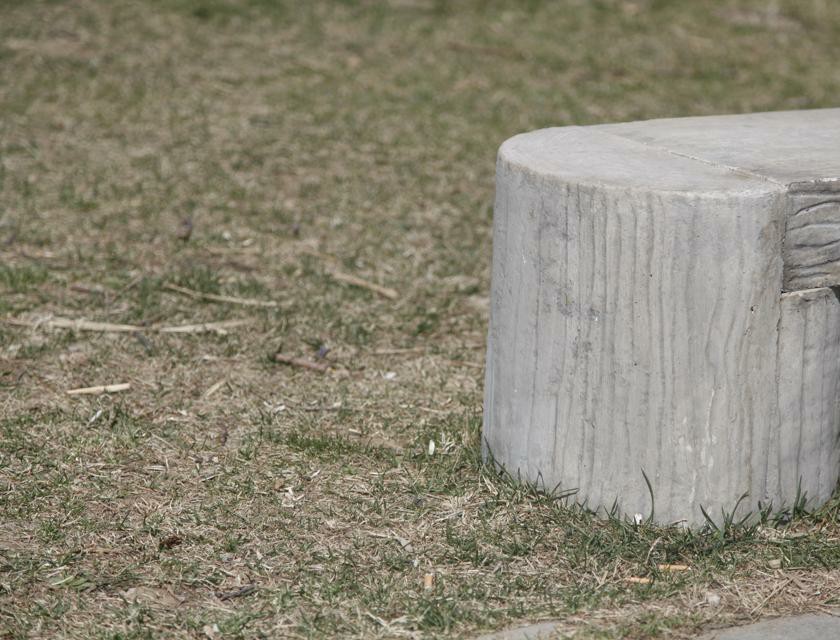}\\
\includegraphics[width=1.86cm]{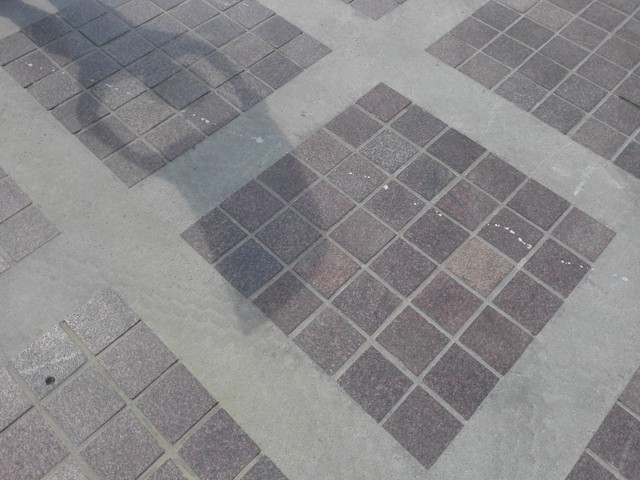}
\includegraphics[width=1.86cm]{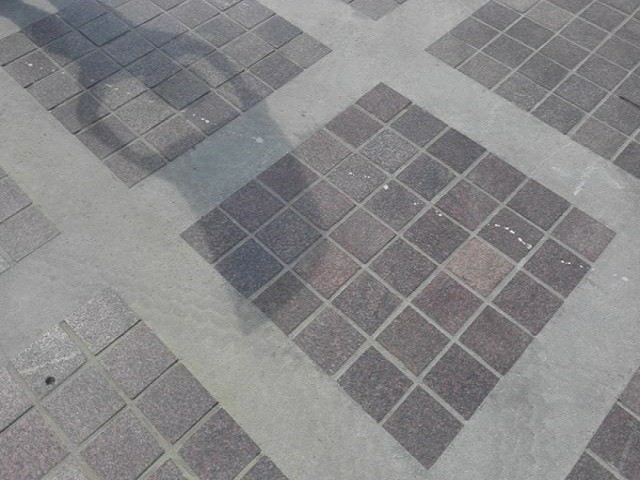}
\includegraphics[width=1.86cm]{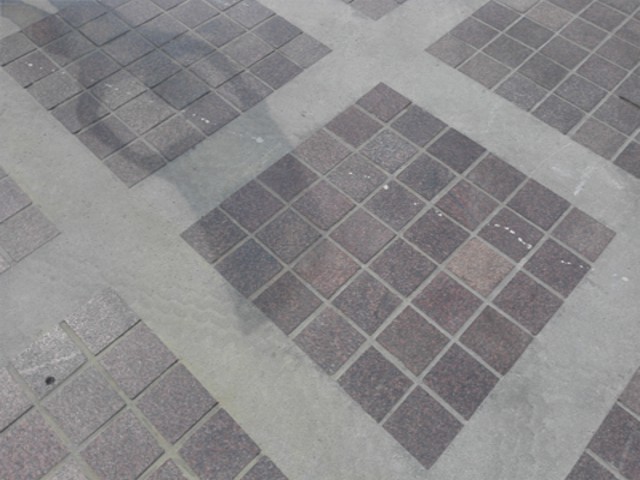}
\includegraphics[width=1.86cm]{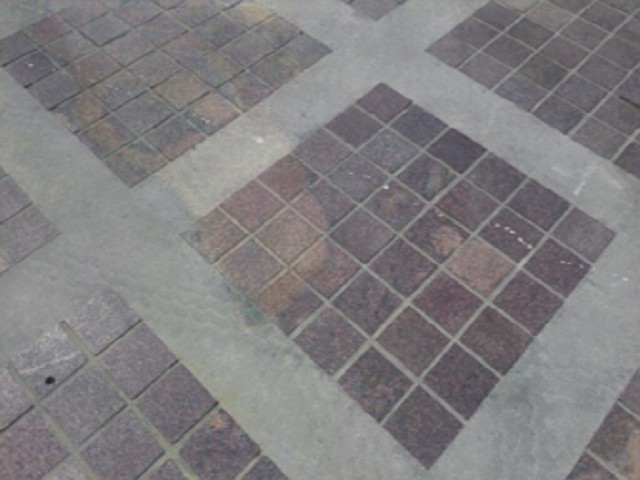}
\includegraphics[width=1.86cm]{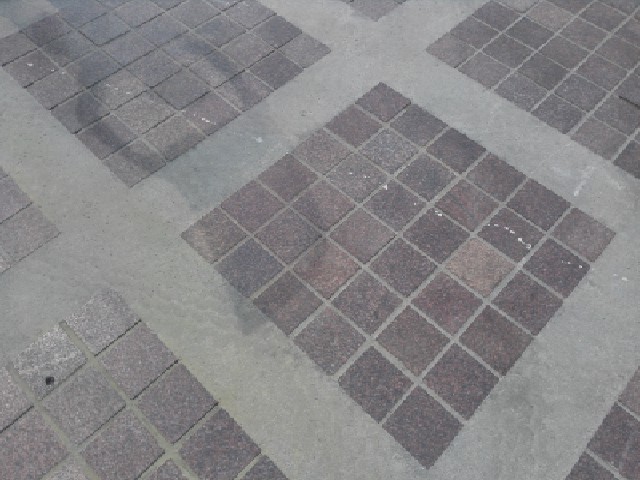}
\includegraphics[width=1.86cm]{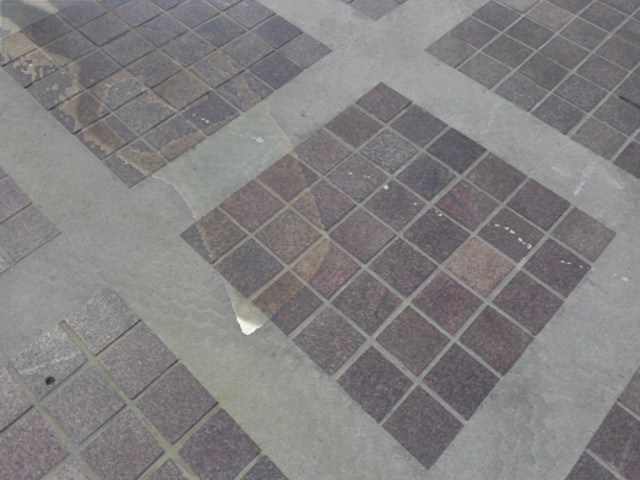}
\includegraphics[width=1.86cm]{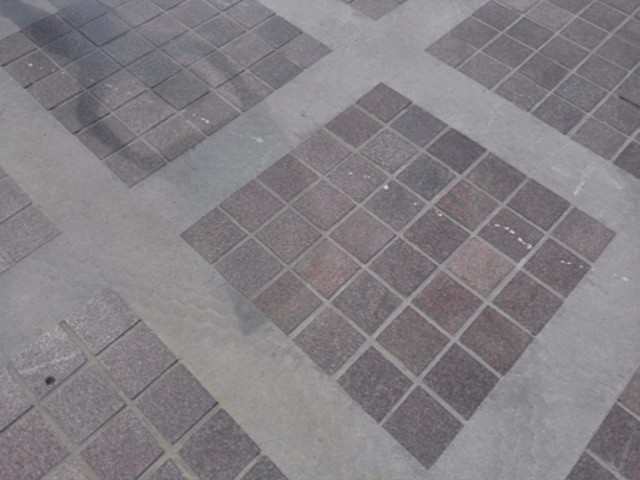}
\includegraphics[width=1.86cm]{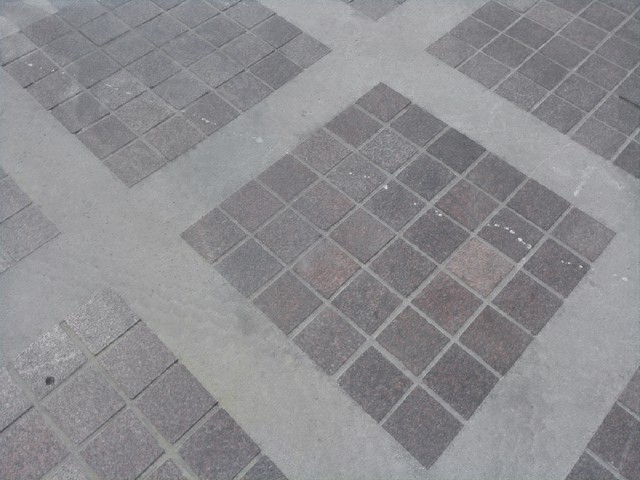}
\includegraphics[width=1.86cm]{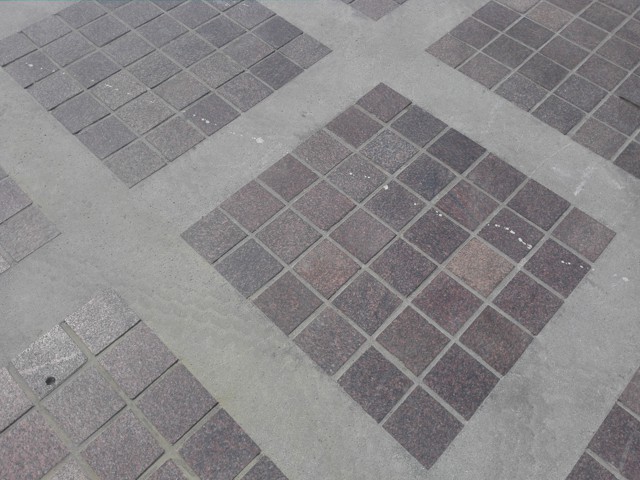}\\
\vspace{-0.15cm}
\subfigure[]{\includegraphics[width=1.86cm]{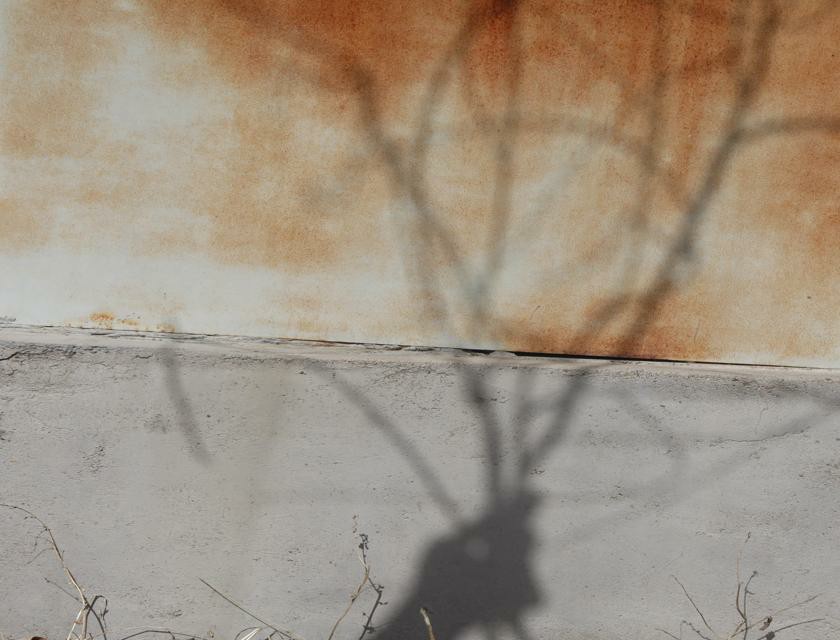}}
\subfigure[]{\includegraphics[width=1.86cm]{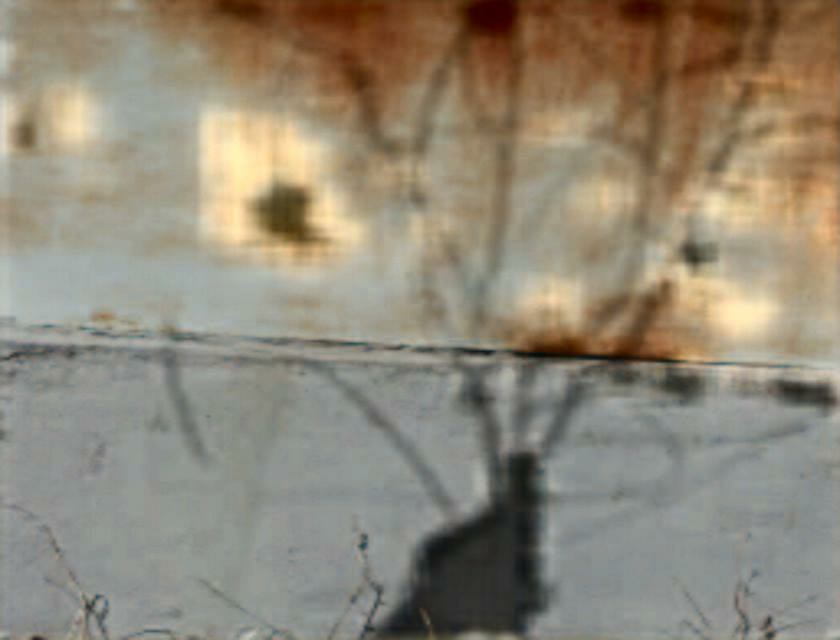}}
\subfigure[]{\includegraphics[width=1.86cm]{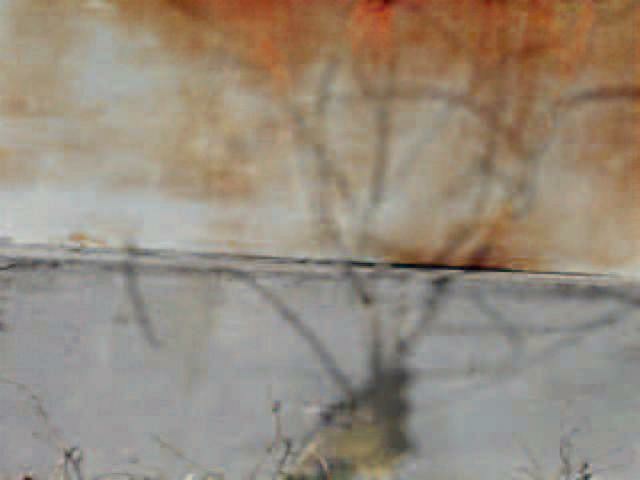}}
\subfigure[]{\includegraphics[width=1.86cm]{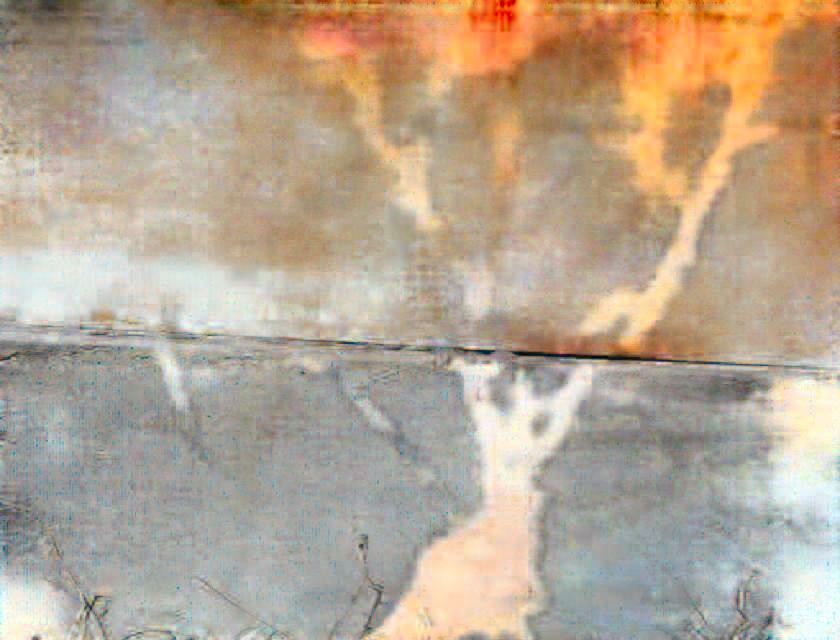}}
\subfigure[]{\includegraphics[width=1.86cm]{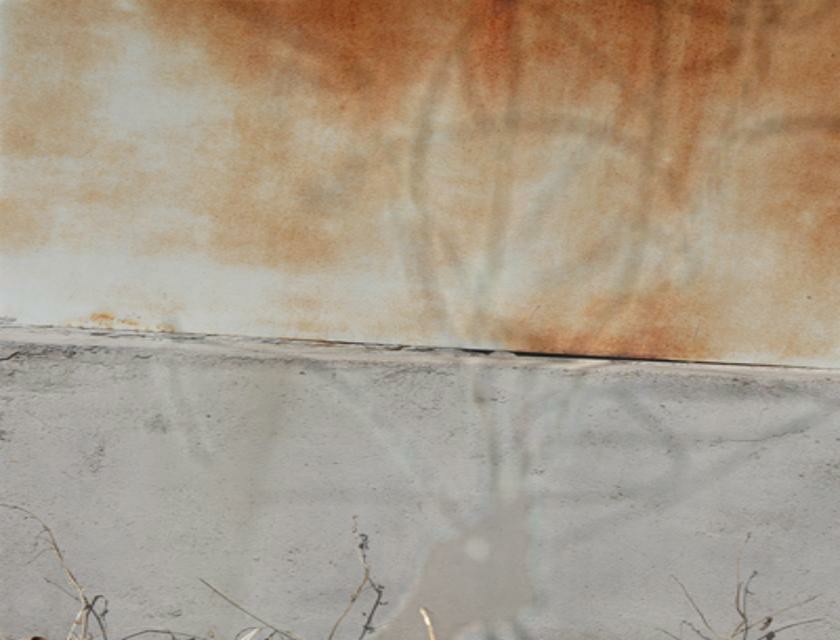}}
\subfigure[]{\includegraphics[width=1.86cm]{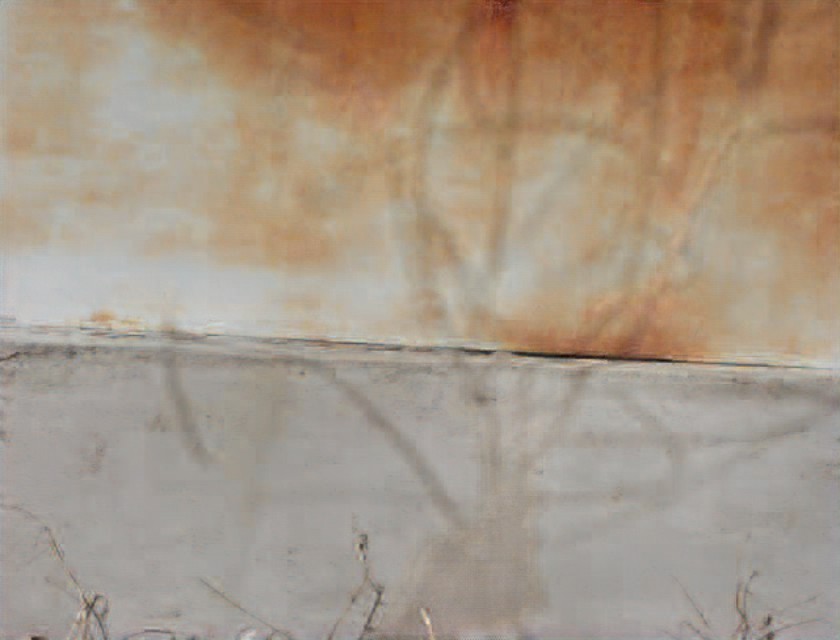}}
\subfigure[]{\includegraphics[width=1.86cm]{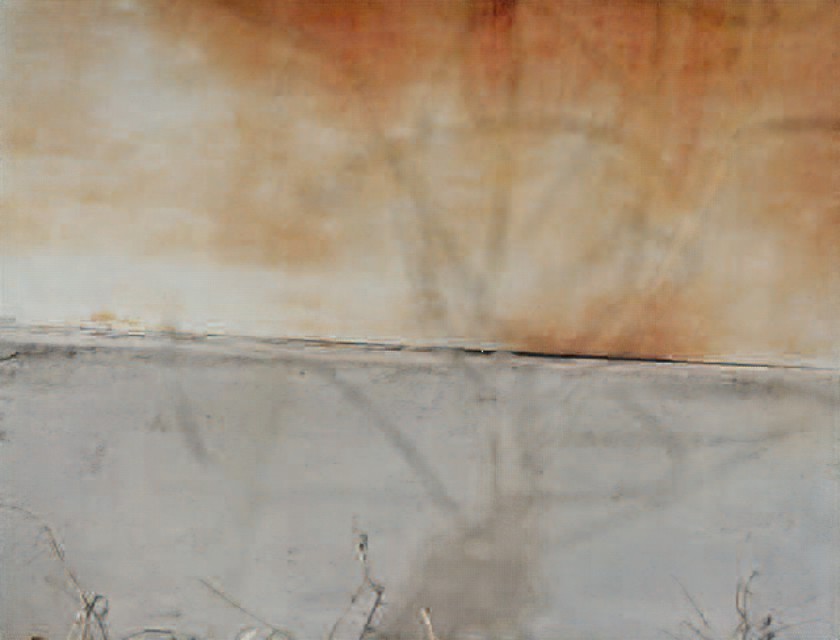}}
\subfigure[]{\includegraphics[width=1.86cm]{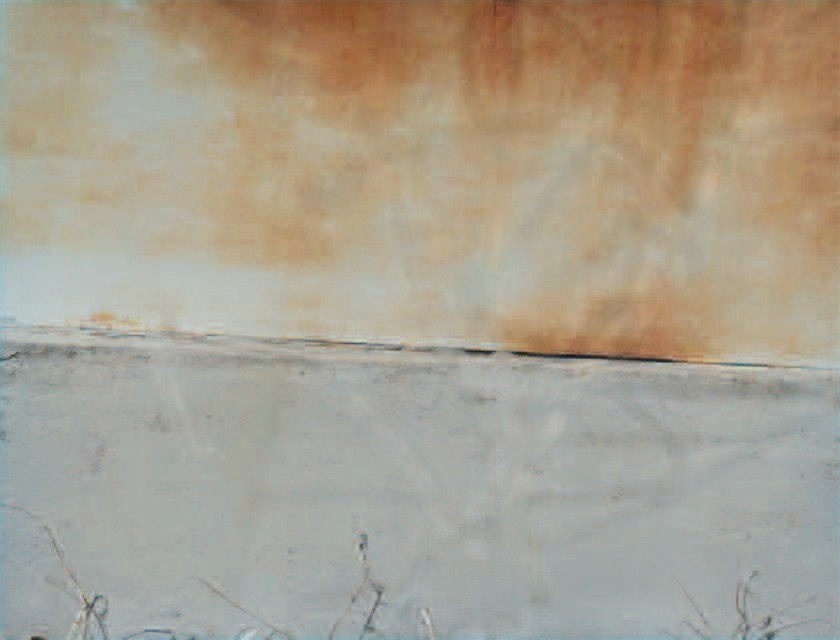}}
\subfigure[]{\includegraphics[width=1.86cm]{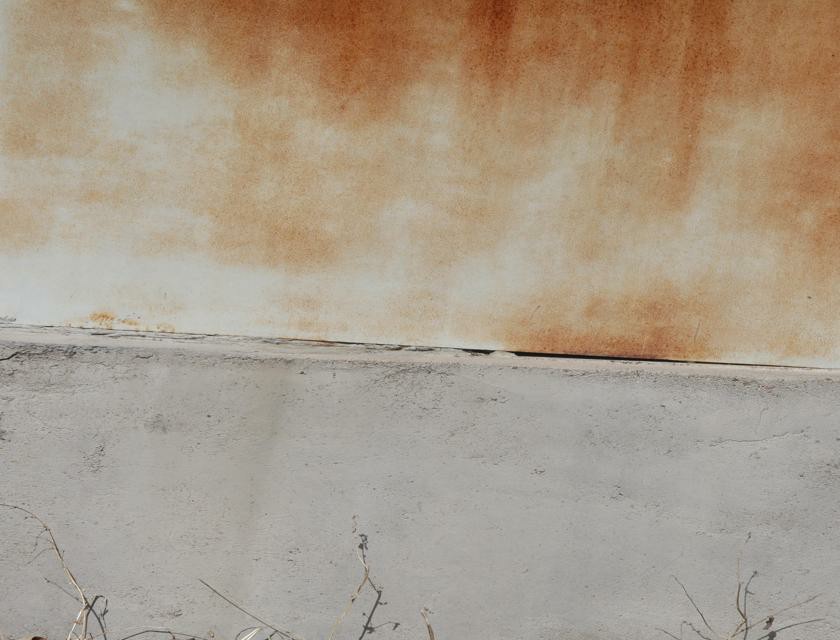}}
   \caption{Shadow removal results. From left to right are: (a) input images; shadow removal results of (b) Guo, (c) Zhang, (d) ST-CGAN, (e) DSC, (f) ARGAN, (g) RIS-GAN, (h) our CANet; and (i) the corresponding shadow-free ground truth images.
  }
\label{fig:compare}
\end{figure*}

\end{appendix}

\end{document}